\documentclass[conference]{IEEEtran}
\IEEEoverridecommandlockouts
\usepackage{cite}
\usepackage{amsmath,amssymb,amsfonts}
\usepackage{algorithm}  % add by kafeng
\usepackage{algorithmic}
\usepackage{graphicx}
\usepackage{textcomp}
\usepackage{xcolor}
\usepackage{graphics} % add by kafeng

\usepackage{float}  % add by kafeng
\usepackage{subfig} % add by kafeng
\usepackage{hyperref}  % add by kafeng  very good
\usepackage{color}  % add by 20210109
\usepackage{adjustbox} % 1007
\usepackage{multirow}
\newcommand{\com}[1]{\textcolor{black}{#1}} % for modify comments  % red  blue

\def\BibTeX{{\rm B\kern-.05em{\sc i\kern-.025em b}\kern-.08em
    T\kern-.1667em\lower.7ex\hbox{E}\kern-.125emX}}
\begin{document}

\newcommand{\TheName}{\emph{E-AFE}}
\title{ \TheName: Approximate Hashing for Automated Feature Engineering  \\
% {\footnotesize \textsuperscript{*}Note: Sub-titles are not captured in Xplore and should not be used}
\thanks{Identify applicable funding agency here. If none, delete this.}
}
\title{Accelerating Automated Feature Engineering via Sample Hashing}
\title{Toward Efficient Automated Feature Engineering}

% \author{
% \IEEEauthorblockN{Kafeng Wang}
% % {Kafeng Wang, Pengyang Wang, Chengzhong Xu}
% % {Kafeng Wang}
% % {1\textsuperscript{st} Kafeng Wang}
% \IEEEauthorblockA{\textit{Shenzhen Institute of Advanced Technology,} \\
% \textit{Chinese Academy of Sciences}\\
% \textit{University of Chinese Academy of Sciences} \\
% Shenzhen, China \\
% Email: kf.wang@siat.ac.cn}
% \and
% % \IEEEauthorblockN{2\textsuperscript{nd} Pengyang Wang}
% \IEEEauthorblockN{Pengyang Wang}
% \IEEEauthorblockA{\textit{State Key Laboratory of Internet of Things for Smart City} \\
% \textit{University of Macau}\\
% Macau, China \\
% Email: pywang@um.edu.mo}
% \and
% % \IEEEauthorblockN{3\textsuperscript{rd} Chengzhong Xu}
% \IEEEauthorblockN{Chengzhong Xu}
% \IEEEauthorblockA{\textit{State Key Laboratory of Internet of Things for Smart City,} \\
% \textit{and Department of Computer and Information Science} \\
% \textit{University of Macau}\\
% Macau, China \\
% Email: czxu@um.edu.mo}
% }

\author{\IEEEauthorblockN{Kafeng Wang\IEEEauthorrefmark{1}, Pengyang Wang \IEEEauthorrefmark{2} and Chengzhong xu\IEEEauthorrefmark{2}} \IEEEauthorblockA{\IEEEauthorrefmark{1}Shenzhen Institute of Advanced Technology, Chinese Academy of Sciences, and\\
University of Chinese Academy of Sciences, and \\
University of Macau, Macau, China, and \\
Guangdong-Hong Kong-Macao Joint Laboratory of Human-Machine Intelligence-Synergy Systems, Shenzhen, China } 
\IEEEauthorblockA{\IEEEauthorrefmark{2}State Key Laboratory of Internet of Things for Smart City, and \\
Department of Computer and Information Science, University of Macau, Macau, China \\
Email: kf.wang@siat.ac.cn, pywang@um.edu.mo, czxu@um.edu.mo}}

\maketitle

\begin{abstract} 
Automated Feature Engineering (AFE) refers to automatically generate and select optimal feature sets for downstream tasks, which has achieved great success in real-world applications. 
Current AFE methods mainly focus on improving the effectiveness of the produced features, but ignoring the low-efficiency issue for large-scale deployment. 
Therefore, in this work, we propose a generic framework to improve the efficiency of AFE. 
Specifically, we construct the AFE pipeline based on reinforcement learning setting, where each feature is assigned an agent to perform feature transformation \com{and} selection, and the evaluation score of the produced features in downstream tasks serve as the reward to update the policy. 
We improve the efficiency of AFE in two perspectives. 
On the one hand, we develop a Feature Pre-Evaluation (FPE) Model to reduce the sample size and feature size that are two main factors on undermining the efficiency of feature evaluation. 
On the other hand, we devise a two-stage policy training strategy by running FPE on the pre-evaluation task as the initialization of the policy to avoid training policy from scratch. 
We conduct comprehensive experiments on 36 datasets in terms of both classification and regression tasks. 
The results show $2.9\%$ higher performance in average and 2x higher computational efficiency comparing to state-of-the-art AFE methods.

\end{abstract}


\section{Introduction \label{section:introduction}}
Feature engineering refers to the process of feature generation and selection to convert raw data into effective features for machine learning tasks. 
Due to the lack of domain knowledge, traditional manual feature engineering is labor-intensive and time-consuming~\cite{shi2020safe}. 
To overcome the limitation, \underline{\textbf{A}}utomated \underline{\textbf{F}}eature \underline{\textbf{E}}ngineering (AFE) is proposed to automatically generate and select optimal feature sets~\cite{kaul2017autolearn,kotthoff2019auto}. 
On the one hand, AFE can significantly reduce human efforts and benefit the deployment in large-scale big data systems; on the other hand, AFE can also discover new knowledge from the data that is hardly achieved by traditional manual data engineering.

\begin{figure}[!t]
    \centering
    \vspace{-0.3cm}
    \subfloat[Sample Percentage vs. Performance.]{\includegraphics[width=0.23\textwidth, height=0.12\textheight]{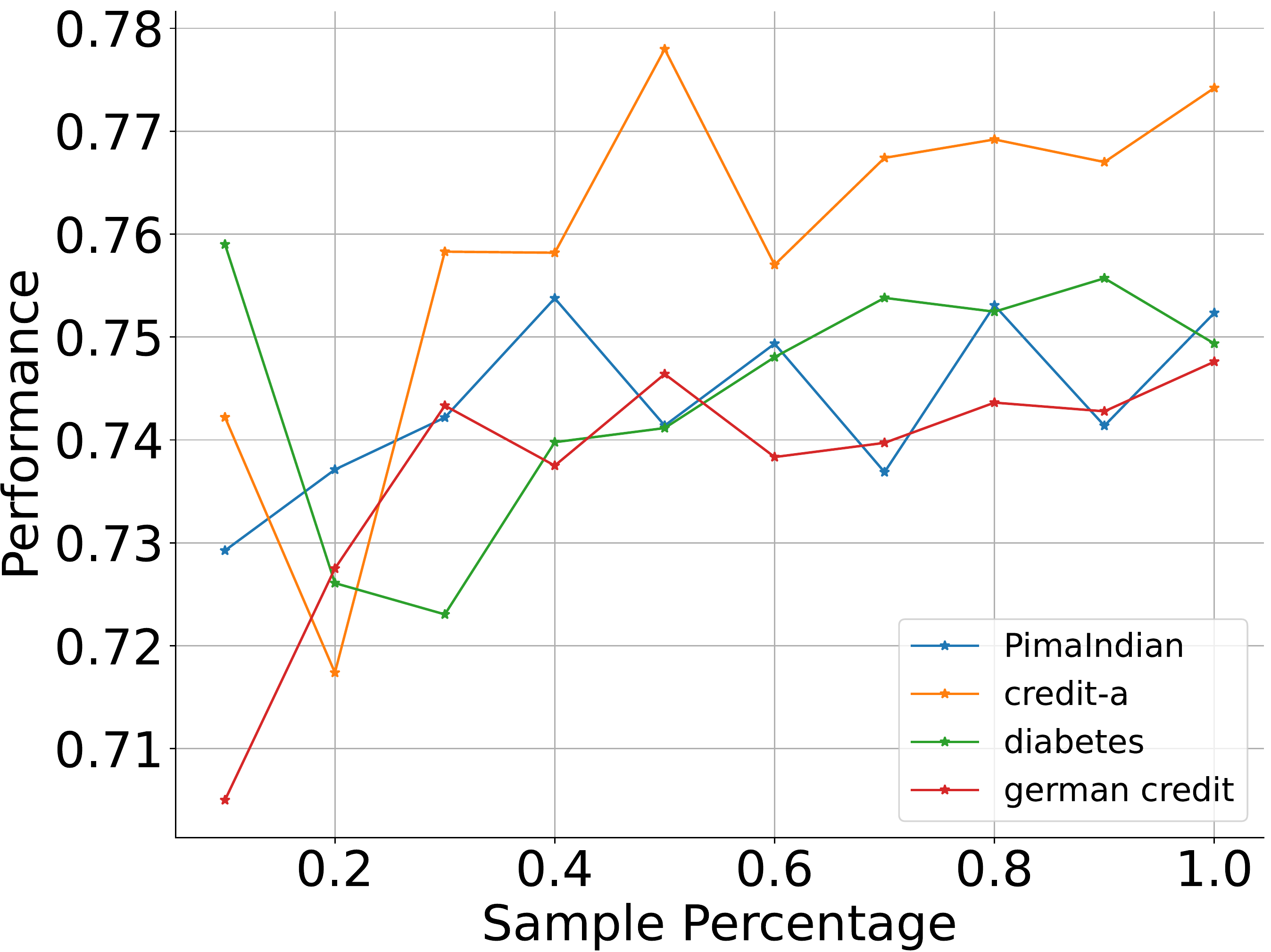}}
    \subfloat[Sample Percentage vs. Computation Time.]{\includegraphics[width=0.23\textwidth, height=0.12\textheight]{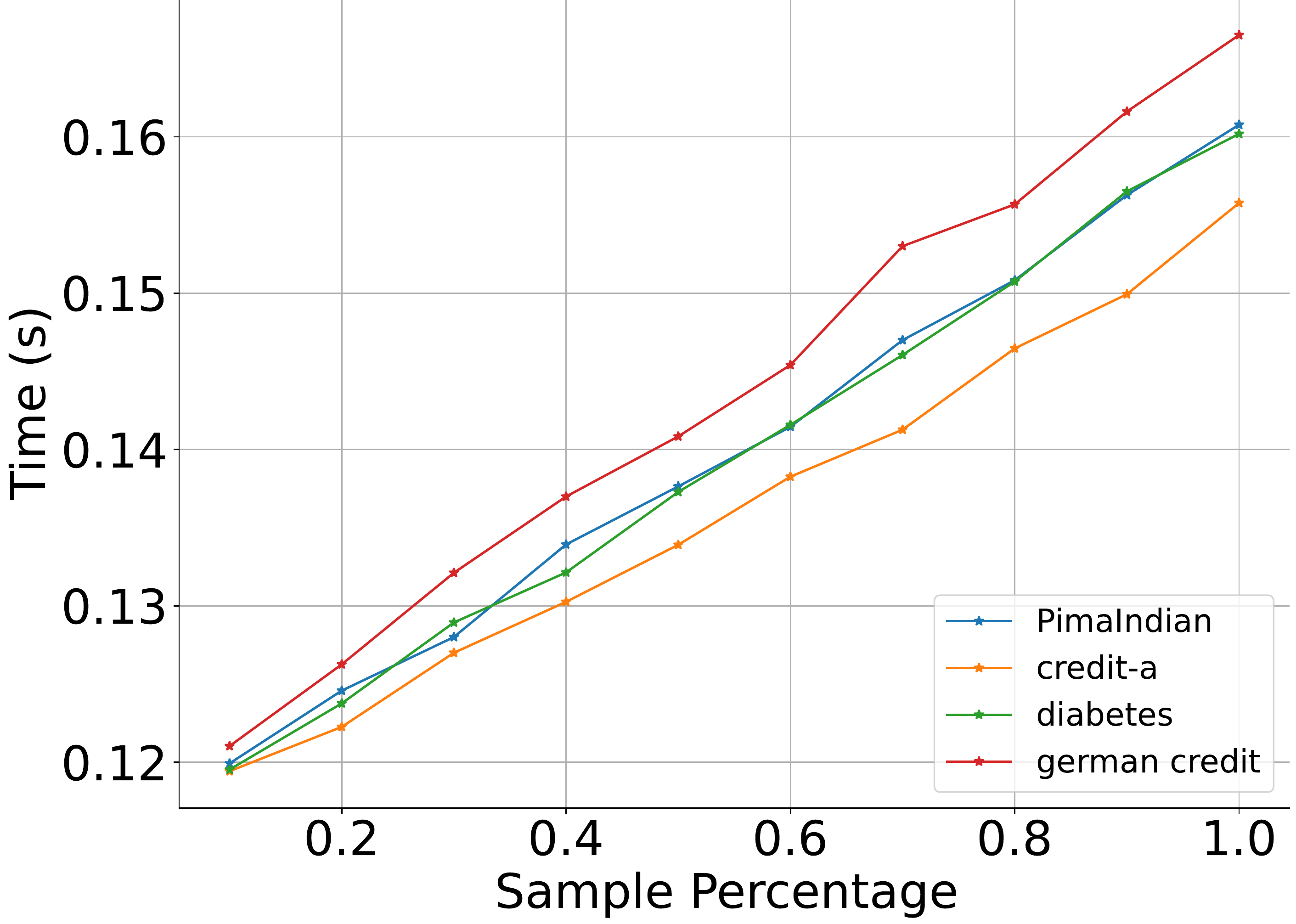}}\
    \vspace{-0.2cm}
    \caption{Sample size vs. performance and computation Time.}
    \label{fig:sample_score_time}
    \vspace{-0.5cm}
\end{figure}

Current studies in AFE focus on leveraging the Reinforcement Learning paradigm to explore possible feature candidates and exploit generated features to downstream machine learning tasks for selecting optimal feature sets. 
For example, 
Nargesian {\it et al.} proposed Learning Feature Engineering (LFE) with focusing on extracts useful transformations from past feature engineering experiences~\cite{nargesian2017learning}; 
Khurana {\it et al.} developed a new transformation graph with Directed Acyclic Graph (DAG) to represent relationships between different transformed versions of the data, and Q-learning was used to learn a performance-guided strategy for effective feature transformation from historical instances~\cite{khurana2018feature}. 
In order to resolve the feature space explosion problem on high-order transformations, Chen {\it et al.} proposed Neural Feature Search (NFS) that utilized Recurrent Neural Network (RNN)-based controller to transform each raw feature through a series of transformation functions, and policy gradient was used to train this controller \cite{chen2019neural}. 
% Although these state-of-the-art methods produce promising feature sets for downstream  machine learning tasks, few efforts are put on the efficiency of AFE pipeline, especially in terms of the running time.

\begin{table*}[!t]
\centering
\vspace{-0.5cm}
\begin{tabular}{cccccc}
\hline
Dataset  & Instances$\backslash$Features & New Features & Generation Time & Eval. New Features Time & Total Time \\
\hline
% hepatitis   & 155$\backslash$6 & 885  & 89ms  & 125s  &  142s \\
% labor  & 57$\backslash$8  & 1179 & 53ms  & 156s  &  177s \\
% fertility & 100$\backslash$9 & 1356  & 89ms  & 188s   &  211s \\
PimaIndian  & 768$\backslash$8 & 1195 & 354ms  & 205s  &  225s \\
credit-a   & 690$\backslash$6 & 890  & 192ms  & 139s  &  155s \\
diabetes  & 768$\backslash$8  & 1174 & 334ms  & 203s  &  223s \\
german credit & 1001$\backslash$24 & 3730 & 1291ms  & 654s   &  732s \\
\hline
\end{tabular}
\vspace{-0.2cm}
\caption{One epoch of NFS time consuming on four datasets.}
\vspace{-0.2cm}
\label{tab:time_comsuming}
\vspace{-0.45cm}
\end{table*}

While promising feature sets produced, these state-of-the-art AFE methods suffer from low-efficiency issue, especially in terms of  time consumption for complex machine learning tasks. 
Taking NFS as an example. 
We investigate the time consumption of NFS on four datasets with different sizes, and present the empirical results in Table~\ref{tab:time_comsuming}. 
The results show that for each dataset, only about 0.1\% of the time is spent on feature generation, but about 90\% of the time is consumed on evaluating new features. 
The observation indicates that the efficiency of AFE is largely impeded by the feature evaluation step. 
Moreover, with the sample size increasing, the feature evaluation would take much more time, which even exacerbates the low-efficiency issue. 
Unfortunately, few efforts are put on improving the efficiency of AFE methods.

Therefore, in this work, we aim to study how to accelerate AFE. 
As the feature evaluation step takes the most of time, we propose to improve the efficiency of AFE by optimizing the feature evaluation process. 
Specifically, we optimize the feature evaluation process from two perspectives: 
(1) reducing the sample size for evaluation; 
(2) reducing the size of candidate features for evaluation.
Next, we introduce the two perspectives as follows:

\noindent\textbf{(1) Reducing the sample size for evaluation}. 
In traditional pipeline of AFE, the feature evaluation is conducted over the whole dataset. 
However, not all samples are necessary for determining the quality of the generated features. 
We empirically investigate the contribution of the sample size on the feature performance. 
Specifically, we randomly sample different percentage of four datasets, and calculate the performance of the selected features and the corresponding running time.
We conduct the operation for ten times, and present the average results in Figure~\ref{fig:sample_score_time}. 
The results show that ignoring when the sample size achieves a certain scale, the performance of the selected features remains relatively stable. 
With more samples getting involved in, it barely enhances the quality of the selected features, but raises unacceptable surge in time consumption.  
The observation suggests that reducing the sample size is a promising direction for accelerating the feature evaluation process without sacrificing the feature quality.

\noindent\textbf{(2) reducing the size of candidate features for evaluation}. 
In addition to sample size, the size of candidate features is another important factor to affect the running time of evaluation. 
Intuitively, more features lead to larger candidate space for evaluation in feature selection, resulting in larger running time.
Moreover, traditional AFE methods directly evaluate the features on the downstream tasks. 
However, the downstream tasks are usually complex and time-consuming, which even significantly retards the feature evaluation process. 
Therefore, reducing the size of candidate features prior to the formal evaluation (downstream tasks) is a potential solution. 

Nevertheless, it is challenging to achieve the goal. 
The reasons are as follows: 
\underline{First}, since AFE aims to provide a generic feature-refinement pipeline for different datasets with various sizes, the reducing process demands to work for arbitrary sample sizes. 
\underline{Second}, different datasets have features with diverse semantic meanings, thus, how to determine a feature as a candidate for evaluation across datasets in a unified manner is non-trivial. 
\underline{Third}, it is even more challenging to combine reducing the size of samples and candidate features without violating each requirement.
Therefore, to tackle the above challenges, we propose to develop a simple and fast yet reasonable auxiliary evaluation task to pre-evaluate the validness of features to filter the candidate features prior to the formal evaluation. 
The proposed auxiliary evaluation task, namely Feature-Validness Task, is a binary classification task that justifies whether one feature is selected as a candidate for the formal evaluation. 
The Feature-Validness Task takes features as input, where one feature is denoted by respective values in samples. 
We pre-train a binary classification model, namely Feature Pre-Evaluation (FPE) Model, for the Feature-Validness Task using a group of public datasets as prior knowledge, and apply the well-trained model for pre-evaluation of the candidate features. 
In this case, all the features are required to be represented as a fixed size of samples. 
To achieve the goal, we further propose a Hashing-based method to project arbitrary sample sizes into a fixed number, which naturally fulfill the requirement for reducing sample sizes. 

Along this line, we propose an efficient AFE framework with accelerating AFE via reducing the size of samples and candidate features simultaneously. 
Specifically, we formulate AFE as an off-policy reinforcement learning problem following the convention of AFE. 
Different from existing AFE methods, generated features from agents would be first pre-evaluated by the pre-trained FPE model to reduce the feature size with compressing samples to improve the efficiency. 
Then, to further reduce the costs on exploring promising feature transformation actions, we develop a two-stage training strategy: (1) quick initialization with FPE model, and (2) formal training.
In stage 1, we only use FPE model as the evaluation to quickly discover promising feature transformation actions. 
During stage 1, the policy is initialized by borrowing knowledge from the pre-trained FPE model, and one replay buffer is constructed to record potentially good actions. 
After several epochs of training in stage 1, we formally convert the evaluation to the downstream tasks in stage 2. 
Trough this way, the proposed framework can improve the training efficiency by avoiding learning the policy from scratch. 
Our contributions can be listed as follows: 
\begin{itemize}
\item To the best of our knowledge, we are the first to study how to improve the efficiency of AFE.
\item Through empirical studies, we identify the core reason of the low-efficiency issue of AFE as the inefficient feature evaluation process. 
\item We propose an efficient AFE framework by reducing the size of samples and candidate features simultaneously through a faster auxiliary feature evaluation task with sample hashing. 
\item We develop a two-stage training strategy to improve the efficiency of policy learning. 
\item We conduct comprehensive experiments on 36 datasets in terms of both classification and regression tasks. The results show $2.9\%$ higher performance in average and 2x higher computational efficiency comparing to state-of-the-art methods.
\end{itemize}

\section{Problem Formulation}
% Pengyang
\begin{figure*}[!t]
\vspace{-0.2cm}
\centering
% \subfloat[General Process of Automated Feature Engineering]{\includegraphics[width=0.40\textwidth, height=0.23\textheight]{images/feature_engineering.pdf}}\
\subfloat[General Process of Automated Feature Engineering]{\includegraphics[width=0.35\textwidth, height=0.20\textheight]{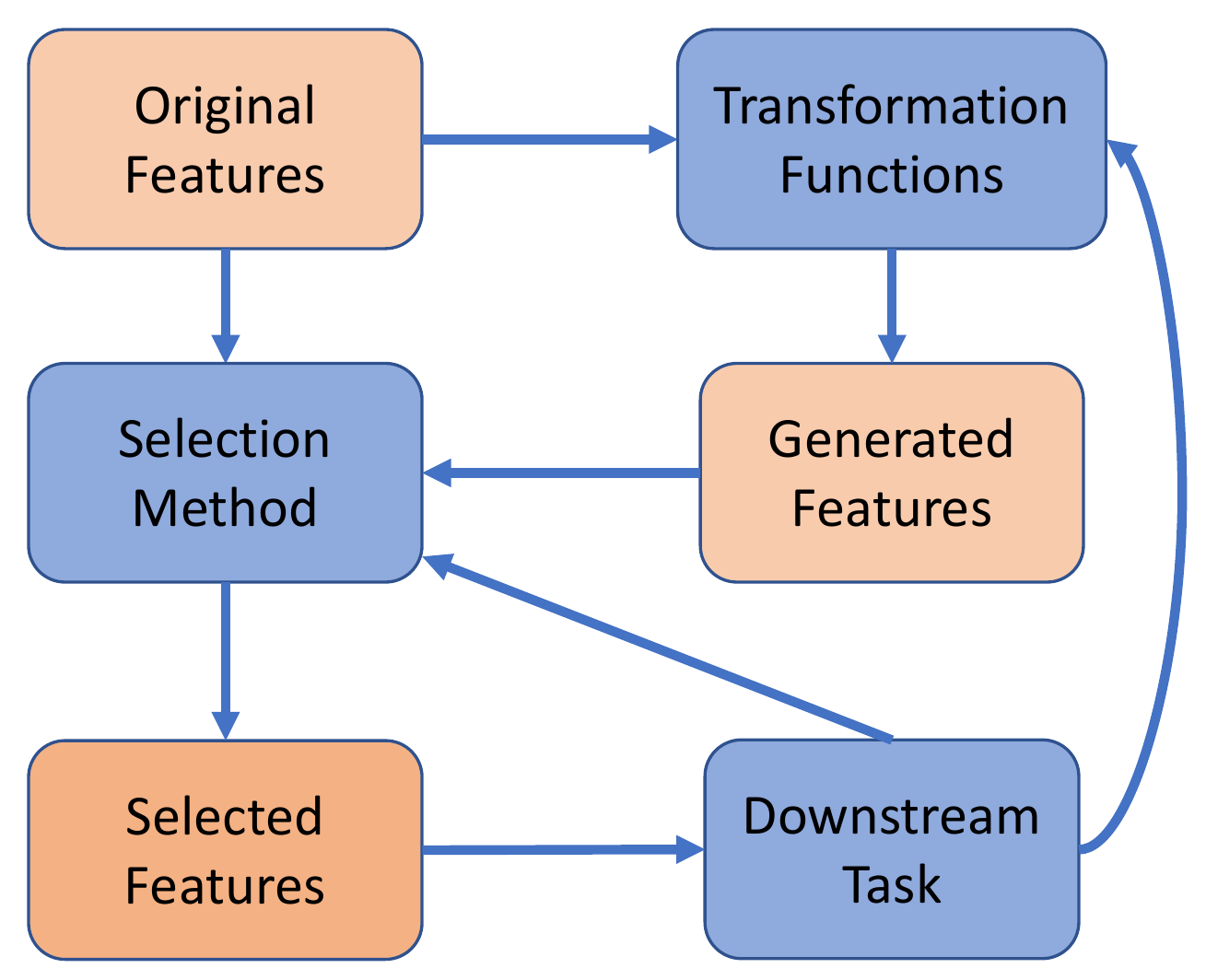}}\
% \subfloat[Automated Feature Engineering with RL]{\includegraphics[width=0.45\textwidth, height=0.23\textheight]{images/multi_agent.pdf}}\
\subfloat[Automated Feature Engineering with RL]{\includegraphics[width=0.40\textwidth, height=0.20\textheight]{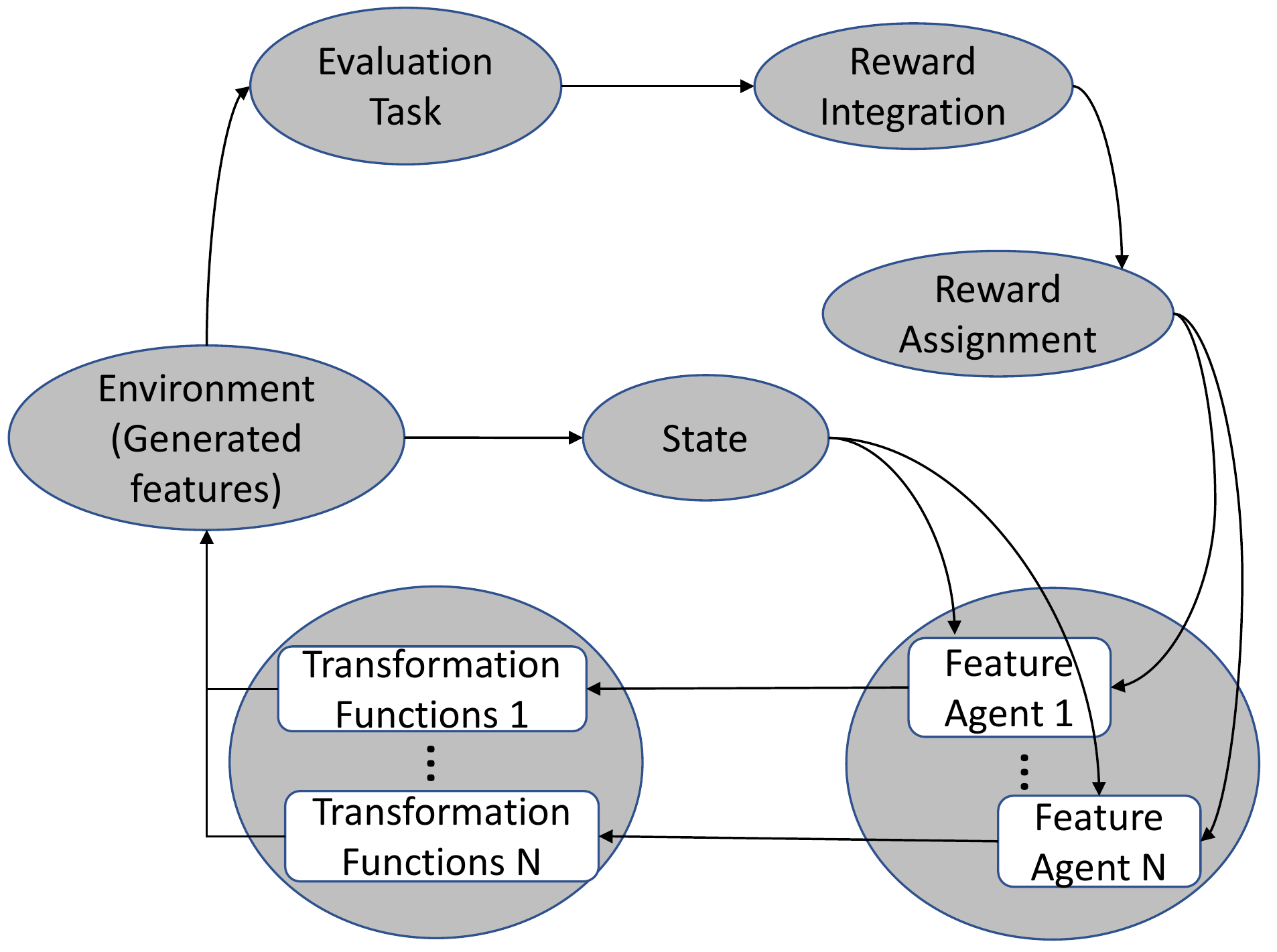}}\
\vspace{-0.2cm}
\caption{From traditional AFE to RL-based AFE.}
\label{fig:multi_agent_afe}
% \vspace{-0.6cm}
\vspace{-0.5cm}
\end{figure*}

In this section, we introduce the formulation of AFE. 
For tabular datasets, the general pipeline of AFE can be abstracted into two steps: feature generation and selection. 
% Feature generation is a combinatorial operation on the search space of hierarchical tree composed of features and operators.
As shown in Figure~\ref{fig:multi_agent_afe}(1), the original features are first transformed into generated features through a series of combinatorial transformation functions. 
\com{Then} the selection method \com{selects} effective features from the generated features based on the evaluation results in downstream tasks. 
Then, the evaluation results will be further used as the feedback to improve the selection method and transformation functions to continue generating and selecting features until the selected features satisfy the evaluation requirement for the downstream tasks. 
Formally, given a dataset $\mathcal{D} \langle \mathbf{F}, \mathbf{y}\rangle$ with $N$ original features $\mathbf{F}=\{f^{[1]}, f^{[2]},...,f^{[N]}\}$ with label $\mathbf{y}$, 
we use $A_{\mathcal{T}}(\mathbf{F},\mathbf{y})$ to represent the evaluation score of the downstream task $\mathcal{T}$ on dataset $\mathcal{D}$. 
%$\mathcal{T}$ is a downstream task (classification or regression). 
The original feature set $\mathbf{F}$ is transformed into $\mathbf{ \check{F} }$ through a set of transformation functions ({\it e.g.}, addition, multiplication, logarithm, etc). 
% The transformation is feature operator, such as addition, multiplication, logarithm. The transformation order is feature have been operated times.
The AFE is to find the optimal transformed feature set $\mathbf{F}^*$ where $A_{\mathcal{T}}(\mathbf{F}^*,\mathbf{y})$ is maximized, such that $\mathbf{F}^*=\underset{\mathbf{\check{F}}}{\mathrm{argmax}}{A_{\mathcal{T}}(\mathbf{\check{F}},\mathbf{y})}$.

In this work, we formulate AFE in a reinforcement learning setting, as shown in Figure~\ref{fig:multi_agent_afe}(2). 
We outline the definitions for the key elements as follows:
% Specifically,  AFE is a defined as a  Markov Decision Process (MDP) process, denoted by a tuple $\{\textbf{s}, \textbf{a}, \textbf{r}, \gamma, \textbf{p} \}$, where state space $\textbf{s}$ is finite, action space $\textbf{a}$ is pre-defined, reward function $\textbf{r}: \textbf{s} \times \textbf{a} \rightarrow \textbf{r}$ is a mapping function from state-action pair to a scalar, $\gamma \in [0, 1]$ is a discount factor and $\textbf{p}: \textbf{s}  \times \textbf{a}  \times \textbf{s}  \rightarrow \textbf{r}$ is the transition.% probability.
% The policy evaluation and improvement need many iterations, and each iteration 
% needs one episode $e$ ($e^i$ for the $i$-th iteration)  
% consisting of $T$ samples.

\begin{itemize}
\item \textbf{Environment.} 
In our design, the environment is the feature subspace of generated feature space. Whenever a feature agent issues an action to generate a new feature, the state of the feature subspace (environment) changes.

\item \textbf{Agents.} 
% Assuming there are $N$ features on the target dataset, we define $N$ agents for those features. For each agent, it is designed to make the generation decision for the corresponding feature.
\com{Suppose there are $N$ original features in the dataset, we define $N$ agents to generate features. Specifically, each agent will generate new features based on the given original feature. In other words, considering one original feature and the respective newly generated features as a subgroup of the state space, there are $N$ feature subgroups. Each agent will be responsible for one feature subgroup.}

% \item \textbf{State Space.} The state $s$ is to describe the selected feature subset. \com{This selected subset includes original feature and new features. Each agent has a state space. For every agent, the initialize state only selects the original feature. As new features are generated, the selected subset is changed, and the state space is extended. For one agent, supposed there are $N$ original features, $M$ operators, which include unary and binary transformation, and $K$ feature transformation orders. The state space dimension is $R^{N \cdot M \cdot K}$. In our state, the agent will select new and original features at the same time. However, new features did not replace the original feature.}
\item \textbf{State Space.} \com{We define the state $s$ as the selected features. The dimension (size) of the state space is the size of the selected feature. The selected features include the original features and the newly generated features. Once a newly generated feature is discriminated as a good feature, the feature will be included into the state. Therefore, with more newly generated features added into the state, the state space will keep expanding until the feature engineering process finishes.}

\item \textbf{Action.} 
% We define the action as the feature transformation. \com{Each agent will pick one transformation to generate a new feature. Each transformation consists of an operator and one or zero features. If it is a unary operator, there is no feature.} We use 4 unary operations, such as logarithm, min-max-normalization, square root, and reciprocal, and 5 binary operations, such as addition, subtraction, multiplication, division, and modulo operation.
\com{We define the action $a$ as the feature transformation. Each agent will take actions (feature transformation) over the respective feature subgroup. The feature transformation is in the format of $\text{OPERATOR}(\text{feature}_1, \text{feature}_2)$, where $\text{OPERATOR}$ is a feature transformation operator that takes two features as input and outputs a new feature, and $\text{feature}_1$ and $ \text{feature}_2$ are two features from one feature subgroup. We consider two types of feature transformation operators:
(i) unary operator (including logarithm, min-max-normalization, square root, and reciprocal), in this case, $\text{feature}_1$ and $\text{feature}_2$ are the same feature; 
(ii) binary operator (including addition, subtraction, multiplication, division, and modulo operation), in this case, $\text{feature}_1$ and $\text{feature}_2$ are two different features.}

\item \textbf{Transition.} 
% \com{After the agent operates an action, a new feature is generated, and the agent selects a new feature subset and transitions to the next state.}
\com{We illustrate the transition process in Figure~\ref{fig:transition}. 
For simplicity, we take the transition from $s_t$ to $s_{t+1}$ as an example. 
The agent first sample two features from the respective feature subgroup in $s_t$ with replacement. 
Then, the agent takes actions according to the policy to generate new features. 
The newly generated features will be discriminated as qualified or unqualified. 
Once confirmed as qualified, the newly generated feature will be selected and added to the respective feature subgroup to construct new state $s_{t+1}$.}

\begin{figure}[!t]
    \centering
    \vspace{-0.2cm}
    \includegraphics[scale=0.34]{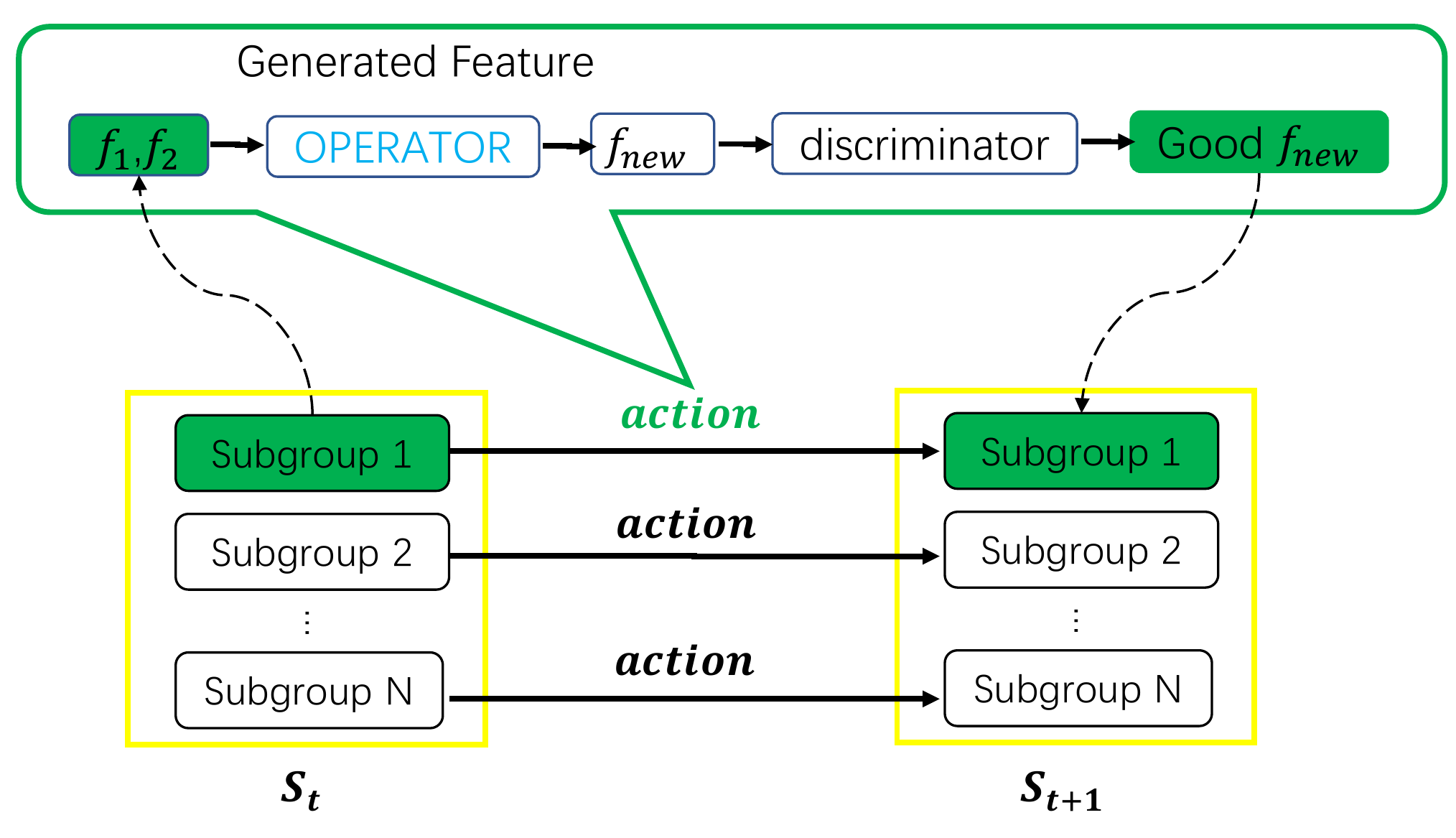}
    % \vspace{-0.3cm}
    \vspace{-0.2cm}
    \caption{\com{The transition process.}}
    % \vspace{-0.6cm}
    \vspace{-0.4cm}
    \label{fig:transition}
\end{figure}

\item \textbf{Evaluation Task.} 
The evaluation task aims to examine the effectiveness of the generated and selected features. 
There are two types of evaluation tasks: (1) pre-evaluation task, that is to reduce the feature size through quick binary classification; and (2) downstream task, that is the formal evaluation task to evaluate and select the generated features. 
Following the convention of AFE~\cite{chen2019neural}, we utilize Random Forest as the model for downstream tasks.
Details will be introduced in Section~\ref{section:FPE}.

\item \textbf{Reward.} 
We define the performance gain on the evaluation task as the reward $r$. 

\end{itemize}

\begin{figure}[h]%[!t]
    \centering
    \vspace{-0.2cm}
    \includegraphics[scale=0.4]{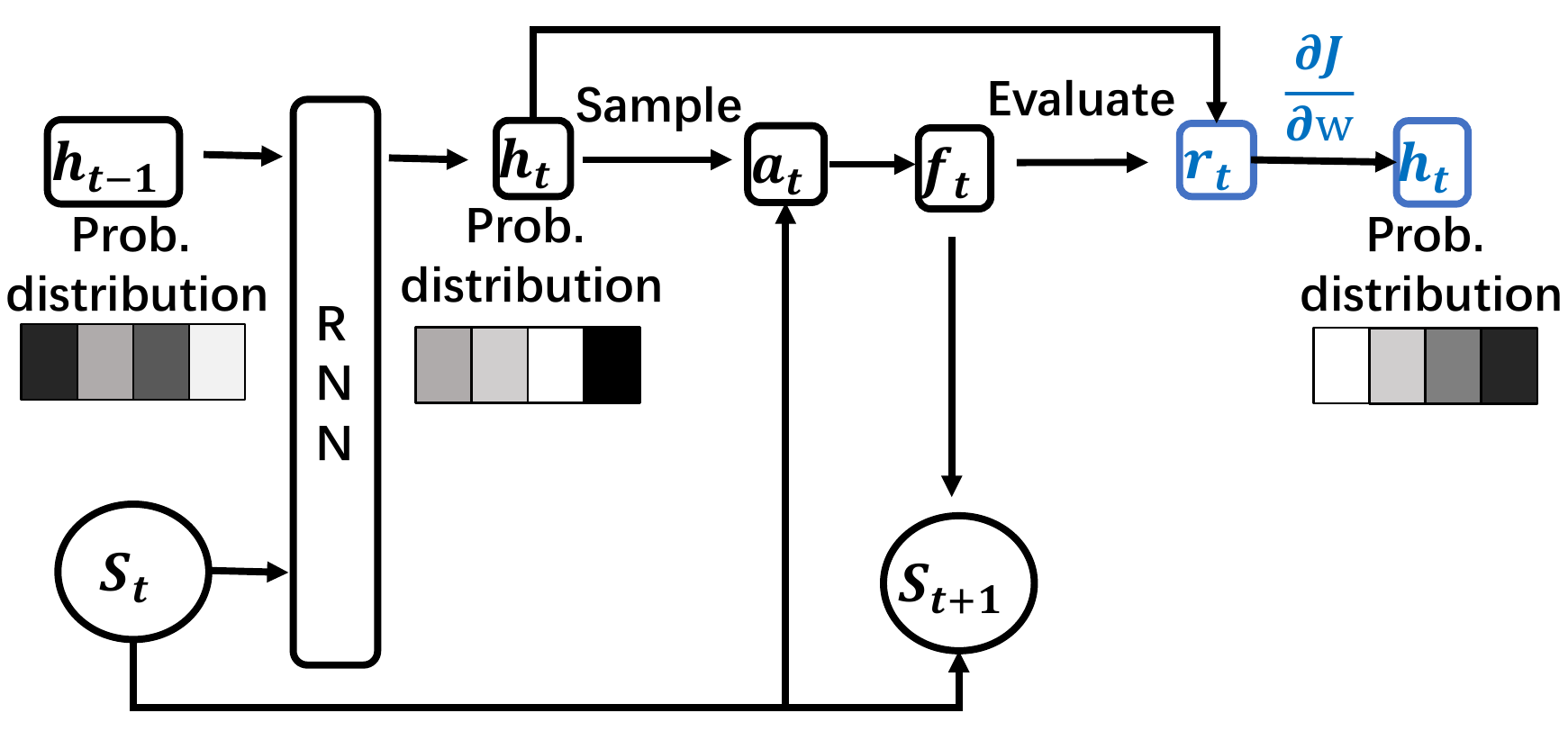}
    \vspace{-0.2cm}
    \caption{An illustration of feature generation using RNN-based agent. An RNN implements an agent of the raw feature. $s_t$ is state, $a_t$ is action, $h_t$ is action probability distribution, $f_t$ is generated features, $r_t$ is the reward of new features. The initial state is the original feature. The evaluation task returns the reward $r_t$ to the $h_t$ through backpropagation on the loss function Equ. \ref{equ:rnn_loss} and updates $h_t$.}
    % \vspace{-0.5cm}
    \vspace{-0.4cm}
    \label{fig:rnn_agent}
\end{figure}

Based on the above definition, we take one agent as example to show how the proposed RL-based \com{AFE} generates features. 
For each feature, we exploit a RNN as the agent to take actions. 
Specifically, we maintain the hidden state of RNN as the probability distribution to sample \com{an} action (feature transformation function) to perform. 
For the first round generation, we set the action probability distribution as uniform distribution, and the original feature set as the input.
For the $t$-th round generation, we take the probability distribution $h_{t-1}$ updated from the $(t-1)$-th round generation, and current state (feature set) $s_t$ of RL as the input to RNN. 
RNN would output the updated probability distribution $h_t$. 
We sample an action $a_t$ based on the updated probability distribution $h_t$. 
Then, we generate new features $f_t$ by applying the action $a_t$ on the feature set $s_t$. 
The updated state (feature set) $s_{t+1}$ is the combination of the input feature set $s_t$ and the generated feature set $f_t$. 
Moreover, the generated features $f_t$ will be evaluated on the evaluation tasks and obtain the reward $r_t$. 
The action probability distribution $h_t$ will be further updated based on the reward $r_t$ by Equ. \ref{equ:rnn_loss}. 
Finally, the updated distribution $h_t$ will be taken as the hidden state of RNN to perform the next round feature generation.

% the loss function of our RNN is .  to integrated reward to 
The loss function of our RNN is as Equ. \ref{equ:rnn_loss}. Agent loss function is constructed by three parts.  $r$ is the reward, $h$ is the action probability distribution, and $\theta$ is the weight of RNN.
\begin{equation}
\begin{split}
\label{equ:rnn_loss}
&\mathcal{L}(\theta,h,r) = \log(\mathrm{argmax}(h))*r + \log(h)*h + ||\theta||^2 \\
\end{split}
\end{equation}

\begin{table}[h]%[!t]
% \vspace{-0.2cm}
\caption{Notations for \TheName}
\vspace{-0.2cm}
\label{table:Notations}
\centering
\begin{tabular}{|l|l|}
\hline
Notations &  Meaning \\
\hline
$\mathcal{D} \langle \mathbf{F}, \mathbf{y}\rangle$  & %Original 
Dataset $\mathcal{D}$ with feature $\mathbf{F}$ and label $\mathbf{y}$ \\
$\mathcal{T}$ & Downstream task (classification or regression) \\
$A_{\mathcal{T}}(\mathbf{F},\mathbf{y})$ & Score of a downstream task $\mathcal{T}$ on $(\mathbf{F}, \mathbf{y})$ \\
$\mathbf{ \check{F} }$ & A set of generated feature  \\  
$\mathcal{D}^i$ & The $i$-th taining set \\
$\mathcal{D}^i_j$ & The $i$-th residual dataset (drawed $j$-th feature) \\
$\mathcal{V}$ & Validation set \\
$A^i_0$ & The score of $i$-th original training set  \\
$A^i_j$ & The score of residual dataset $\mathcal{D}^i_j$ \\
$d$ & \com{MinHash signature output dimension} \\
$\textbf{H}$ & Approximate hashing features  \\
$\textbf{L}$ & Labels of negative/positive feature (0 or 1) \\
$thre$ & \com{Threshold of score gain for feature labels (0 or 1)}\\
$Rec, Prec$ & Recall, Precision \\
$\mathcal{C}_{\mathcal{D}}$ & The FPE model \\
$N$ & The number of features on a target dataset \\
% $f, agt$ & The feature, agent \\
% $\theta$ & Parameter of one agent network \\
$T$ & Sample of an episode \\
$s_t$ & State at time $t$ for one agent \\
$ a_t $ & Action at time $t$ of one agent \\
$r_t$ & Reward at time $t$ of one agent \\
$\gamma$  & Discount factor in range [0,1] \\
$h_t$ & Action probability distribution at time $t$ of one agent \\
\hline
\end{tabular}
\vspace{-0.5cm}
\end{table}

The notations used in this paper are summarized in Table~\ref{table:Notations}.

\section{\TheName}
% Pengyang
In this section, we introduce our proposed framework in detail. We start with overall framework of our proposed \TheName. Then, we present FPE model for reducing sample and feature size. Moreover, we introduce the two-stage training strategy for enhancing the learning efficiency. Finally, we provide a theoretical analysis of the algorithm complexity. 
\subsection{Framework Overview}

\begin{figure*}%[!htbp]
    \centering
    \vspace{-0.4cm}
    \includegraphics[scale=0.46]{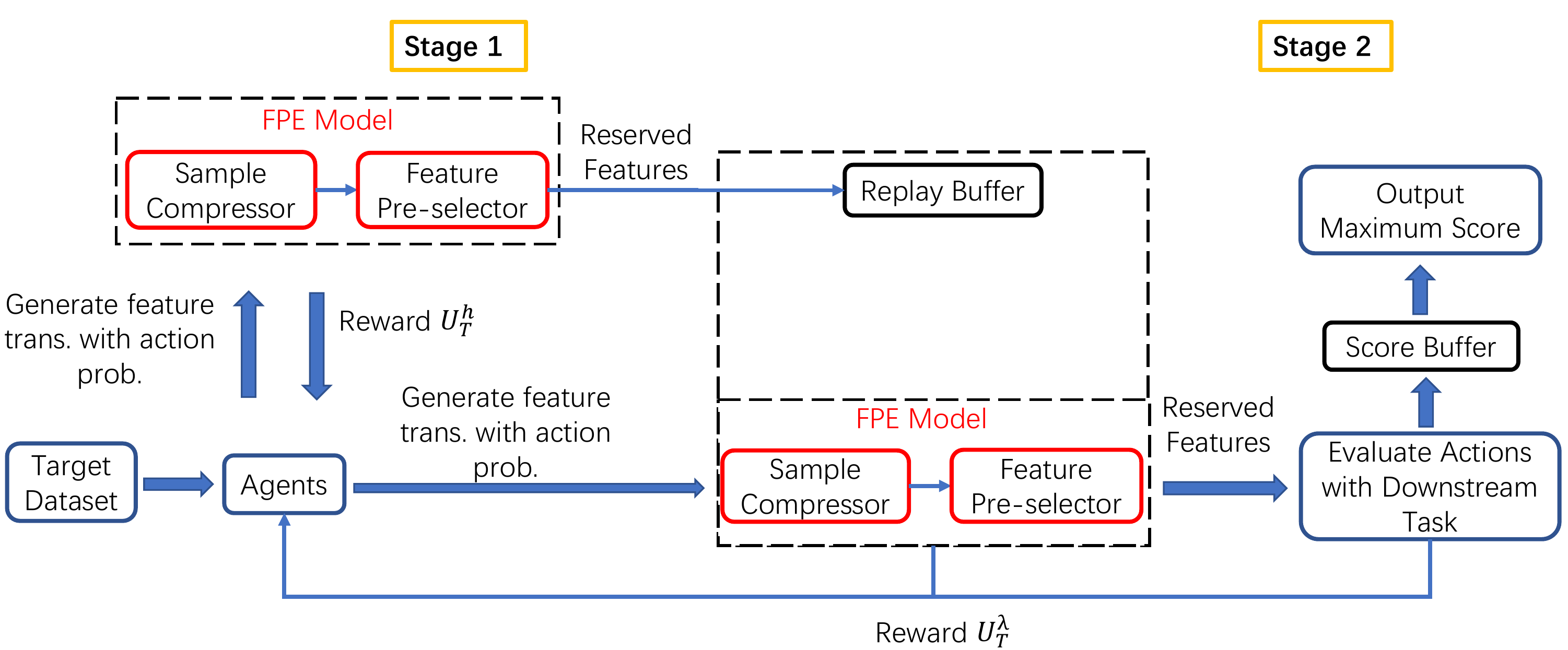}
    % \caption{Online process of ~\TheName ~ with batch agents off-policy RL.}
    \vspace{-0.2cm}
    \caption{Framework Overview.}
    \vspace{-0.5cm}
    \label{fig:overview}
\end{figure*}

% \begin{figure*}%[!htbp]
%     \centering
%     \includegraphics[scale=0.52]{images/leaning-to-hash.pdf}
%     \caption{FPE model training, validation and testing.}
%     \label{fig:hash_learning}
% \end{figure*}

Figure~\ref{fig:overview} shows an overview of the proposed framework. 
The core elements of ~\TheName~  includes two parts: 
(1) FPE model, which is to reduce the sample size and candidate feature size for improving the evaluation efficiency; 
(2) two-stage policy-training strategy, which is to boost the learning efficiency by avoiding training from scratch. 
Specifically, for each original feature, the agent will first generate features following the pipeline as shown in Figure~\ref{fig:rnn_agent}. 
Then, the generated features are fed into the FPE model to reduce sample size with the hashing module, and reduce the candidate feature size with a binary classifier pre-trained by the pre-evaluation task. 
During the policy training procedure, we will initialize the policy with taking the pre-evaluation task as the evaluation task to train for several epochs. 
Meanwhile, we construct a replay buffer to record promising actions. 
Then, we continue to train the policy with real downstream tasks. 
The proposed framework continue to be trained until the optimal features are achieved.

\subsection{Feature Pre-Evaluation (FPE) Model \label{section:FPE}}
As discussed in Section~\ref{section:introduction}, sample size and feature size are two vital factors to compromise the efficiency of AFE. 
Therefore, we develop FPE model to help reduce the sample and feature size to accelerate AFE. 
Specifically, FPE model consists of two modules: (1) sample compressor, which is to reduce sample size with hashing operations; 
and (2) feature pre-selector, which is to reduce candidate feature size for evaluation tasks with pre-trained binary classifier.

\noindent \textbf{Sample Compressor.} 
Figure~\ref{fig:sample_score_time} indicates that not all samples are necessary for feature evaluation. 
Therefore, a method for reducing the sample size is highly desired. 
Moreover, different datasets have various sizes. 
To be generalized across datasets, the compression method also expects to compress arbitrary input sizes into the fix one. 
Therefore, in this work, we adopt hashing techniques as the sample compressor to reduce sample size. 
Specifically, we take $\text{MinHash}$ as the hashing function family. 
The basic idea of $\text{MinHash}$ is to assign the target dimension hashing values, and select $d$ instances with the minimum hashing values as the compressed results~\cite{wu2020review}. 

In our sample compressor case, given a dataset (tabular data) $\mathcal{D} \in \mathbb{R}^{M \times N}$ with $N$ features (column) and $M$ samples (row). 
We take samples (row) as the target dimension, and input into $\text{MinHash}$. 
Suppose the expected sample size is $d$, we expect the compressed dataset with selected $d$ samples should also preserve the sample similarities in the original $M$-sample dataset. 
Then the compression process can be represented as 
\begin{equation}
\begin{split}
    \Tilde{\mathcal{D}} &= \text{MinHash}(\mathcal{D}, d), \\
   ~ s.t.~ |\text{sim}&(\mathcal{D}^1, \mathcal{D}^2)-\text{sim}(\Tilde{\mathcal{D}}^1, \Tilde{\mathcal{D}}^2)| < \epsilon \\
\end{split}
\end{equation}
where $\mathcal{D} \in \mathbb{R}^{M\times N}$ and $\Tilde{\mathcal{D}} \in \mathbb{R}^{d\times N}$ denote the orignal dataset and compressed dataset, respectively, and $\epsilon$ denotes a very small constant. \com{$D^1$ and $D^2$ are the two datasets whose similarity is to be calculated.}

% \com{Feature quality screening is a feature distance measurement model, that is, feature similarity measurement. MinHash can be used to estimate the similarity of two sets quickly. The MinHash algorithm uses Jaccard similarity to measure how similar objects are. In the case of a fixed number of features, MinHash maps the samples of the original data set to the sparse matrix through the element dictionary and then selects the d-dimensional signature as the output according to the Hash function, thereby reducing the number of samples \cite{wu2020review}. E-AFE uses MinHash to extract d-dimensional signatures from different numbers of redundant samples to reduce the sample size. Efficiently reuse knowledge of public datasets and pre-trained FPE models according to specified downstream tasks.}  % kafeng

\begin{figure}
    \centering
    \vspace{-0.3cm}
    \includegraphics[scale=0.18]{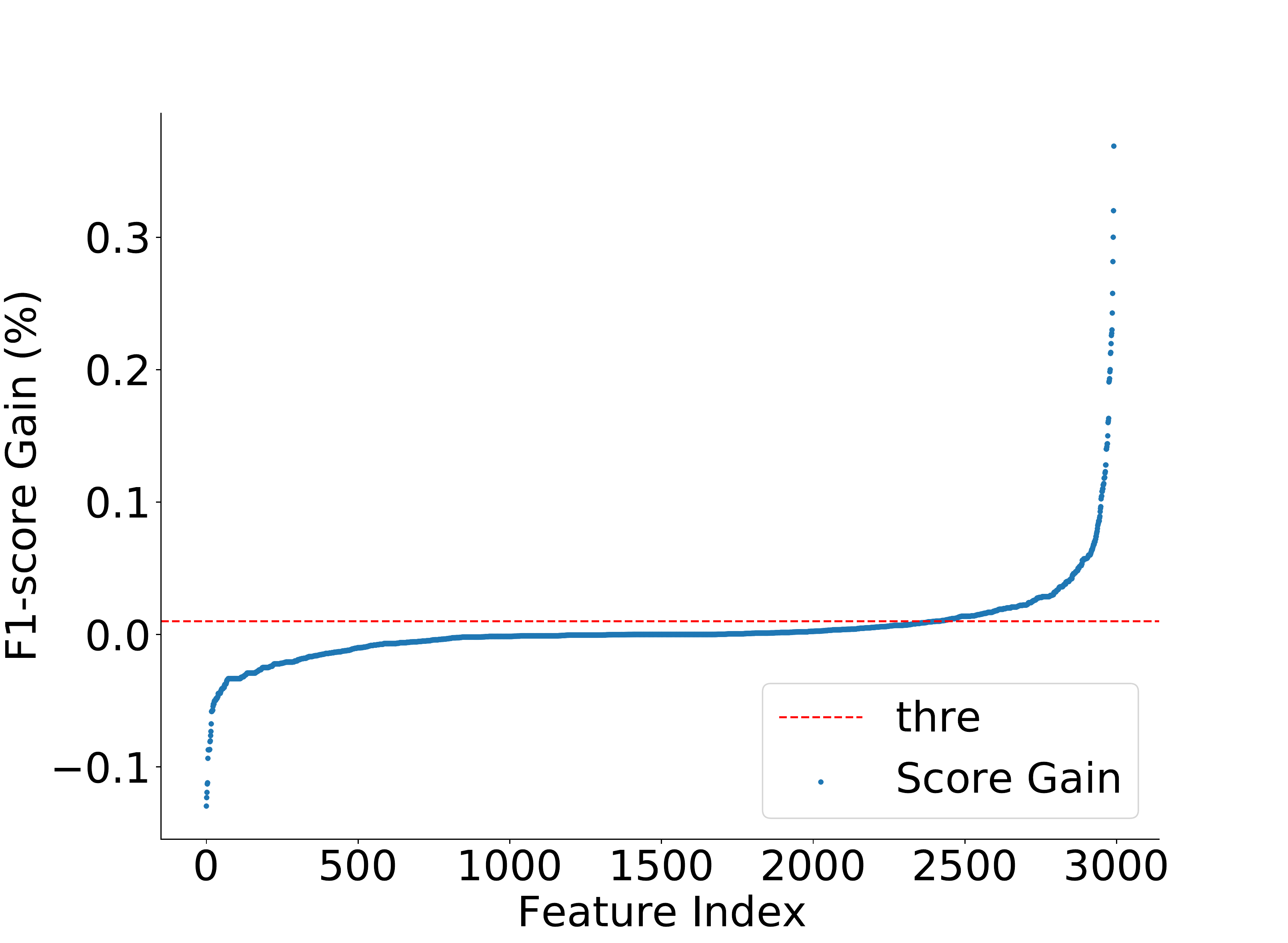}
    \vspace{-0.3cm}
    \caption{The $thre$  and score gain for classification.}
    \label{fig:thre_score_gain}
    \vspace{-0.6cm}
\end{figure}

\noindent \textbf{Feature Pre-Selector.}
Since evaluating features directly on downstream tasks is time-consuming, we develop a fast yet effective pre-evaluation task to pre-select candidate features. 
Specifically, we pre-train a binary classifier on public datasets to distinguish the effective features from the generated feature set. 
Formally, given $n$ public datasets, and each dataset is $\mathcal{D}^i=[k^i, m^i], i \in [1,n]$, which include $m^i$ original features and $k^i$ instances. 
The downstream task $\mathcal{T}$ calculates the original performance score as $A_0$. 
Then, we leave out the $j$-th feature from $i$-th original dataset to construct the residual dataset $\mathcal{D}^i_j$. 
We continue to calculate the performance score $A_j^i$ for $\mathcal{D}^i_j$.
We compare the performance score between $A_0$ and $A_j^i$ to determine whether the $j$-th feature is effective. 
Then, the labeling process of feature effectiveness can be represented as 
\begin{equation}
\begin{split}
\label{equ:acc0_label}
&\mathbf{A}_0^i=\mathcal{T}( \mathcal{D}^i), ~ i\in[1,n] \\
% &  \mathcal{D}_j^i = \mathcal{D}^i - \mathbf{F}_j^i \\
&  \mathcal{D}_j^i \leftarrow \mathcal{D}^i - \mathbf{F}_j^i \\
&\mathbf{A}_j^i=\mathcal{T}( \mathcal{D}_j^i), ~ j\in[1,m^i], i\in[1,n] \\
&\mathbf{L}_j^i = \mathrm{sgn}(\mathbf{A}_0^i - \mathbf{A}_j^i + thre)
\end{split}
\end{equation}
where $thre$ is the threshold of score gain; $\mathrm{sgn}(z) = 1$ if $z > 0$ and $\mathrm{sgn}(z) = 0$ (or equivalently -1) otherwise; $\mathbf{L}$ is the label vector. 
Consequently, the effective feature is labeled as $1$, otherwise $0$.
\com{$thre$ is set a small value, such as 0.01. This value is to set the boundary of the binary classification, which is larger than 0, so that better features can be found. If set to 0, many features that are not significantly improved for downstream tasks may be retained. The size of this threshold is also set according to the recall effect. For example, as shown in Figure \ref{fig:thre_score_gain}.}

Once the label vector is obtained, we pre-train a binary classifier $\mathcal{C}_\mathcal{D}$ paramterized by $\phi$ through taking the sample-compressed features as input. 
We select effective features (labeled as $1$) as the candidate as the input to evaluation tasks in RL-based AFE. 
Although the pre-evaluation task is still based on real downstream tasks, once well-trained, the binary classifier can infer the effectiveness of given features quickly.

\noindent \textbf{Model Optimization.}
We exploit cross-entropy as the loss function and take maximizing recall as the optimization target. 
The objective is to find the optimal sample compressor with the optimal compression size $d$ complying with best feature-effectiveness classifier. 
Therefore, we consider the sample compressor as the hyperparameter by fine-tuning the hashing function options and compression size $d$ to train the feature-effectiveness classifier. 
Formally, the predicted label vector $\Tilde{\mathbf{L}}$ can be represented as 
\begin{equation}
\label{equ:hash_label}
\begin{split}
\mathbf{H} &= \text{MinHash}(\mathcal{D}, d) \\
\Tilde{\mathbf{L}} &= \mathrm{sgn}(\mathcal{C}_\mathcal{D} (\mathbf{H})) \\
\end{split}
\end{equation} 

Then, according to the ground-truth $\mathbf{L}$ and prediction $\Tilde{\mathbf{L}}$, we calculate the precision $Prec$ and recall $Rec$ as 
\begin{equation}
\label{equ:rec_prec}
\begin{split}
     Prec &=\text{Precision}(\mathbf{L}, \Tilde{\mathbf{L}}) \\
     Rec &=\text{Recall}(\mathbf{L}, \Tilde{\mathbf{L}}) \\
\end{split}
\end{equation}
Then, the objective can be represented as 
\begin{equation}
\begin{split}
\label{equ:argmax}
&\mathcal{C}_\mathcal{D}^*, \text{MinHash}^*, d^*  \leftarrow \underset{\mathcal{C}_\mathcal{D}, \text{MinHash}, d}{\mathrm{argmax}}~ Rec, \\
&~ s.t.~ Prec > 0 ~ and~ Rec < 1, \\
\end{split}
\end{equation}
where $\mathcal{C}_\mathcal{D}^*$, $\text{MinHash}^*$ and $d^*$ are optimal classifier, and the corresponding hashing function and compressed sample size, respectively.
We perform stochastic gradient descent (SGD) to train and fine-tune the FPE model.
Algorithm~\ref{alg:offline} shows the calculation details of FPE model.

\begin{algorithm}[!tb]
% \caption{Training and Validation of Feature Evaluation Model}
\caption{Feature Pre-Evaluation (FPE) Model}
\label{alg:offline}
\textbf{Input}: $n$ public dataset $\mathcal{D}$ for training, \com{every dataset has $k^i, i \in [1,n]$ samples and $m^i, i \in [1,n]$ features}. $\mathcal{V}$ is the validation set. $\mathcal{T}$ is the downstream task. $\mathbf{d}$ is the vector of \com{MinHash signature output dimension}
% reduced sample size
for an approximate feature. \com{$thre$ is the threshold of score gain for feature labels (0 or 1)}. $\text{MinHash}$ is hash function set.   \\
\textbf{Output}: Trained FPE model $\mathcal{C}_\mathcal{D}$, MinHash function and MinHash  output dimension $d$. 

\begin{algorithmic}[1] %[1] enables line numbers
% \FOR{$d_0, d_1, ...,d_t,...$}
% \FOR{$d \in \textbf{d}$}
\WHILE{Select $d \in \textbf{d}$ or $\text{MinHash}$ Function}
\STATE Let $\mathbf{A}=[], \mathbf{L}=[], Rec=[]$.
\STATE \textcolor{blue}{/*Training model with cross-entropy*/}
\FOR{$i = 1 \rightarrow n$} % for all train dataset
\STATE Get $\mathbf{A}_0^i$ by Equ. (\ref{equ:acc0_label}) on $\mathcal{D}^i$.
\FOR{$j = 1 \rightarrow m^i$} % for every feature
\STATE Draw sub dataset $\mathcal{D}_{j}^i  \leftarrow \mathcal{D}^i - \mathbf{F}_j^i$.  
\STATE Get $\mathbf{A}_j^i$ by Equ. (\ref{equ:acc0_label}) on $\mathcal{D}_j^i$.
\IF {$\mathbf{A}_0^i - \mathbf{A}_j^i > thre $}
\STATE $\mathbf{L}_j^i \leftarrow 1$
\ELSE
\STATE $\mathbf{L}_j^i \leftarrow 0$
\ENDIF
\STATE $\mathbf{H}_j^i \leftarrow \text{MinHash}(\mathbf{F}_j^i, d)$ as Equ. (\ref{equ:hash_label})
\ENDFOR
\ENDFOR
\STATE \textcolor{blue}{/*Validating model for recall*/}
% \STATE $\mathcal{C}_{\mathcal{D}}_t \leftarrow  \mathbf{\mathcal{C}}_t(\mathbf{H},\mathbf{L})$ 
\STATE $\Tilde{\mathbf{L}} = \mathrm{sgn}(\mathbf{\mathcal{C}}_{\mathcal{D}}(\mathbf{H}))$ as Equ. (\ref{equ:hash_label})
\STATE $Rec =\text{Recall}(\mathbf{L}, \Tilde{\mathbf{L}})$ as Equ. (\ref{equ:rec_prec})
% \STATE $\mathbf{Prec}_t, \mathbf{Rec}_t \leftarrow \mathbf{\mathcal{C}}_{\mathcal{D}}_t(\text{MinHash}(\mathcal{V}))$ as Equ. (\ref{equ:rec_prec})
\ENDWHILE

% \STATE \textcolor{blue}{/*Selecting Model with Precision-Recall Space*/}
\STATE \textcolor{blue}{/*Train and fine-tune the FPE model*/}
\STATE $\mathcal{C}_{\mathcal{D}}^*, \text{MinHash}^*, d^* \leftarrow  \underset{\mathcal{C}_\mathcal{D}, \text{MinHash}, d}{\mathrm{argmax}}( Rec)$ as Equ. (\ref{equ:argmax})
\STATE \textbf{return} $\mathcal{C}_{\mathcal{D}}^*, \text{MinHash}^*, d^*$
\end{algorithmic}
\end{algorithm}
\vspace{-0.2cm}

\begin{algorithm}[!tb]
\caption{Two-Stage Policy Training Strategy}
\label{alg:online}
\textbf{Input}: 
Trained FPE model $\mathcal{C}_\mathcal{D}$. 
Features $\mathbf{F}$ of the target dataset. \text{MinHash} function. 
\text{MinHash} output dimension $d$. \\
\textbf{Output}: Maximize score of downstream task

\begin{algorithmic}[1] %[1] enables line numbers
% \STATE Let $p=0, f^h=[]$.
\STATE \textcolor{blue}{/*Stage 1: Quick Initialization with FPE Model*/}  % initialization ???
\WHILE{Choose agent one by one}
\STATE Agent sample action with equal rate
\STATE Action generates new feature $ \check{f} \in \mathbf{ \check{F} }$ 
% \STATE $f^h \leftarrow \text{MinHash}(\check{f}, d)$
% \STATE $p \leftarrow \mathcal{M}(f^h)$
\STATE Evaluate $\check{f}$ by Equ. (\ref{equ:probability})
\IF {$\check{f}$ is positive feature} % reserved
% \STATE Store transition $\{ s_t, a_t, r_t, p_t, s_{t+1} \}$ of sample to replay buffer
\STATE Store this feature to replay buffer
\ELSE
% \STATE Dropout transition $\{ s_t, a_t, r_t, p_t, s_{t+1} \}$ of sample
\STATE Dropout this feature 
\ENDIF
\STATE Get $A^h_t$ by Equ. (\ref{equ:reward_hash})
\STATE Get return $U^h_t$ by Equ. (\ref{equ:return_single}) for every agent
\ENDWHILE
\STATE Agents initialization complete 
\STATE \textcolor{blue}{/*Stage 2: Formal Training*/}
% \WHILE{Get transition of sample from replay buffer}
\WHILE{Get feature from replay buffer}
\STATE Sample feature transformation with action prob.
\STATE Generate new features by transformation
\STATE Save score of the downstream task
\STATE Get return $U^{\lambda}_t$ by Equ. (\ref{equ:return_multi}).
\ENDWHILE
\STATE Trained agents % Batch
\STATE \textbf{return} Maximize score of downstream task
\end{algorithmic}
\end{algorithm}

\subsection{Two-Stage Training Strategy} 
We integrate the well-trained FPE model into the RL-based AFE framework to construct ~\TheName~, as shown in Figure~\ref{fig:overview}. 
The conventional training process of RL-based AFE methods suffer from exhaustively exploring the action-state space, which wastes a lot of running time to reach the optimum. 
Therefore, to solve the problem, we propose the two-stage training strategy by borrowing external knowledge from pre-trained FPE model to avoid learning policy from scratch. 
The two-stage training strategy includes: (1) quick initialization with FPE model, and (2) formal training. 

\noindent \textbf{Stage 1: Quick Initialization with FPE Model} 
The benefits of FPE model lie in two aspect. 
On the one hand, FPE model is trained on various public datasets, which contain rich external knowledge of feature characteristics; 
On the other hand, the well-trained FPE can directly infer the effectiveness of the generated features, which saves substantial time comparing to the downstream task evaluation. 
Therefore, at the initialization stage, we omit the downstream task evaluation, but directly take the results from FPE model as the reward to update the policy of ~\TheName~. 
Formally, we compute the reward of each action as follows: for taking action $a_t$, the improvement $r_t$ is the score $A_t$ of state $s_t$ minus $A_{t-1}$ of state $s_{t-1}$ , which means the gain of performing $a_t$ intuitively. 
Then, we utilize the $\lambda -return$ $U^ \lambda _t$ that combines all $k-step$ returns $U_t$ as the final reward signal for action $a_t$. 
% Then, we utilize the $U_t$ that combines all steps returns $U_t$ as the final reward signal for transformation $a_t$. 
We define $A^{O}$ and $A^{h}$ as the performance score of the original dataset and compressed dataset by FPE model, respectively.
$\Delta A_{max}$ and $\Delta A_{min}$ are the maximize and minimize score gain of input space. 
Then, the accumulated reward can be computed as 
\begin{equation}
\begin{split}
\label{equ:probability}
&p = \mathcal{C}_{\mathcal{D}}(\text{MinHash}(\check{f}, d)) \\
\end{split}
\end{equation}

\begin{equation}
\label{equ:reward_hash}
A^{h}_t = \left \{
\begin{aligned}
A^{O} + \frac{0.5-p}{0.5}  (\Delta A_{max} - thre), p \in [0, 0.5) \\
A^{O} + \frac{0.5-p}{0.5}  (thre - \Delta A_{min}), p \in [0.5, 1] \\
\end{aligned}
\right. 
\end{equation}
where, $thre$ is the threshold of score gain for feature labels (0 or 1).
$\check{f}$ is a new feature from the test set.
$p$ is the output probability of the binary classifier.

\begin{equation}
\begin{split}
\label{equ:return_single}
&r^{h}_t = A^{h}_t - A^{h}_{t-1}  \\  % FPE model
% &U^{h(k)}_t = r^{h}_t + \gamma r^{h}_{t+1} + ... + \gamma ^k r^{h}_{t+k} \\  % return
&U^h_t = r^{h}_t + \gamma r^{h}_{t+1} + ... + \gamma ^k r^{h}_{t+k} = \sum ^t _ {k=0} \gamma^{(t-k)} r^h_k \\ 
% &U_t = \sum ^t _ {j=0} \gamma^{(t-j)} r^h_j \\
\end{split}
\end{equation}

\noindent \textbf{Stage 2: Formal Training}
After several epochs of training in stage 1, we then change the evaluation task to the real downstream task. We take the output of FPE model (pre-selected feature with reduced sample and feature size) as the candidate features to formally train the policy, where the reward is computed as the performance gain on the downstream tasks. 
To be consistent with the stage 1, we represent the accumulated rewards for stage 2 as:
\begin{equation}
\begin{split}
\label{equ:return_multi}
&r_t = A_t - A_{t-1}  \\  % NFS
% &U^{(k)}_t = r_t + \gamma r_{t+1} + ... + \gamma ^k r_{t+k} \\
&U_t = r_t + \gamma r_{t+1} + ... + \gamma ^k r_{t+k} = \sum ^ t _ {k=0} \gamma^{(t-k)} r_k  \\
% &U_t = \sum ^ t _ {j=0} \gamma^{(t-j)} r_j \\
% &U^{\lambda}_t = (1- \lambda ) \sum^{N \times T}_{k=1} \lambda ^{k-1} U^{(k)}_t \\
&U^{\lambda}_t = (1- \lambda ) \sum^{N \times T}_{k=1} \lambda ^{k-1} U_t, \\ 
\end{split}
\end{equation}
where $A_t$ and $A_{t-1}$ are evaluation scores of two successive feature sets on the downstream tasks, respectively. 

\noindent \textbf{Policy Learning.} 
To explore the optimal actions (feature transformations), we train agents to maximize their expected return, represented by $J(\theta ^{[1]}, ..., \theta ^{[j]},  \theta ^{[N]})$, where the $j$-th agent is parameterized by $\theta ^{[j]}$. 
Then, the expected return can be represented as 
\begin{equation}
\begin{split}
\label{equ:target_policy}
&J(\theta ^{[1]}, ..., \theta ^{[N]}) = \mathbb{E}_{P(a_{1:N \times T}; ~ \theta ^{[1]}, ..., \theta ^{[N]})} \Bigg[ \sum^{N \times T}_{t=1} U^{\lambda}_t  \Bigg] \\ 
\end{split}
\end{equation}
To achieve an empirical approximation, we utilize the REINFORCE \cite{williams1992simple} rule and Monte Carlo simulation \cite{robert2010monte} to update the parameters:   %  kafeng this is  Behavior Policy

\begin{equation}
\begin{split}
\label{equ:target}
&\nabla _{\theta ^{[1]}, ..., \theta ^{[N]}} J(\theta ^{[1]}, ..., \theta ^{[N]}) \approx  \\ 
&\frac{1}{m} \sum ^{m}_{k=1} \sum ^{N \times T}_{t=1} \nabla _{\theta ^{[1]}, ..., \theta ^{[N]}} \log \mathbb{P}(a_t|a_{1,t-1}; ~ \theta ^{[1]}, ..., \theta ^{[N]}) U^{ \lambda}_{t,[k]}    \\ 
\end{split}
\end{equation}
where $m$ is the number of batch size that the agent samples per epoch and $U^{ \lambda}_{t,[k]}$ is the cross-validation score that the $k-th$ sample achieves.
Algorithm~\ref{alg:online} shows the calculation details of the two-stage training strategy.

\begin{table*}
\centering
\vspace{-0.3cm}
\caption{Comparison results on 36 target datasets. C$\backslash$R is the downstream type. C is classification. R is regression. $\mathrm{RTDL_{N}}$ ($\mathrm{DL_{N}}$) is ResNet with RF from RTDL \cite{gorishniy2021revisiting}. $\mathrm{AutoFS_{R}}$ ($\mathrm{FS_{R}}$) \cite{fan2020autofs,fan2021interactive} is feature selection from randomly generated features. $\mathrm{NFS}$ \cite{chen2019neural} is feature generation and evaluation. 
\com{The FE\textbar{}DL method is to put the features selected by feature engineering into the deep learning process.
The DL\textbar{}FE method puts the original features into the deep learning training and then puts the output features into the feature engineering method for feature selection.}
$\TheName _{D}$ is the ablation study of ~\TheName~, replacing the approximate hashing model with a random dropout method. $\TheName _{R}$ is the ablation study of ~\TheName~, replacing the RL framework with policy gradient method like $\mathrm{NFS}$ \cite{chen2019neural} used. \com{$\TheName ^{P}$ is $\TheName$ with PCWS \cite{wu2017consistent}. $\TheName ^{I}$ is $\TheName$ with ICWS \cite{ioffe2010improved}. $\TheName ^{L}$ is $\TheName$ with LICWS \cite{li20150-bit}.
$\TheName$ use CCWS \cite{wu2016canonical}}
}
\label{tab:comparsion_score}
\vspace{-0.2cm}
\begin{adjustbox}{scale=0.95}
\com{
\begin{tabular}{l|l|l|l|l|l|l|l|l|l|c|l|l|l} 
\hline
Dataset           & \begin{tabular}[c]{@{}l@{}}C$\backslash$\\R\end{tabular} & \begin{tabular}[c]{@{}l@{}}Samples$\backslash$\\Features\end{tabular} & $\mathrm{FS_{R}}$ & $\mathrm{DL_{N}}$ & NFS   & FE\textbar{}DL & DL\textbar{}FE & $\TheName _R$ & $\TheName _D$ & $\TheName ^L$ & $\TheName ^P$ & $\TheName ^I$ &   \TheName              \\ 
\hline
Higgs Boson       & C                                                        & 50000$\backslash$28                                                   & 0.723             & 0.756             & 0.731 & 0.779          & 0.836          & 0.725         & 0.802         & 0.821         & 0.833         & 0.816         & \textbf{0.836}  \\ 
\hline
A. Employee       & C                                                        & 32769$\backslash$9                                                    & 0.948             & 0.484             & 0.950 & 0.677          & 0.532          & 0.943         & 0.950         & 0.950         & 0.951         & 0.950         & \textbf{0.951}  \\ 
\hline
PimaIndian        & C                                                        & 768$\backslash$8                                                      & 0.779             & 0.743             & 0.790 & 0.787          & 0.797          & 0.783         & 0.792         & 0.793         & 0.797         & 0.795         & \textbf{0.798}  \\ 
\hline
SpectF            & C                                                        & 267$\backslash$44                                                     & 0.871             & 0.639             & 0.876 & 0.760          & 0.684          & 0.862         & 0.900         & 0.903         & 0.900         & 0.892         & \textbf{0.903}  \\ 
\hline
SVMGuide3         & C                                                        & 1243$\backslash$21                                                    & 0.842             & 0.728             & 0.846 & 0.741          & 0.793          & 0.843         & 0.875         & 0.872         & 0.879         & 0.881         & \textbf{0.881}  \\ 
\hline
German Credit     & C                                                        & 1001$\backslash$24                                                    & 0.775             & 0.681             & 0.780 & 0.692          & 0.712          & 0.773         & 0.810         & 0.796         & 0.816         & 0.813         & \textbf{0.816}  \\ 
\hline
Bikeshare DC      & R                                                        & 10886$\backslash$11                                                   & 0.978             & 0.945             & 0.990 & 0.973          & 0.993          & 0.983         & 0.992         & 0.991         & 0.993         & 0.993         & \textbf{0.993}  \\ 
\hline
Housing Boston    & R                                                        & 506$\backslash$13                                                     & 0.709             & 0.648             & 0.710 & 0.670          & 0.720          & 0.693         & 0.798         & 0.813         & 0.819         & 0.817         & \textbf{0.821}  \\ 
\hline
Airfoil           & R                                                        & 1503$\backslash$5                                                     & 0.784             & 0.697             & 0.796 & 0.723          & 0.741          & 0.781         & 0.790         & 0.795         & 0.810         & 0.807         & \textbf{0.810}  \\ 
\hline
AP. ovary         & C                                                        & 275$\backslash$10936                                                  & 0.852             & 0.659             & 0.864 & 0.684          & 0.671          & 0.858         & 0.879         & 0.880         & 0.876         & 0.879         & \textbf{0.884}  \\ 
\hline
Lymphography      & C                                                        & 148$\backslash$18                                                     & 0.895             & 0.000             & 0.922 & 0.000          & 0.000          & 0.921         & 0.960         & 0.958         & 0.960         & 0.961         & \textbf{0.964}  \\ 
\hline
Ionosphere        & C                                                        & 351$\backslash$34                                                     & 0.934             & 0.883             & 0.957 & 0.906          & 0.913          & 0.944         & 0.973         & 0.977         & 0.964         & 0.970         & \textbf{0.977}  \\ 
\hline
Openml 618        & R                                                        & 1000$\backslash$50                                                    & 0.619             & 0.072             & 0.640 & 0.678          & 0.143          & 0.635         & 0.728         & 0.729         & 0.731         & 0.737         & \textbf{0.737}  \\ 
\hline
Openml 589        & R                                                        & 1000$\backslash$25                                                    & 0.737             & 0.578             & 0.754 & 0.765          & 0.647          & 0.748         & 0.757         & 0.754         & 0.762         & 0.762         & \textbf{0.764}  \\ 
\hline
Openml 616        & R                                                        & 500$\backslash$50                                                     & 0.609             & 0.000             & 0.673 & 0.510          & 0.000          & 0.653         & 0.675         & 0.676         & 0.689         & 0.680         & \textbf{0.689}  \\ 
\hline
Openml 607        & R                                                        & 1000$\backslash$50                                                    & 0.634             & 0.025             & 0.688 & 0.603          & 0.027          & 0.678         & 0.728         & 0.729         & 0.727         & 0.733         & \textbf{0.734}  \\ 
\hline
Openml 620        & R                                                        & 1000$\backslash$25                                                    & 0.705             & 0.047             & 0.732 & 0.741          & 0.078          & 0.727         & 0.736         & 0.734         & 0.747         & 0.741         & \textbf{0.748}  \\ 
\hline
Openml 637        & R                                                        & 500$\backslash$50                                                     & 0.608             & 0.016             & 0.634 & 0.539          & 0.039          & 0.633         & 0.624         & 0.635         & 0.642         & 0.636         & \textbf{0.646}  \\ 
\hline
Openml 586        & R                                                        & 1000$\backslash$25                                                    & 0.749             & 0.521             & 0.780 & 0.612          & 0.542          & 0.780         & 0.790         & 0.793         & 0.781         & 0.784         & \textbf{0.793}  \\ 
\hline
Credit Default    & C                                                        & 30000$\backslash$25                                                   & 0.782             & 0.678             & 0.815 & 0.738          & 0.693          & 0.812         & 0.819         & 0.821         & 0.821         & 0.816         & \textbf{0.822}  \\ 
\hline
Messidor features & C                                                        & 1150$\backslash$19                                                    & 0.743             & 0.731             & 0.757 & 0.765          & 0.793          & 0.756         & 0.781         & 0.783         & 0.790         & 0.784         & \textbf{0.793}  \\ 
\hline
Wine Q. Red       & C                                                        & 999$\backslash$12                                                     & 0.671             & 0.387             & 0.692 & 0.514          & 0.421          & 0.691         & 0.716         & 0.717         & 0.720         & 0.723         & \textbf{0.723}  \\ 
\hline
Wine Q. White     & C                                                        & 4900$\backslash$12                                                    & 0.652             & 0.371             & 0.687 & 0.403          & 0.469          & 0.671         & 0.708         & 0.708         & 0.705         & 0.708         & \textbf{0.708}  \\ 
\hline
SpamBase          & C                                                        & 4601$\backslash$57                                                    & 0.934             & 0.939             & 0.938 & 0.949          & 0.949          & 0.935         & 0.949         & 0.949         & 0.949         & 0.944         & \textbf{0.949}  \\ 
\hline
AP. lung          & C                                                        & 203$\backslash$10936                                                  & 0.962             & 0.944             & 0.966 & 0.955          & 0.985          & 0.964         & 0.983         & 0.985         & 0.979         & 0.985         & \textbf{0.985}  \\ 
\hline
credit-a          & C                                                        & 690$\backslash$6                                                      & 0.782             & 0.798             & 0.793 & 0.802          & 0.815          & 0.789         & 0.810         & 0.814         & 0.893         & 0.815         & \textbf{0.815}  \\ 
\hline
diabetes          & C                                                        & 768$\backslash$8                                                      & 0.778             & 0.695             & 0.784 & 0.714          & 0.748          & 0.782         & 0.798         & 0.798         & 0.781         & 0.784         & \textbf{0.798}  \\ 
\hline
fertility         & C                                                        & 100$\backslash$9                                                      & 0.896             & 0.474             & 0.900 & 0.513          & 0.486          & 0.900         & 0.916         & 0.911         & 0.919         & 0.913         & \textbf{0.920}  \\ 
\hline
gisette           & C                                                        & 2100$\backslash$5000                                                  & 0.951             & 0.950             & 0.952 & 0.960          & 0.978          & 0.950         & 0.971         & 0.973         & 0.969         & 0.978         & \textbf{0.978}  \\ 
\hline
hepatitis         & C                                                        & 155$\backslash$6                                                      & 0.877             & 0.162             & 0.884 & 0.253          & 0.195          & 0.873         & 0.910         & 0.894         & 0.910         & 0.887         & \textbf{0.910}  \\ 
\hline
labor             & C                                                        & 57$\backslash$8                                                       & 0.876             & 0.862             & 0.930 & 0.899          & 0.964          & 0.930         & 0.964         & 0.957         & 0.960         & 0.960         & \textbf{0.964}  \\ 
\hline
lymph             & C                                                        & 138$\backslash$10936                                                  & 0.961             & 1.000             & 0.964 & 1.000          & 1.000          & 0.964         & 0.992         & 1.000         & 0.990         & 0.994         & \textbf{1.000}  \\ 
\hline
madelon           & C                                                        & 780$\backslash$500                                                    & 0.742             & 0.564             & 0.751 & 0.681          & 0.638          & 0.740         & 0.860         & 0.860         & 0.864         & 0.864         & \textbf{0.867}  \\ 
\hline
megawatt1         & C                                                        & 253$\backslash$37                                                     & 0.899             & 0.620             & 0.913 & 0.743          & 0.693          & 0.904         & 0.943         & 0.941         & 0.937         & 0.942         & \textbf{0.945}  \\ 
\hline
secom             & C                                                        & 470$\backslash$590                                                    & 0.922             & 0.030             & 0.929 & 0.092          & 0.092          & 0.927         & 0.930         & 0.929         & 0.932         & 0.932         & \textbf{0.932}  \\ 
\hline
sonar             & C                                                        & 208$\backslash$60                                                     & 0.740             & 0.699             & 0.770 & 0.769          & 0.840          & 0.768         & 0.840         & 0.842         & 0.833         & 0.837         & \textbf{0.842}  \\
\hline
\end{tabular}
}
\end{adjustbox}
\vspace{-0.3cm}
\end{table*}

\subsection{Theoretical Analysis} 
We first analyze the complexity of the proposed FPE model on reducing sample and feature sizes. 
The training complexity of FPE is related to the size of search space and datasets. 
Suppose there are $h$ hash functions in search space, the $\alpha$ candidate options of the sample size $d$. 
In the training process, given $n$ public datasets $\mathcal{D}^i=[k^i, m^i], i \in [1,n]$, and $v$ validation set $\mathcal{V}^{\gamma}=[j^{\gamma}, t^{\gamma}], \gamma \in [1,v]$, 
the training complexity of FPE model \com{in stage 1} is 
$O((h \cdot \alpha \cdot d) \cdot (n \cdot k \cdot m) \cdot (v \cdot j \cdot t))$. 

When we integrate FPE model in RL-based AFE, the FPE model serves as a sample compressor and feature pre-selector via quick inference. 
With the two-stage training strategy, ~\TheName~ evaluates the generated feature with the binary classifier first and then with the downstream tasks. 
Formally, given $N$ agents (original features) with $k$ samples, each agent performs $T$ times of feature transformations to get $m$ generated features. 
The complexity in the first training stage is $O(d \cdot k \cdot m \cdot N \cdot T)$. 
After initialization from stage 1, we continue to train the policy with \com{cross-validation} downstream tasks. 
Suppose the downstream task of RF cross-validation complexity is $O(c)$, the policy update $epoch$, the dropout rate is $ratio$, and the complexity of the stage 2 is 
% $O(c \cdot N \cdot T \cdot ratio)$
\com{$O(c \cdot N \cdot T \cdot epoch \cdot ratio)$}.

Then, the \com{two-stage} policy training complexity of ~\TheName~ is $O((h \cdot \alpha \cdot d) \cdot (n \cdot k \cdot m) \cdot (v \cdot j \cdot t))+O(c \cdot N \cdot T \cdot epoch \cdot ratio)$. 
Stage 1 runs the inference process of the FPE model, which is far less than the cross-validation time of RF and can be excluded.
If you consider deploying to multiple target datasets, the FPE model can be reused, and the training time of the FPE model is much shorter than the deployment time. Therefore
$O(c \cdot N \cdot T \cdot epoch \cdot ratio) >> O((h \cdot \alpha \cdot d) \cdot (n \cdot k \cdot m) \cdot (v \cdot j \cdot t))$, the finally complexity is 
% $O(c \cdot N \cdot T \cdot ratio)$
\com{$O(c \cdot N \cdot T \cdot epoch \cdot ratio)$}
. 
Compared to the state-of-the-art AFE method NFS, its complexity is 
% $O(c \cdot N \cdot T \cdot epoch)$
\com{$O(c \cdot N \cdot T \cdot epoch \cdot ratio)$}.  Our method drop rate is more than $0.5$.
Our algorithm guarantees 2x faster than NFS when running the same epoch without early stopping.
As the drop rate increases, our algorithm is faster. The drop rate increase positively correlates with the FPE model's ability to recall good new features on the current dataset.

\section{Results and Discussion}
% Particularly, our experiments aim to answer the following research questions:
In this section, we conduct extensive experiments to answer the following research questions: %\\
\noindent \textbf{Q1}: How is the performance of our ~\TheName~ in online AFE tasks as compared to state-of-the-art methods? %\\
\noindent \textbf{Q2}: How is the performance of ~\TheName~ variants with different combinations of key components in the RL framework? %\\ 
\noindent \textbf{Q3}: How is the performance of ~\TheName~ with a different RL framework? %\\
\noindent \textbf{Q4}: Is deep learning better than feature engineering for the tabular dataset? %\\
\noindent \textbf{Q5}: How do the key hyperparameter settings impact ~\TheName~’s performance? %\\
\com{
\noindent \textbf{Q6}: Why MinHash is chosen? %\\
\noindent \textbf{Q7}: Are the results robust to other downstream tasks? %\\
\noindent \textbf{Q8}: Is the performance improvement robust? %\\
\noindent \textbf{Q9}: How the method is performing with increasing numbers of features and larger datasets? %\\
}
In the following subsections, we first present the experimental settings and then answer the above research questions in turn.

\subsection{Experimental Settings}
\subsubsection{Data Description} 
We collect 239 public datasets for pre-training FPE, and 36 datasets for downstream task evaluations. 
Specifically, the collected public datasets are from  OpenML 
\footnote{\url{http://www.openml.org}} 
with 141 classification datasets and 98 regression datasets. 
And the datasets for downstream tasks include 26 classification datsets and 10 regression datasets. 
Detials about the datasets can be found in Table~\ref{tab:comparsion_score}.

\subsubsection{Evaluation Protocols and Metrics} 
The following metrics are used for evaluating our proposed method.
$TP, TN, FP, FN$ are true positive, true negative, false positive and false negative for all classes.
% \textbf{Accuracy} is given by $\frac{TP+TN}{TP+TN+FP+FN}$.
\textbf{Precision} is given by $\frac{TP}{TP+FP}$. 
\textbf{Recall} is given by $\frac{TP}{TP+FN}$. 
\textbf{F1-score} is the harmonic mean of precision and recall, given by $\frac{2 \times Precision \times Recall}{Precision + Recall}$.
\textbf{1-relative absolute error (1-rae)} is given by $1-rae=1-\frac{\sum |\check{y} - y|}{\sum |\bar{y} - y|}$, where $y$ is the actual target, $\bar{y}$ is the mean of $y$, and $\check{y}$ is prediction results by model. 
We use F1-score for the classification problem, and use 1-rae for the regression problem.

\subsubsection{Baseline Methods}
% There are three baseline methods: NFS \cite{chen2019neural}, $\mathrm{AutoFS_{R}}$ \cite{fan2020autofs} and $\mathrm{RTDL_{N}}$ \cite{gorishniy2021revisiting}. 
We compare the performance of our method (namely ~\TheName~) against the following baseline algorithms.

\textbf{(1)NFS.} Neural Feature Search (NFS) \cite{chen2019neural} is the most accurate method at present. It uses RF as the downstream task. For a fair comparison, we use RF \com{in} other comparison methods.

\textbf{(2)$RTDL_{N}$.} 
RTDL \cite{gorishniy2021revisiting} concludes that ResNet-like architecture is effective for tabular deep learning.
Our ResNet \cite{he2016deep} method \com{is} derived from RTDL. 
\com{First, we} divide each target dataset into train, validation, and test sets for RTDL.
After training and validating the ResNet in the framework of RTDL, we change the downstream task of ResNet, softmax, into RF, then test the modified ResNet model.

\textbf{(3)$AutoFS_{R}$.} The RL framework of AutoFS \cite{fan2020autofs,fan2021interactive} can’t consider feature generation.
\com{So we} generated features randomly and \com{selected} features by AutoFS.

\subsubsection{Reproducibility and Parameter Settings}
% We summarized the parameter settings of ~\TheName~ in our experiments in Table \ref{tab:parameters}. 
We implemented our RL framework based on TensorFlow and chose Adam \cite{kingma2014adam} as our optimizer to learn the model parameters. The learning rate is 0.01. The batch size is 32. We use four unary operations, such as logarithm, min-max-normalization, square root, and reciprocal, and five binary operations, such as addition, subtraction, multiplication, division, and modulo operation. 
Our default 
% reduced sample size
\com{MinHash signature output dimension}
is 48, and the MinHash function is CCWS. The maximum order is 5. Threshold $thre$ is 0.01. The training epoch of the two-stage policy training strategy is 200, respectively.

\subsection{Performance Comparison (\textbf{Q1})}
We evaluated the performance of all compared algorithms on 26 classification and 10 regression datasets and reported the evaluation results in Table \ref{tab:comparsion_score}, \ref{tab:feature_evaluated}, and Figure \ref{fig:learning_curve}.
To reduce the feature space, ~\TheName~ first conducts feature selection of less than maximum features according to the feature importance via RF on the 36 raw target datasets. 
Then, In the learning curve, we sample score results when the training epoch is 0, 10, 30, 60, 90, 120, 150, or 200.
From the evaluation results, we summarize several key observations as follows:

We can observe that the learning speed of ~\TheName~ is more than 2x that of NFS \cite{chen2019neural} when the learning curve is saturated.
The evaluated features of ~\TheName~ are less than $50\%$ of other methods.
In the \com{two-stage} training strategy, by dropping some bad generated features, ~\TheName~ can significantly improve the learning speed of AFE.
Comparing time with the same score, ~\TheName~ is 10x faster than NFS in some datasets.
\com{$\TheName ^{L}$, $\TheName ^{P}$ and $\TheName ^{I}$ are all variants of ~\TheName~, just using different MinHash functions.}

$\mathrm{AutoFS_{R}}$ \cite{fan2020autofs} has not had enough generated features for feature selection, and the final score is less than ~\TheName~. 
AutoFS \cite{fan2020autofs,fan2021interactive} with random feature generation does not fully mine the knowledge of feature generation. The randomly generated feature set does not have enough good features to give AutoFS for feature selection. 

The score of $\mathrm{RTDL_{N}}$ \cite{gorishniy2021revisiting} is the lowest. The ResNet feature extractor is not suitable for tabular datasets at any time. We believe that the convolution kernel of the CNN network is specially designed for data types such as images, and it is not ideal for tabular data processing.

\subsection{Model Ablation Study of ~\TheName~ (\textbf{Q2})}
\begin{figure*}[!t]
    \centering
    
    \subfloat[Higgs Boson]{\includegraphics[width=0.16\textwidth, height=0.10\textheight]{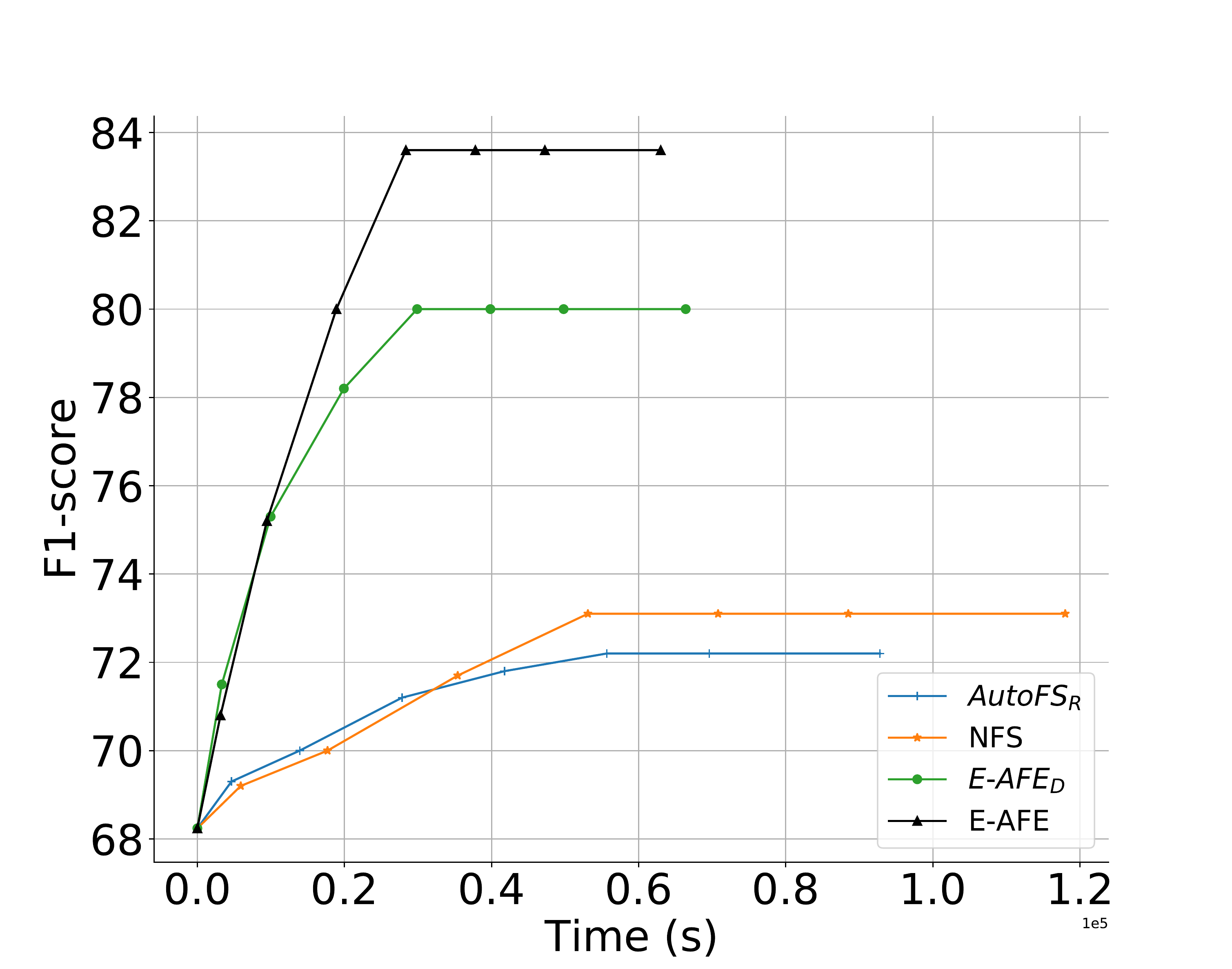}} \
    \subfloat[A. Employee]{\includegraphics[width=0.16\textwidth, height=0.10\textheight]{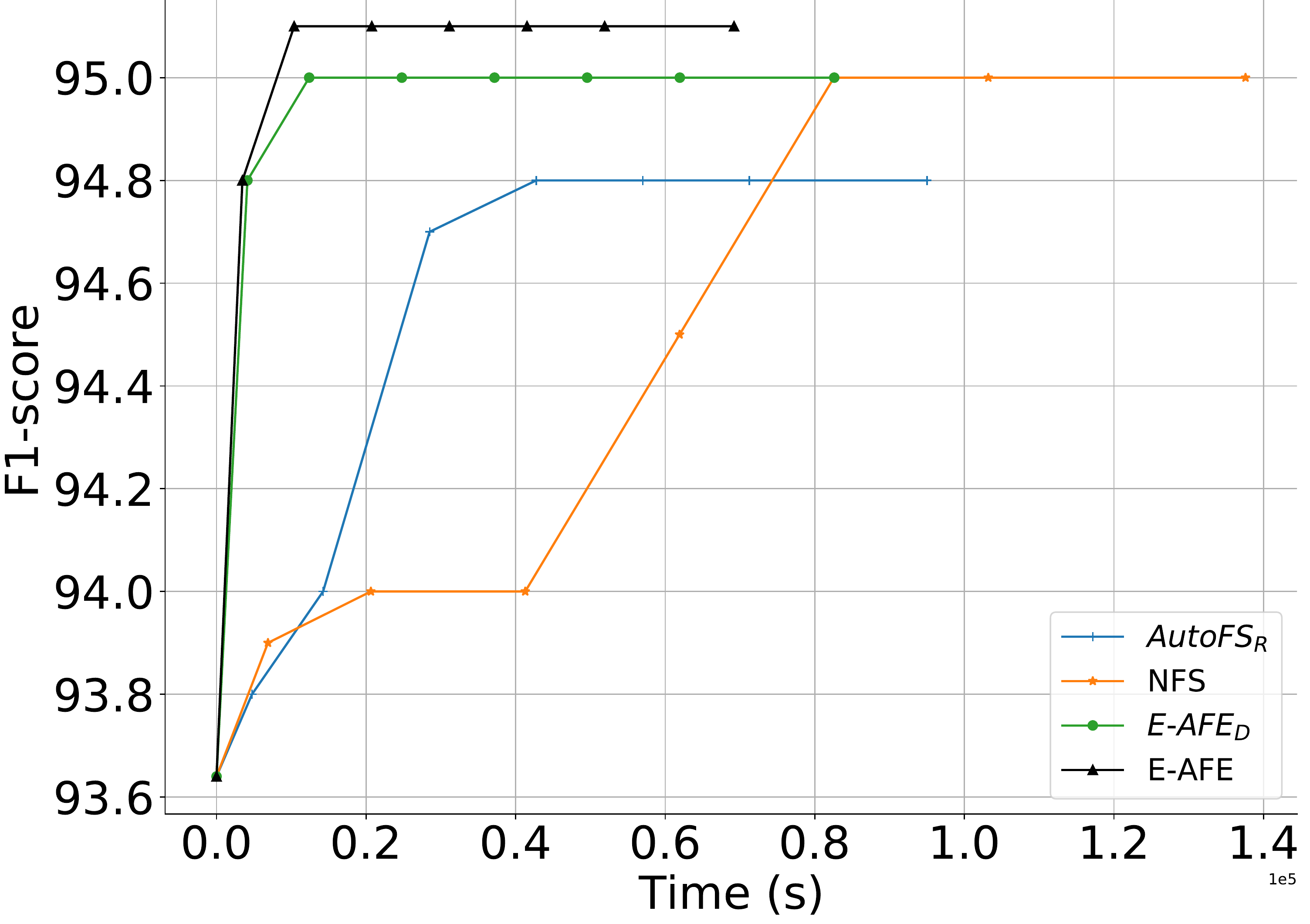}} \
    \subfloat[PimaIndian]{\includegraphics[width=0.16\textwidth, height=0.10\textheight]{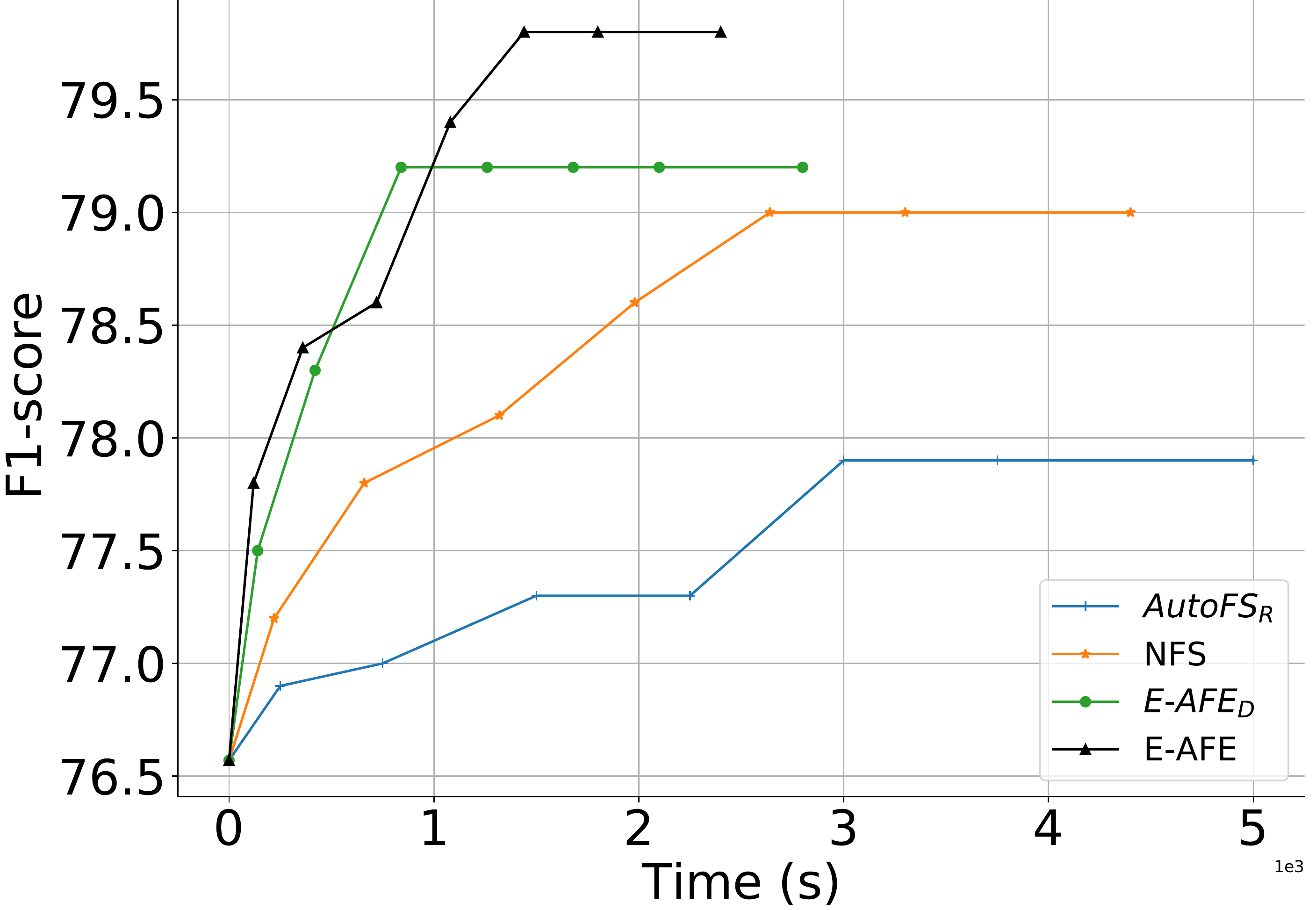}} \
    \subfloat[SpectF]{\includegraphics[width=0.16\textwidth, height=0.10\textheight]{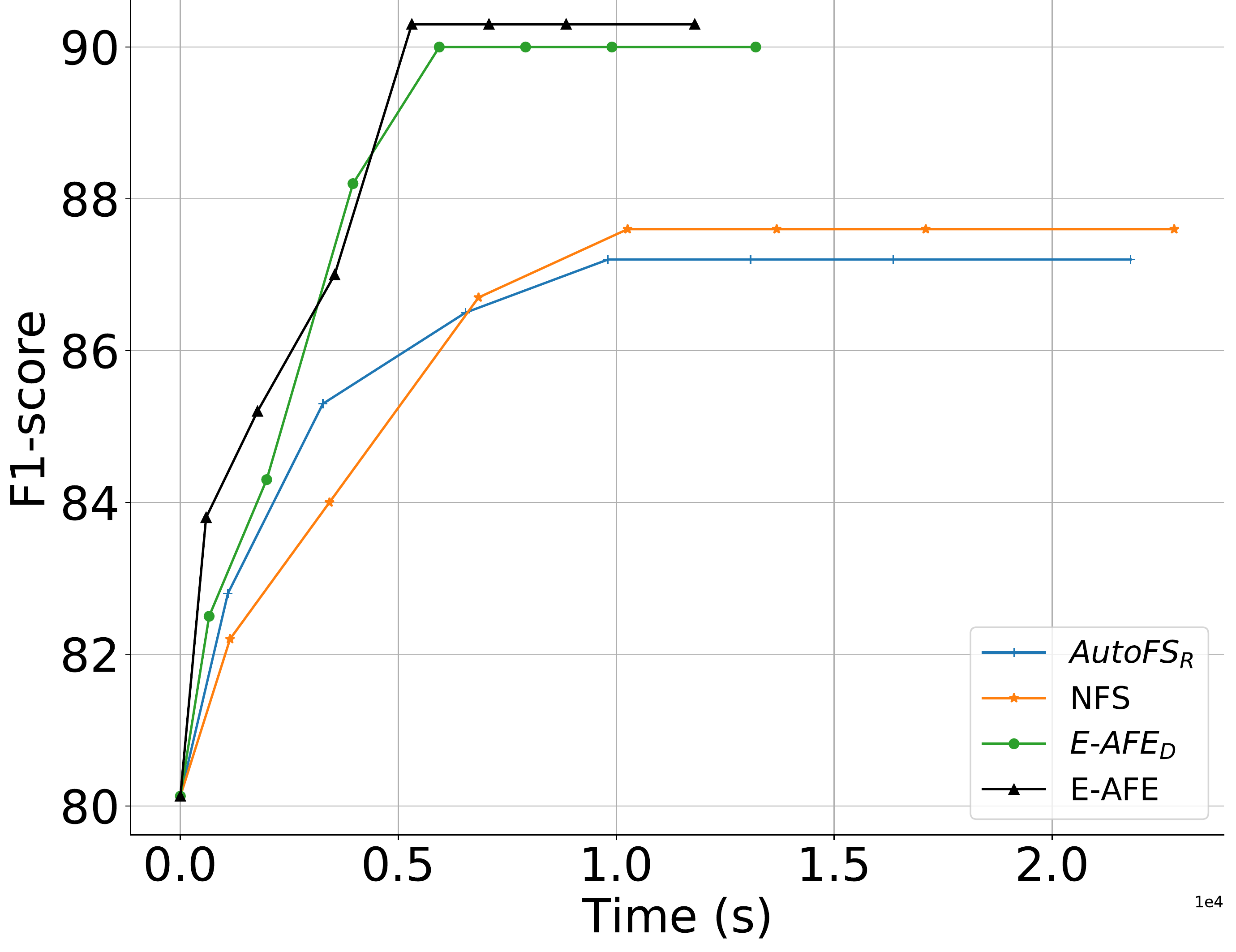}} \
    \subfloat[SVMGuide3]{\includegraphics[width=0.16\textwidth, height=0.10\textheight]{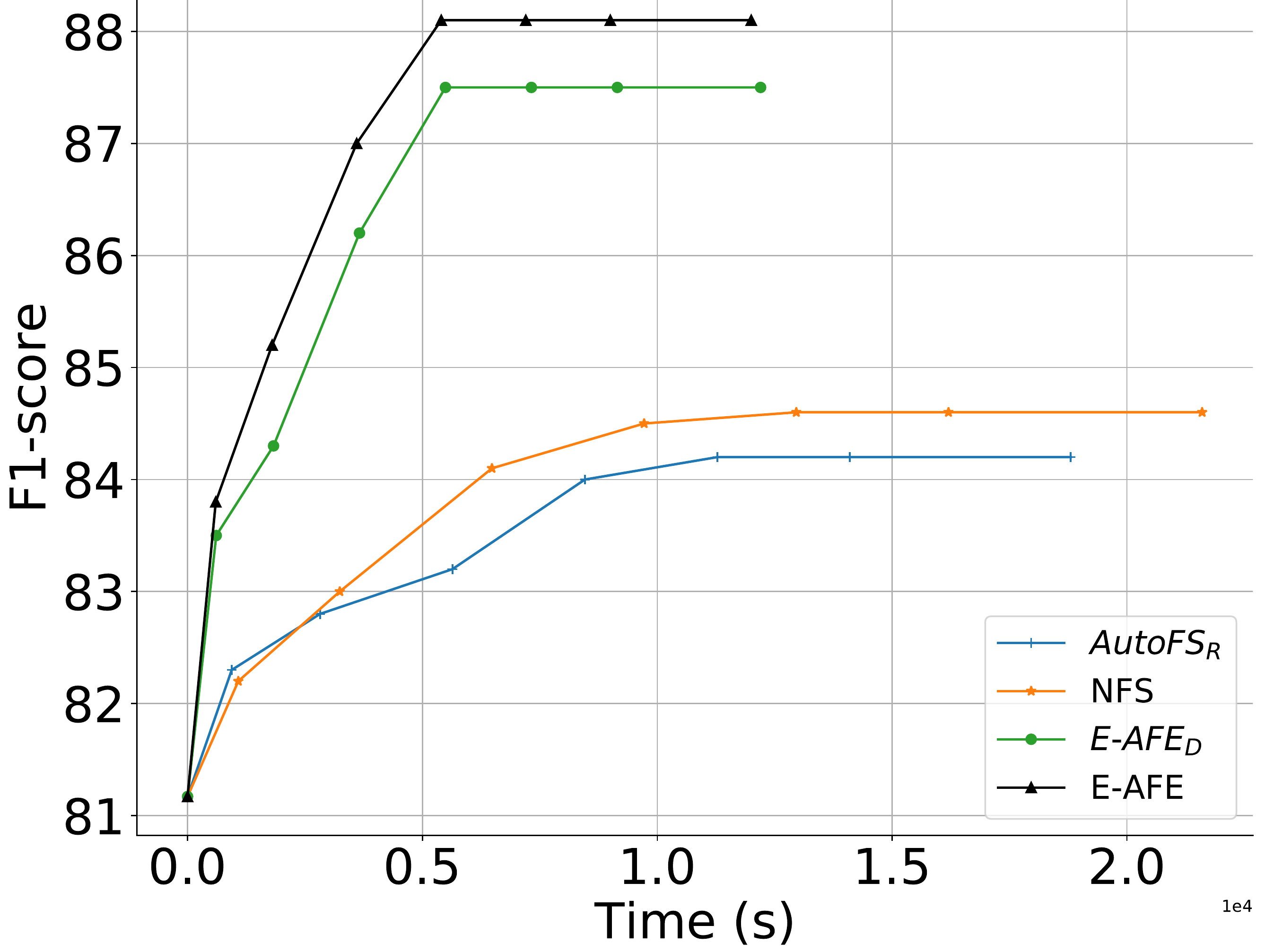}} \
    \subfloat[German Credit]{\includegraphics[width=0.16\textwidth, height=0.10\textheight]{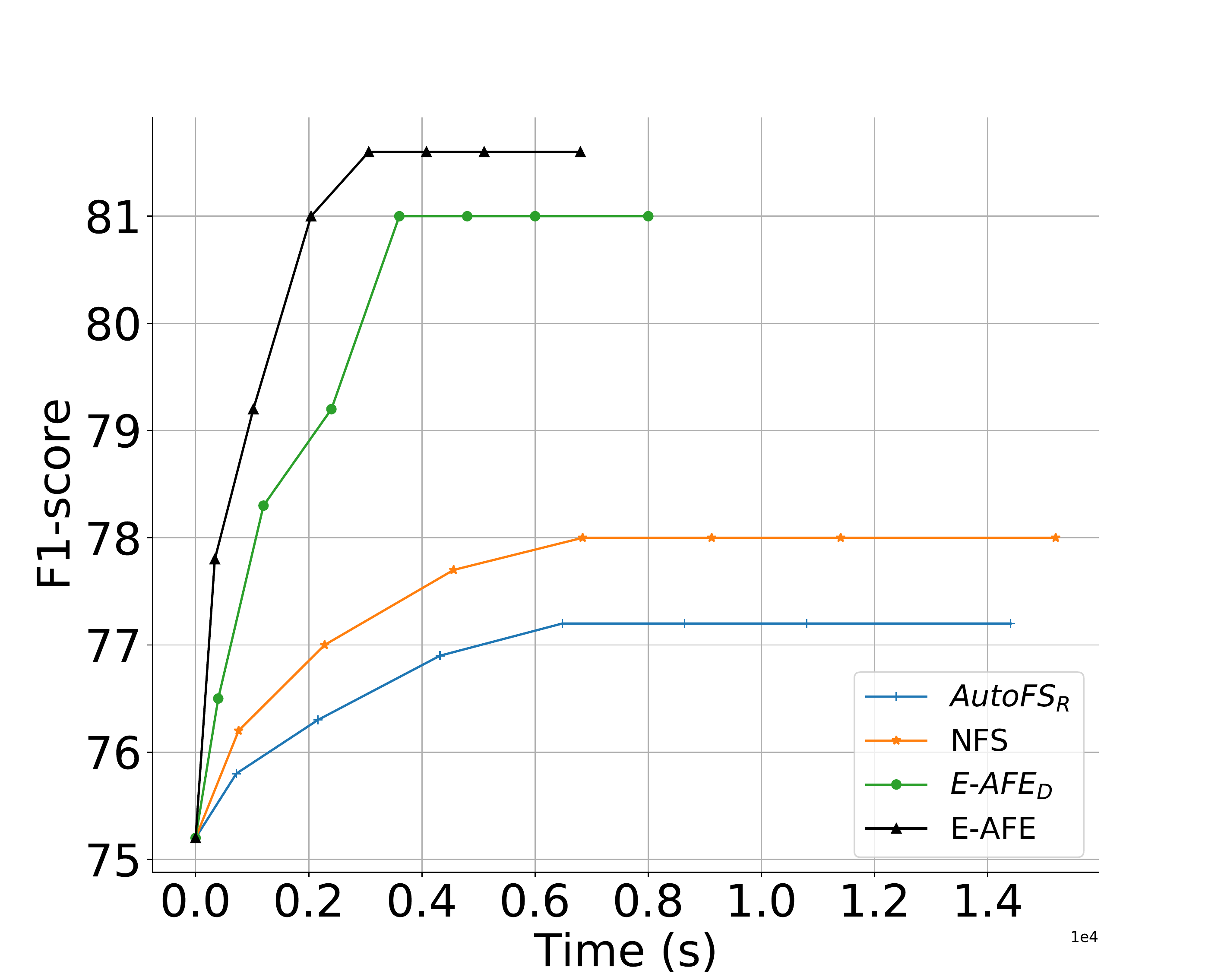}} \
    
    \subfloat[AP. ovary]{\includegraphics[width=0.16\textwidth, height=0.10\textheight]{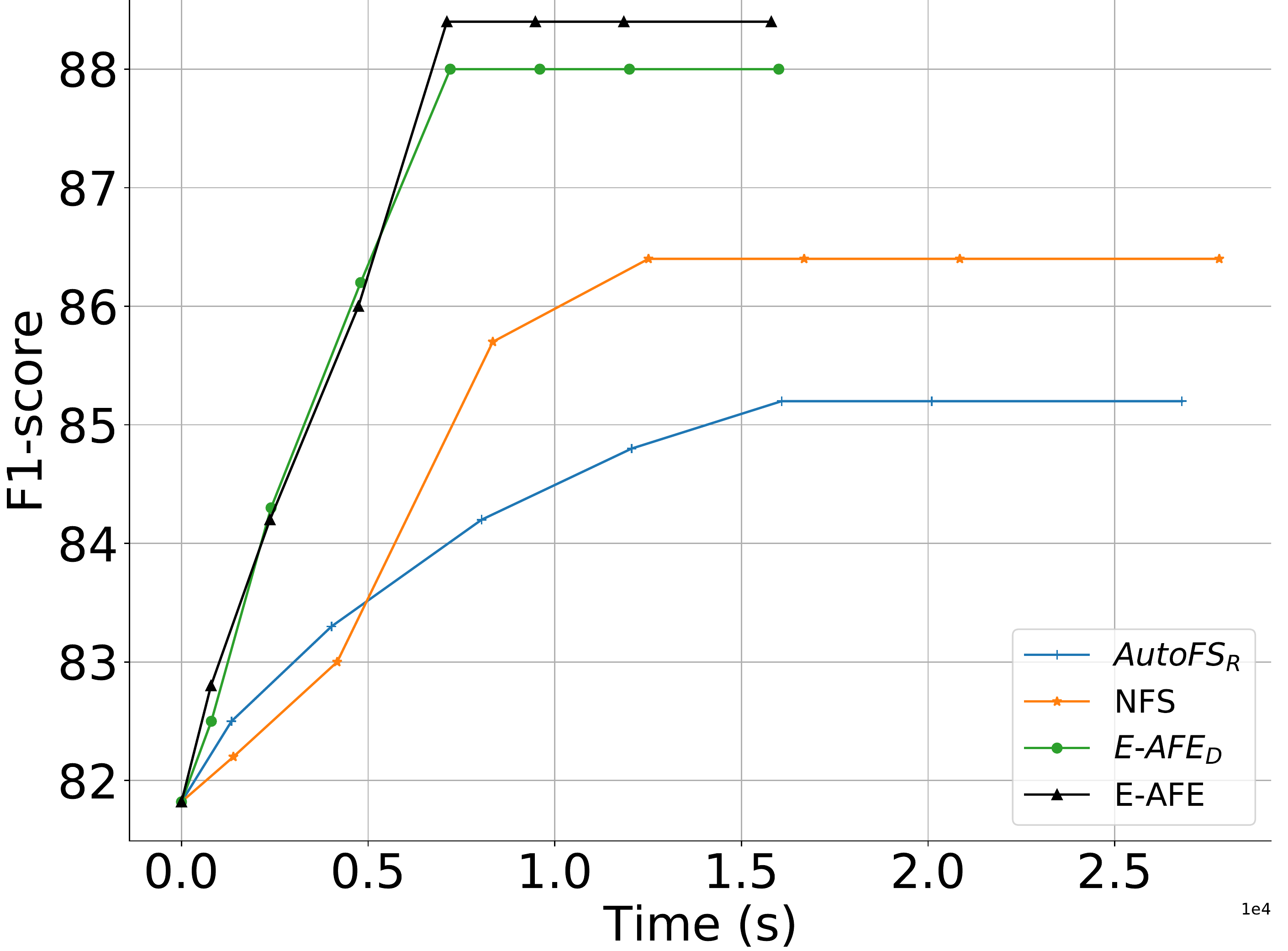}} \
    \subfloat[Lymphography]{\includegraphics[width=0.16\textwidth, height=0.10\textheight]{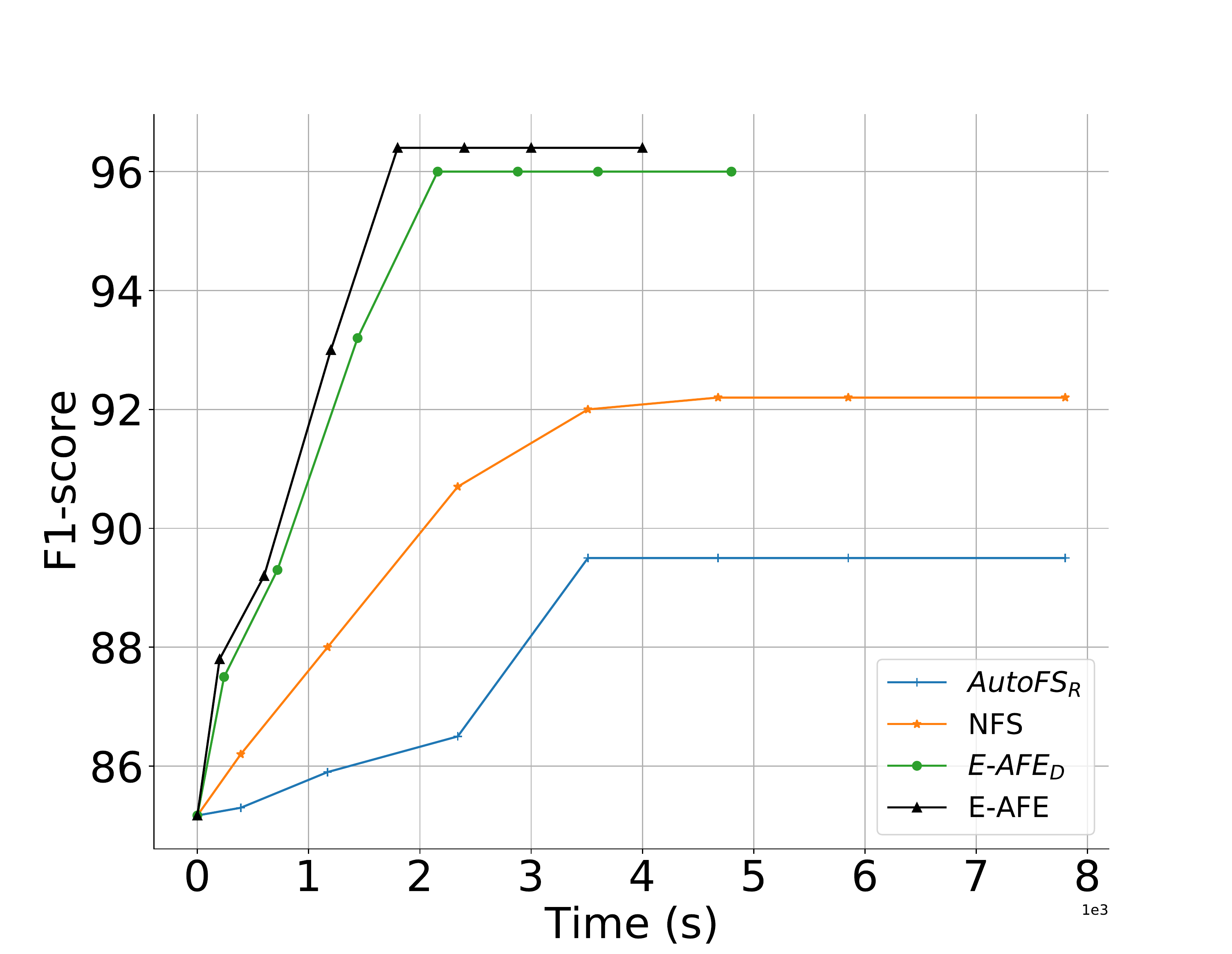}} \
    \subfloat[Ionoshpere]{\includegraphics[width=0.16\textwidth, height=0.10\textheight]{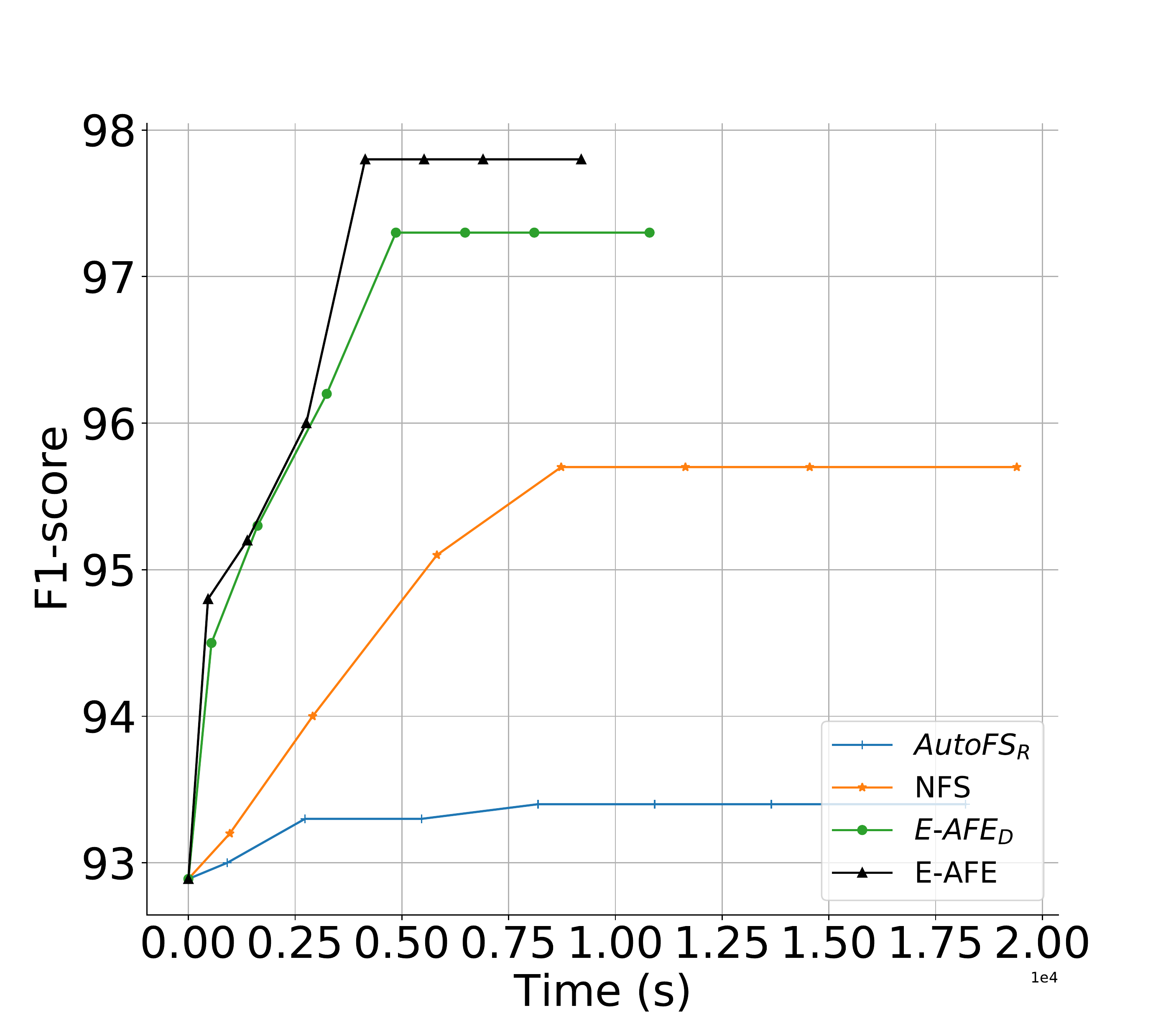}} \
    \subfloat[Credit Default]{\includegraphics[width=0.16\textwidth, height=0.10\textheight]{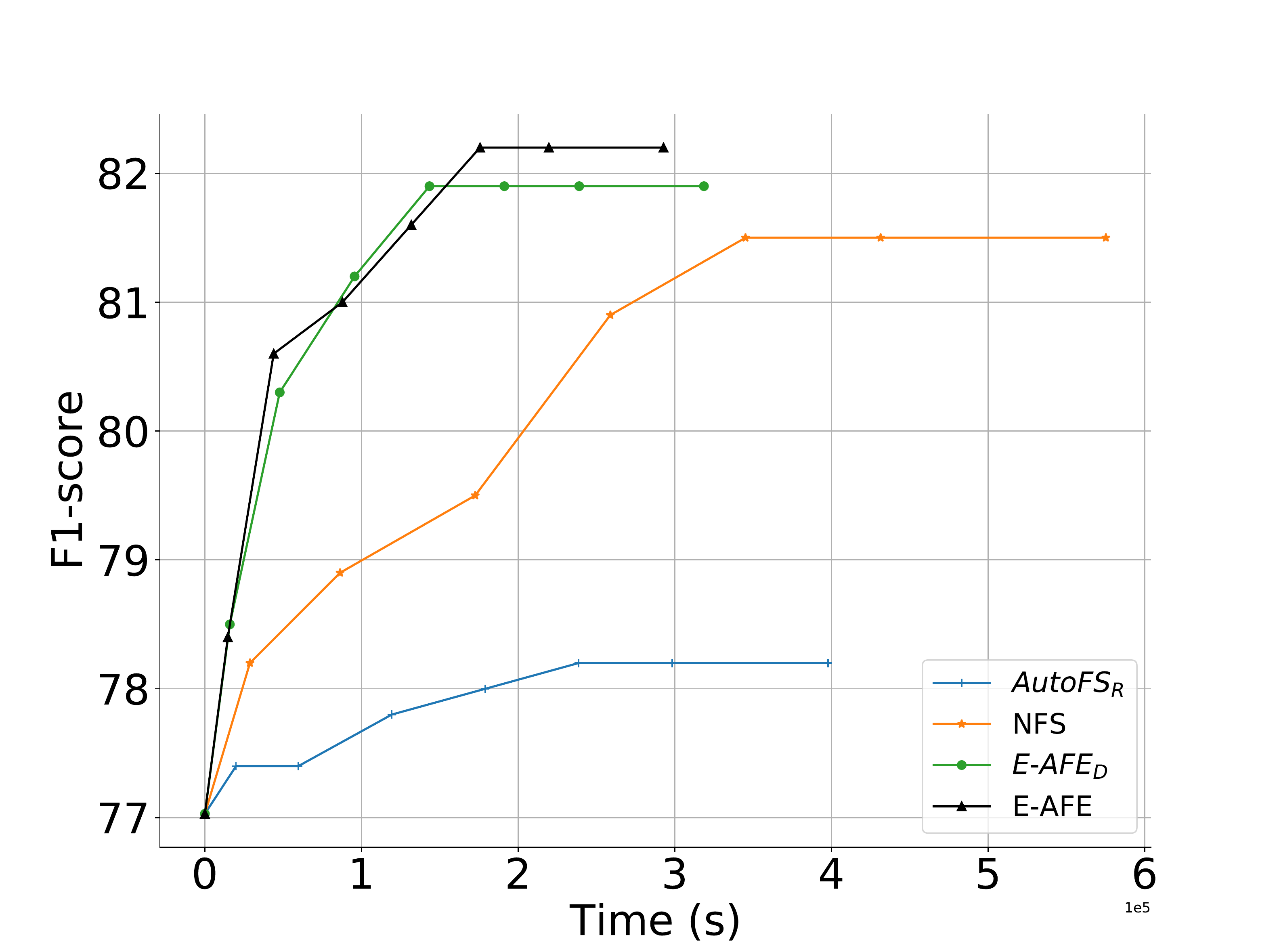}} \
    \subfloat[Messidor features]{\includegraphics[width=0.16\textwidth, height=0.10\textheight]{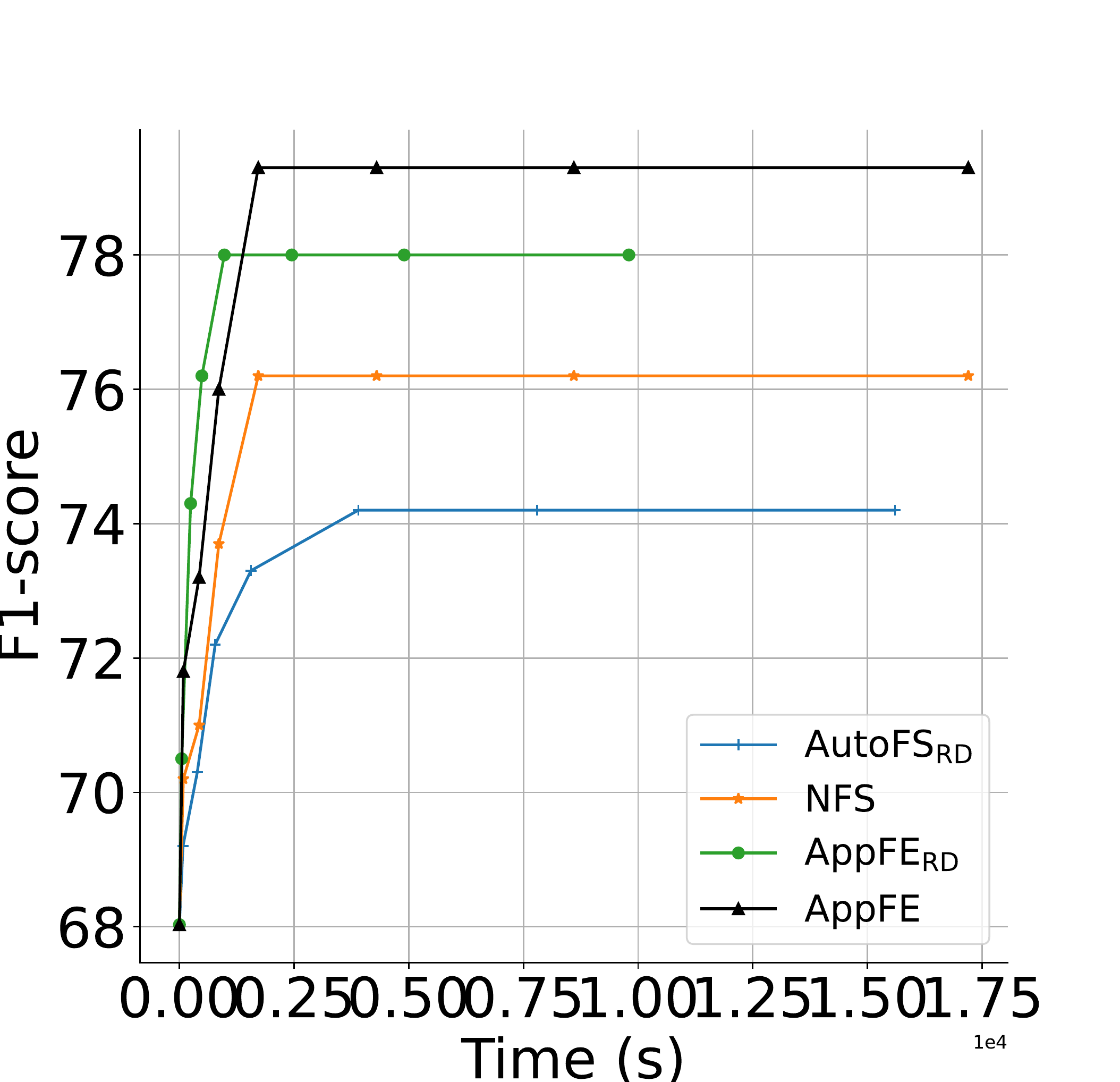}} \
    \subfloat[Wine Q. Red]{\includegraphics[width=0.16\textwidth, height=0.10\textheight]{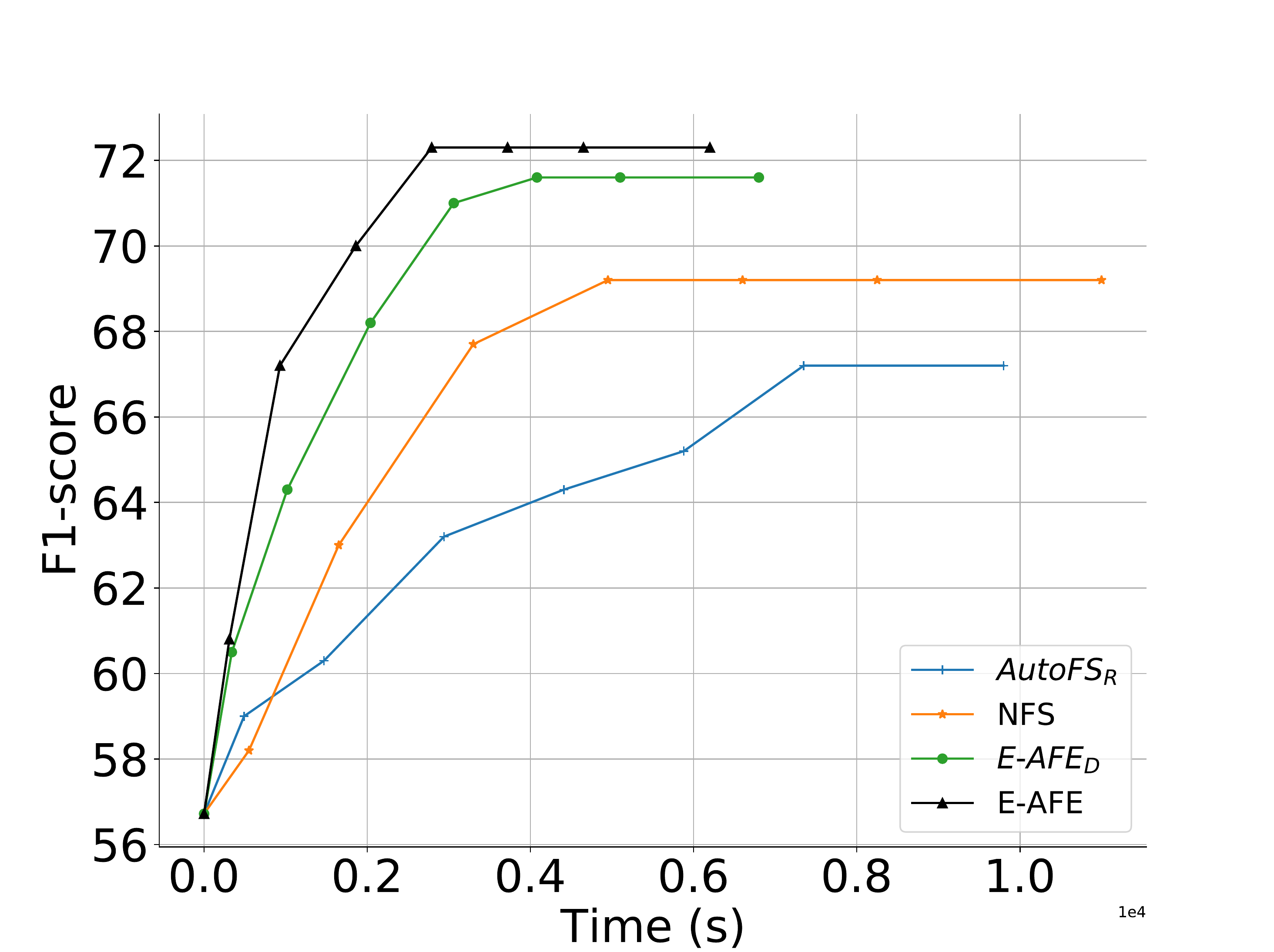}} \
    
    \subfloat[Wine Q. White]{\includegraphics[width=0.16\textwidth, height=0.10\textheight]{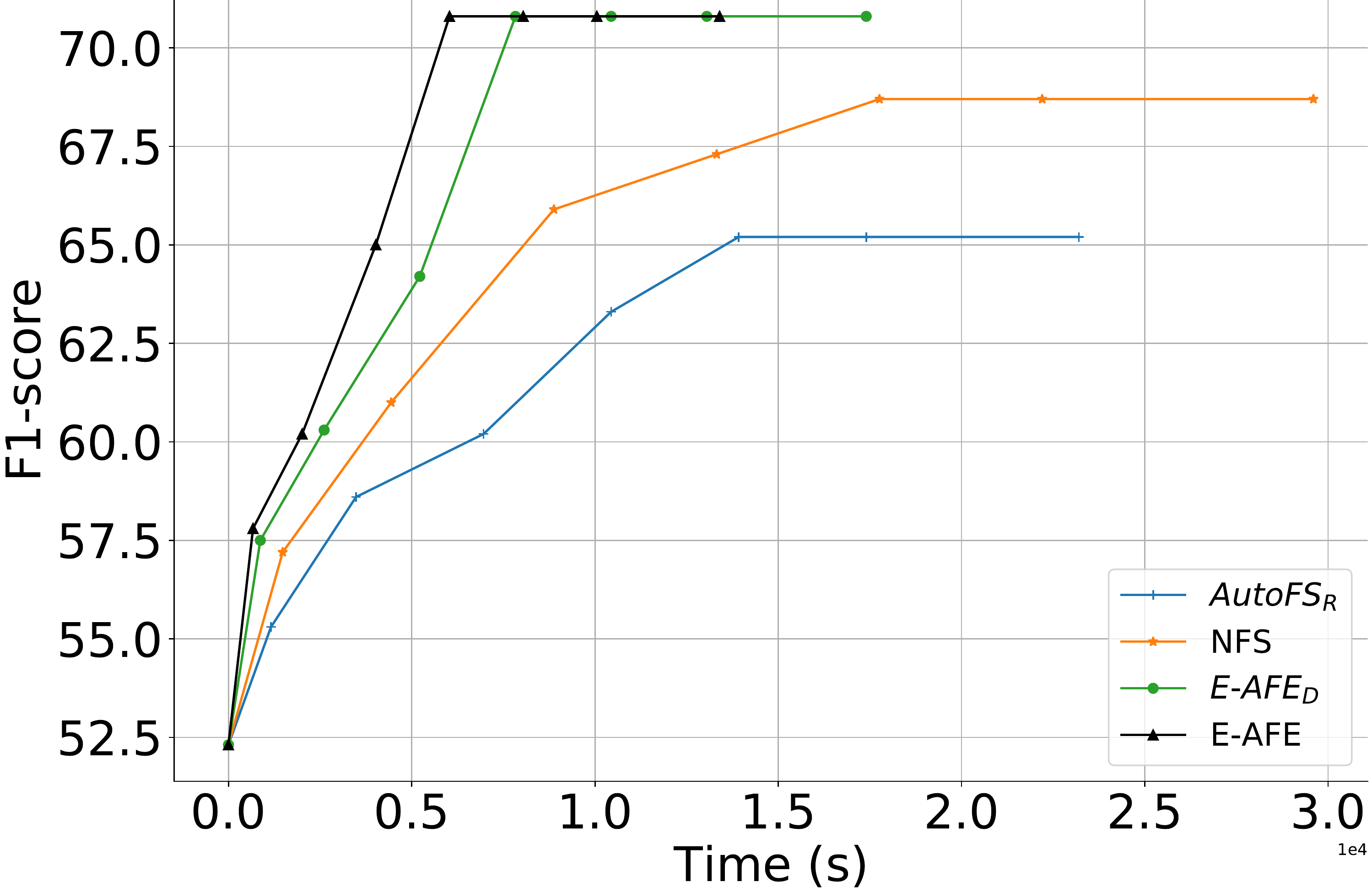}} \
    \subfloat[SpamBase]{\includegraphics[width=0.16\textwidth, height=0.10\textheight]{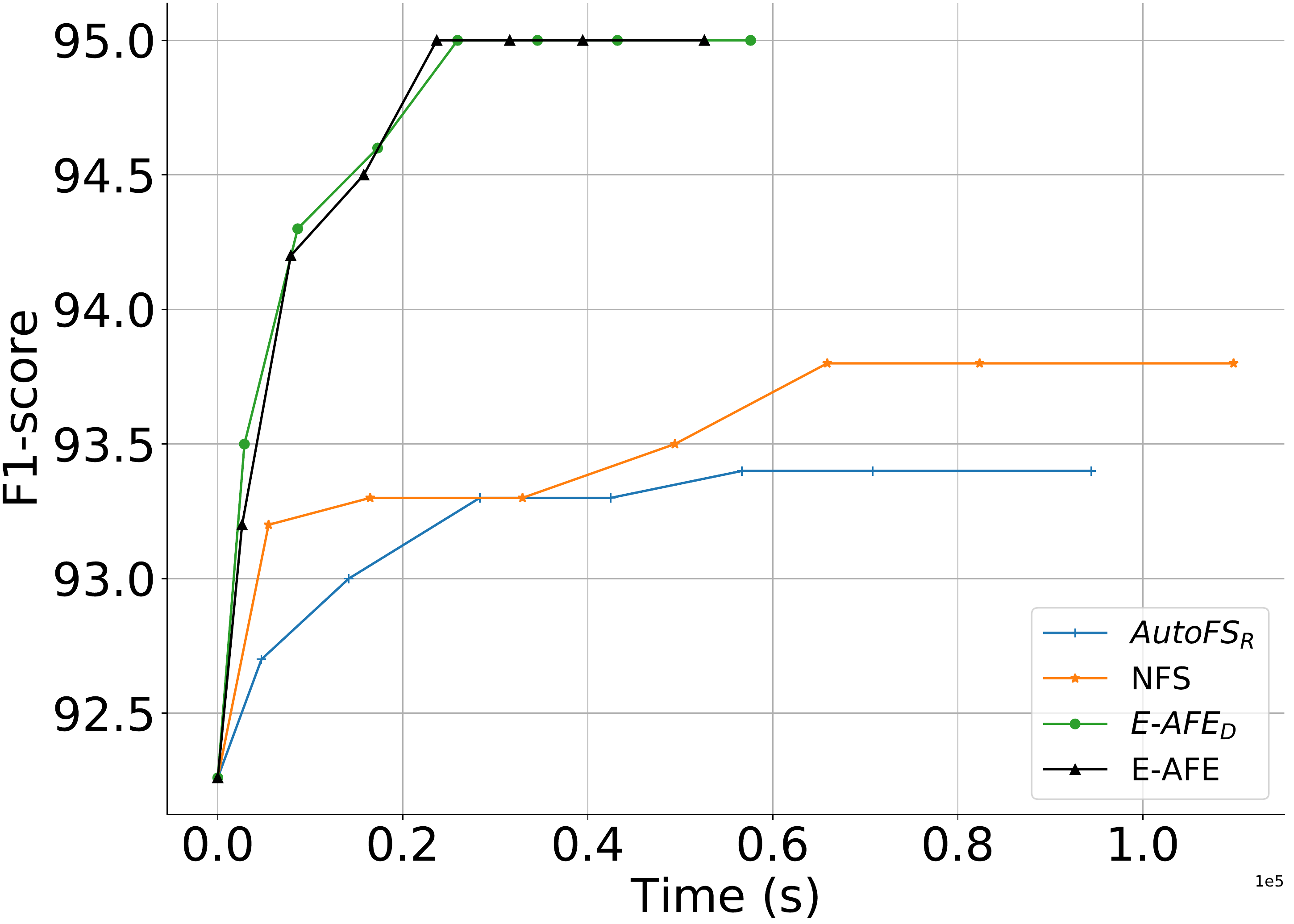}} \
    \subfloat[AP. lung]{\includegraphics[width=0.16\textwidth, height=0.10\textheight]{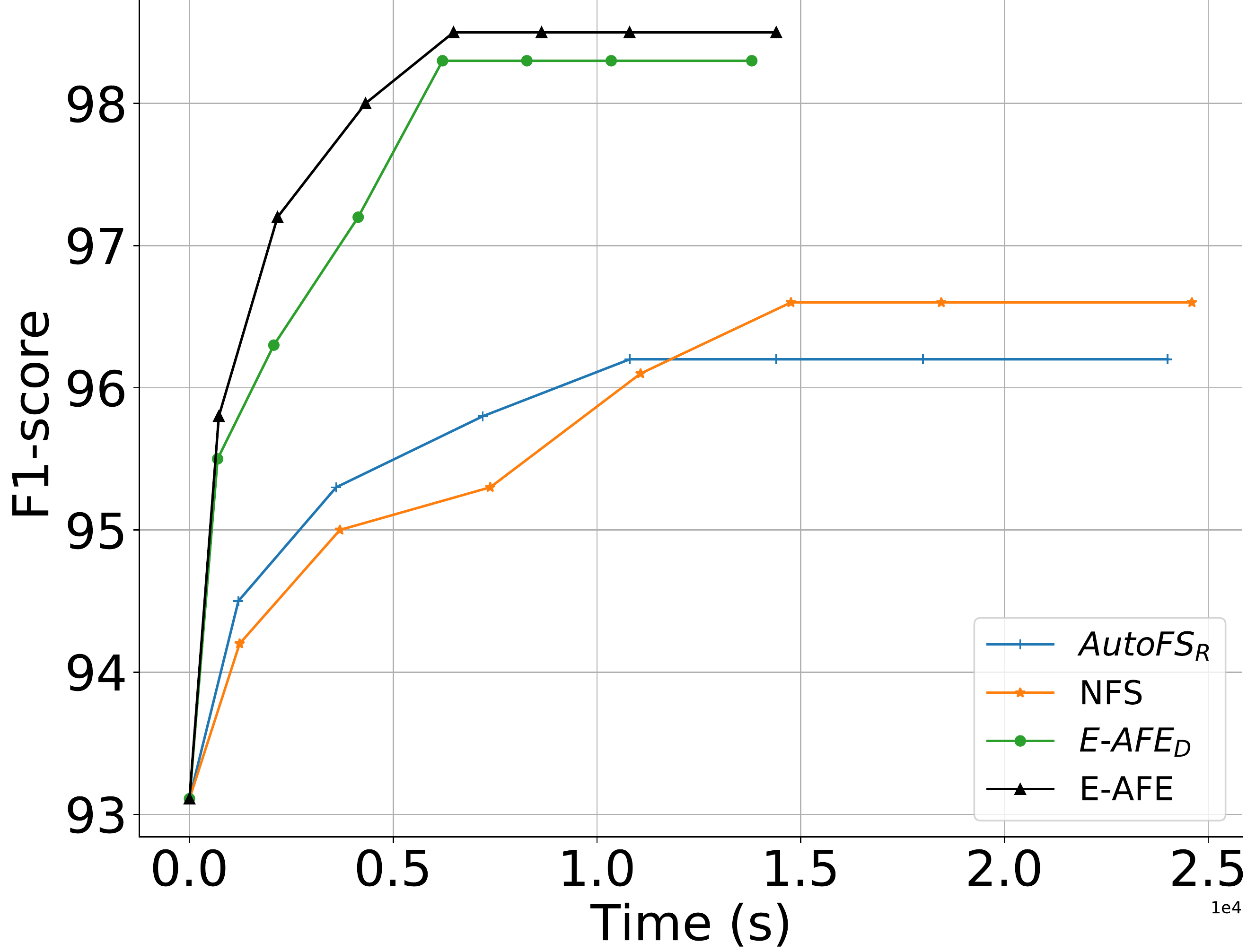}} \
    \subfloat[credit-a]{\includegraphics[width=0.16\textwidth, height=0.10\textheight]{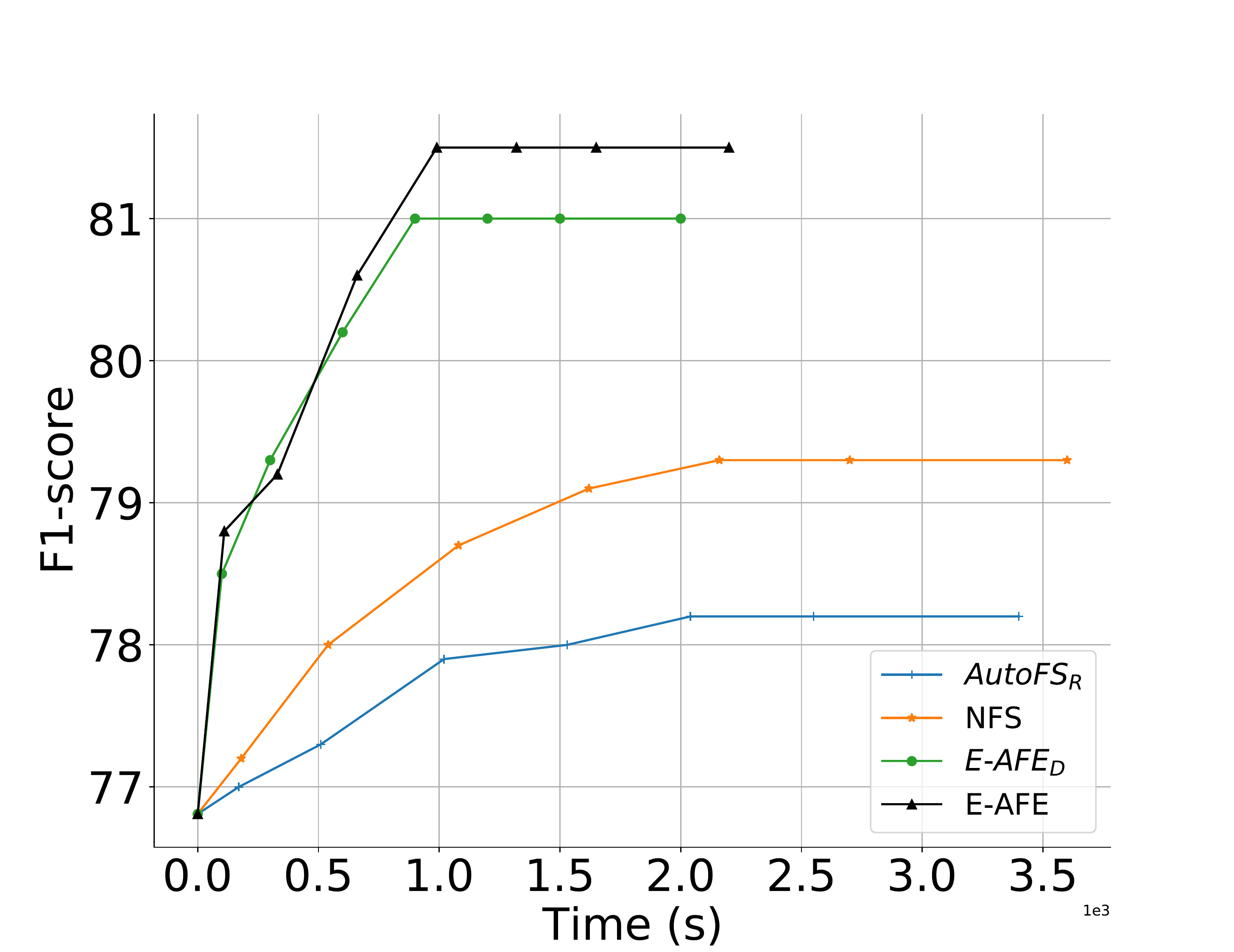}} \
    \subfloat[diabetes]{\includegraphics[width=0.16\textwidth, height=0.10\textheight]{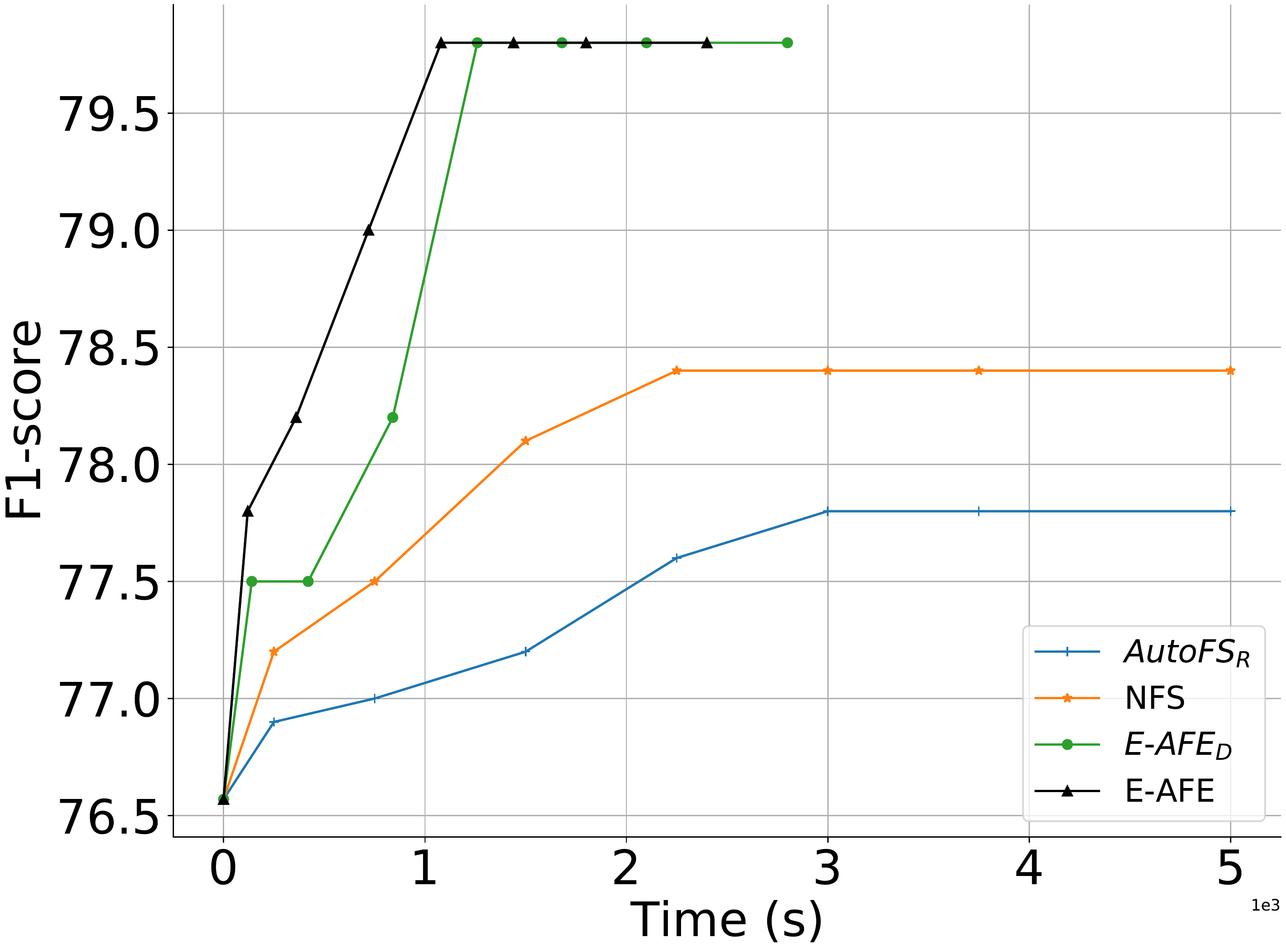}} \
    \subfloat[fertility]{\includegraphics[width=0.16\textwidth, height=0.10\textheight]{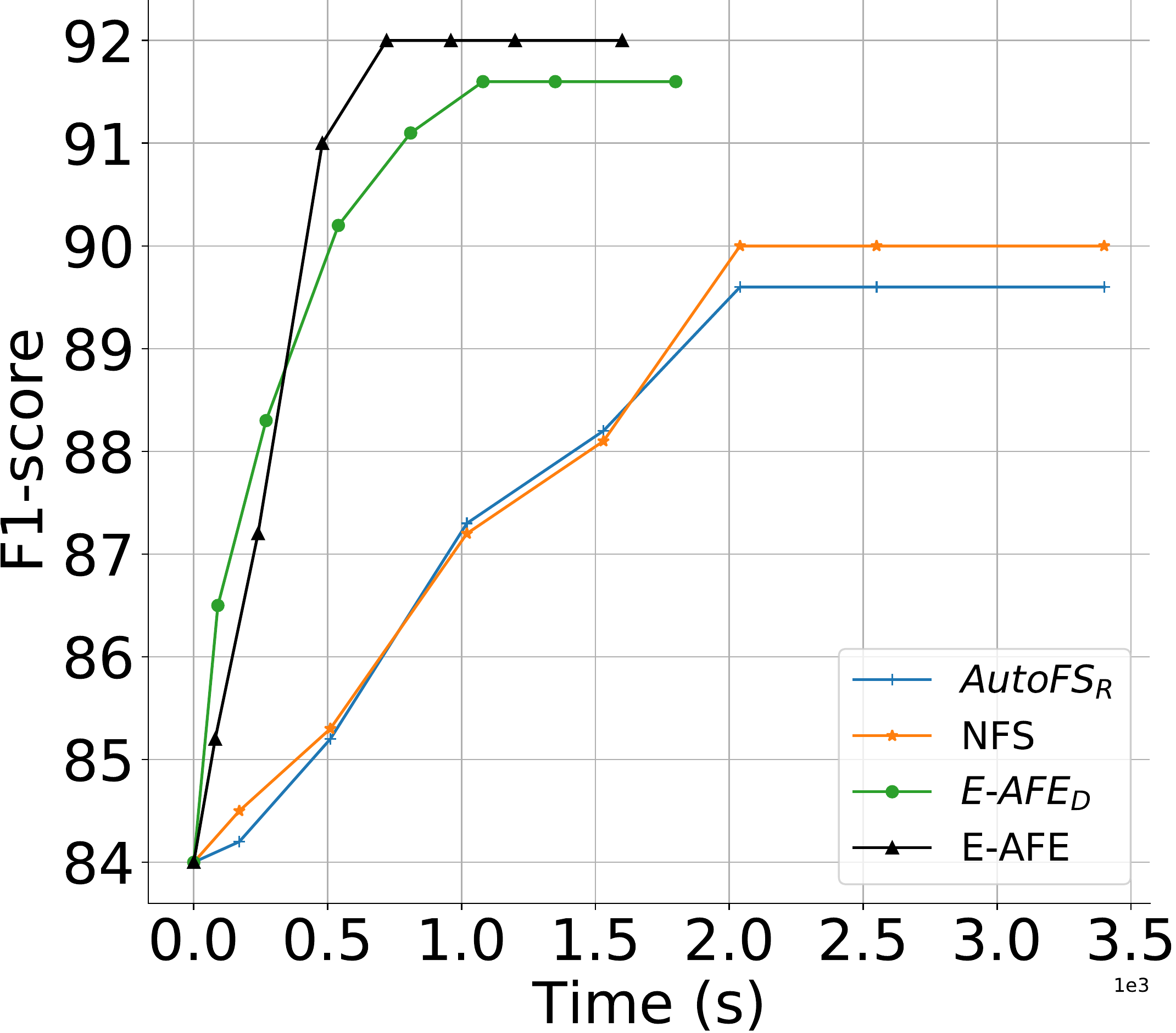}} \
    
    \subfloat[gisette]{\includegraphics[width=0.16\textwidth, height=0.10\textheight]{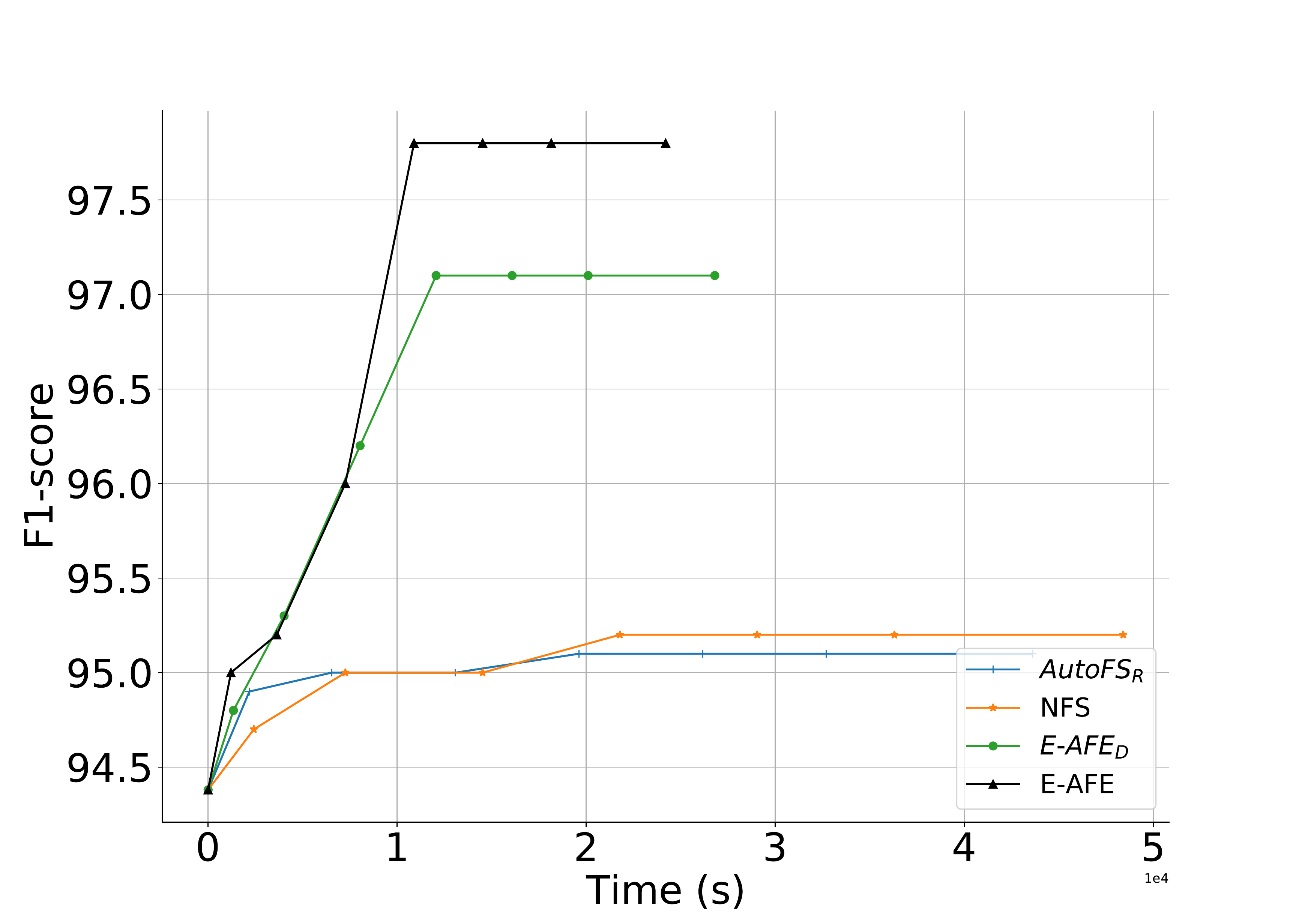}} \
    \subfloat[hepatitis]{\includegraphics[width=0.16\textwidth, height=0.10\textheight]{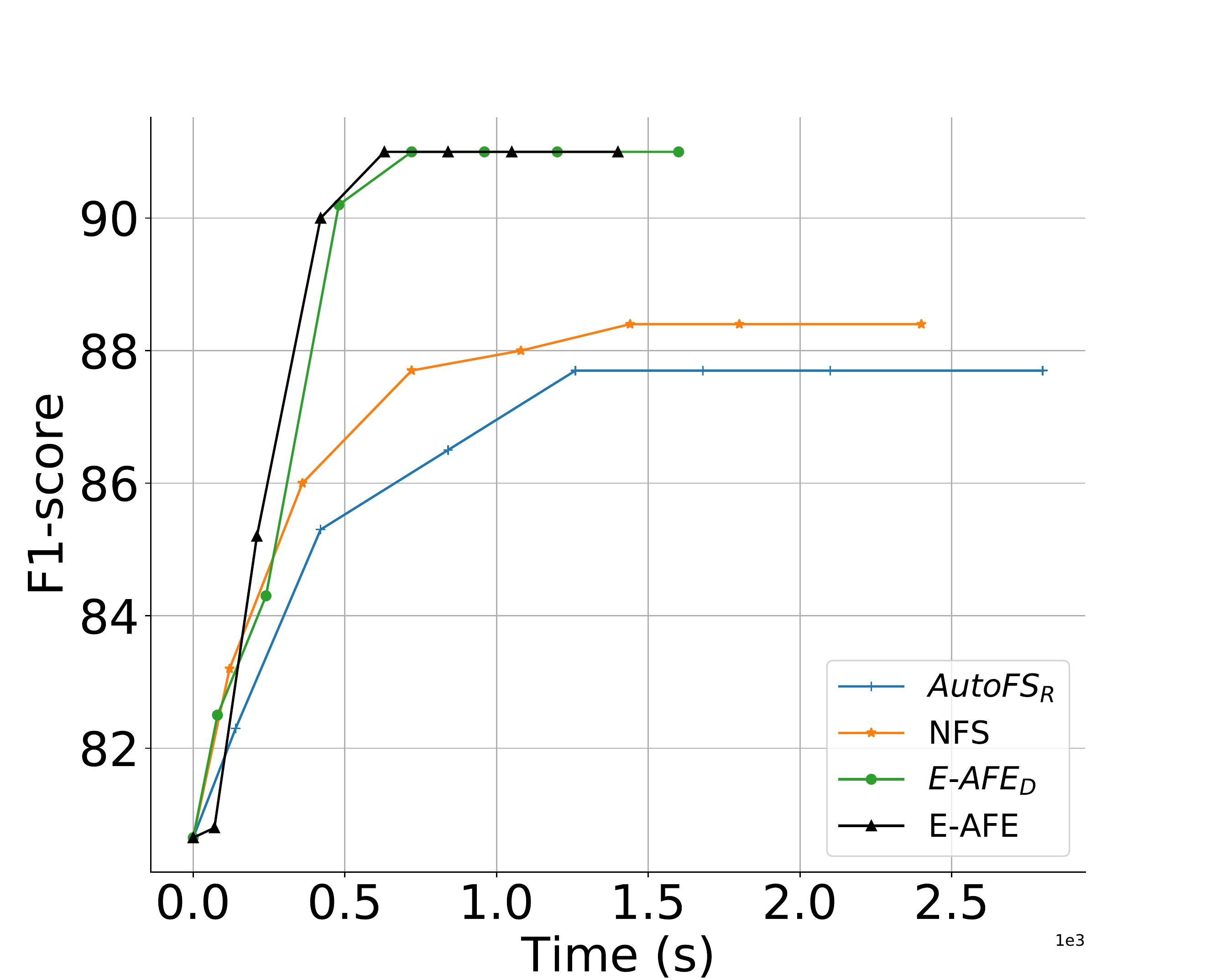}} \
    \subfloat[labor]{\includegraphics[width=0.16\textwidth, height=0.10\textheight]{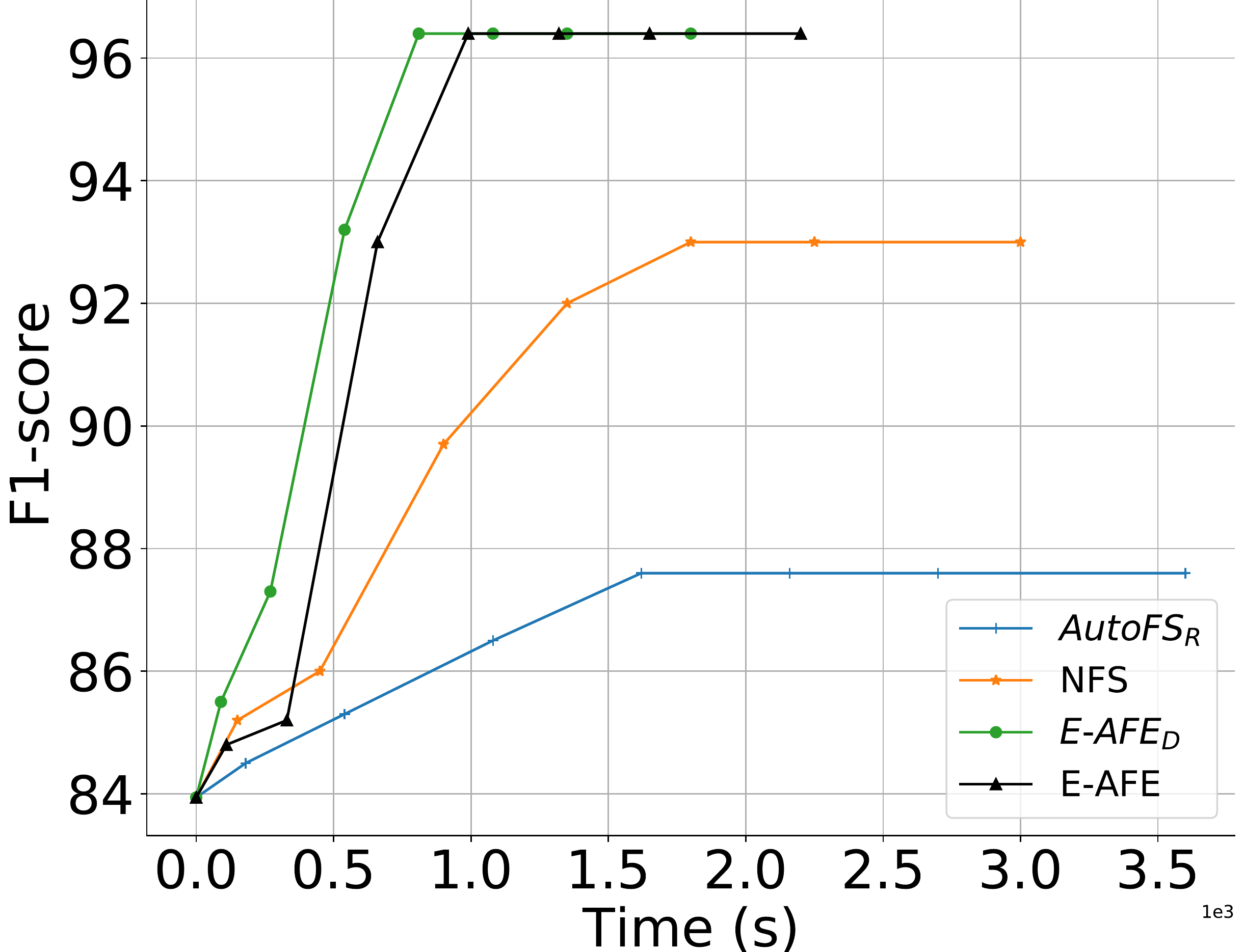}} \
    \subfloat[lymph]{\includegraphics[width=0.16\textwidth, height=0.10\textheight]{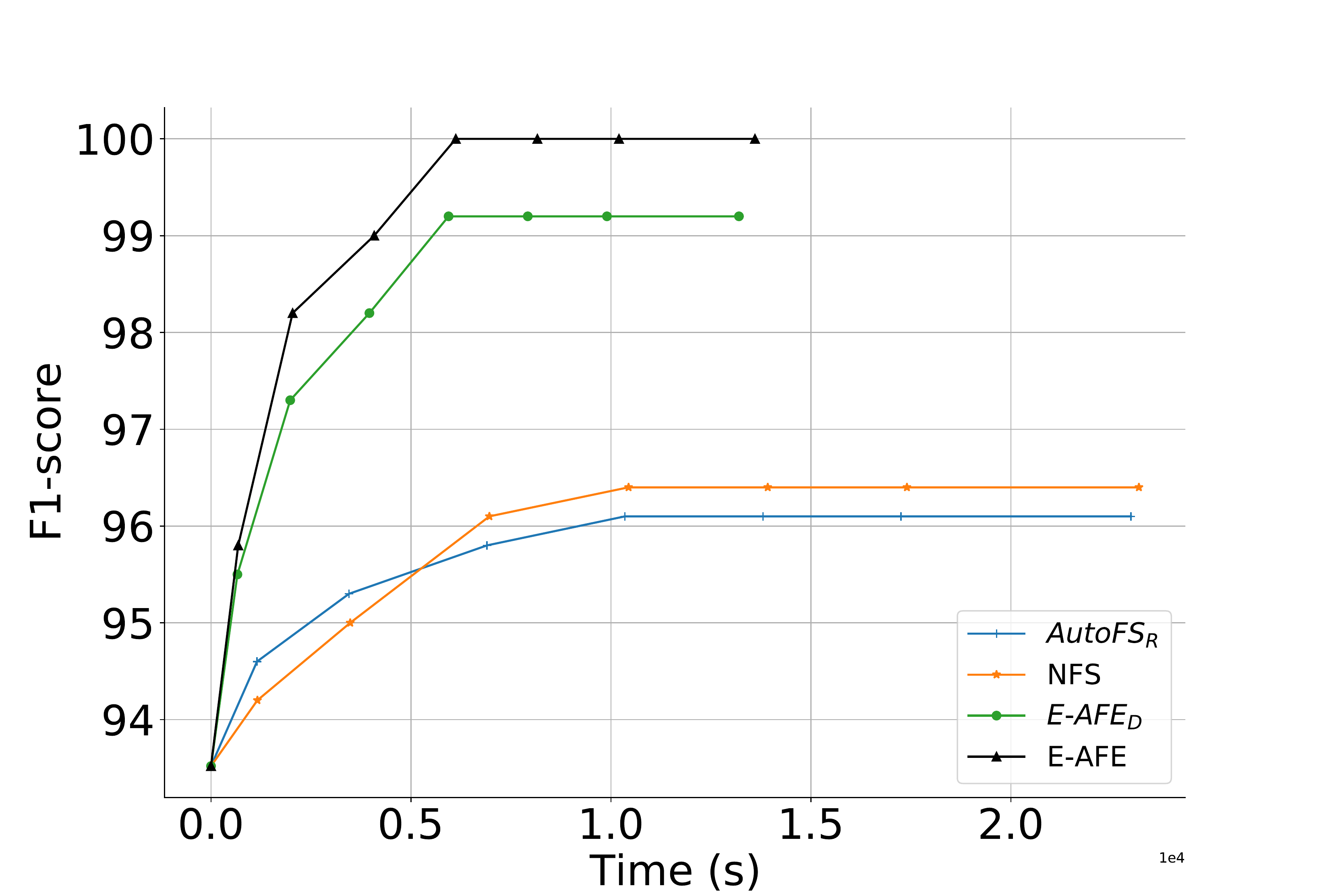}} \
    \subfloat[madelon]{\includegraphics[width=0.16\textwidth, height=0.10\textheight]{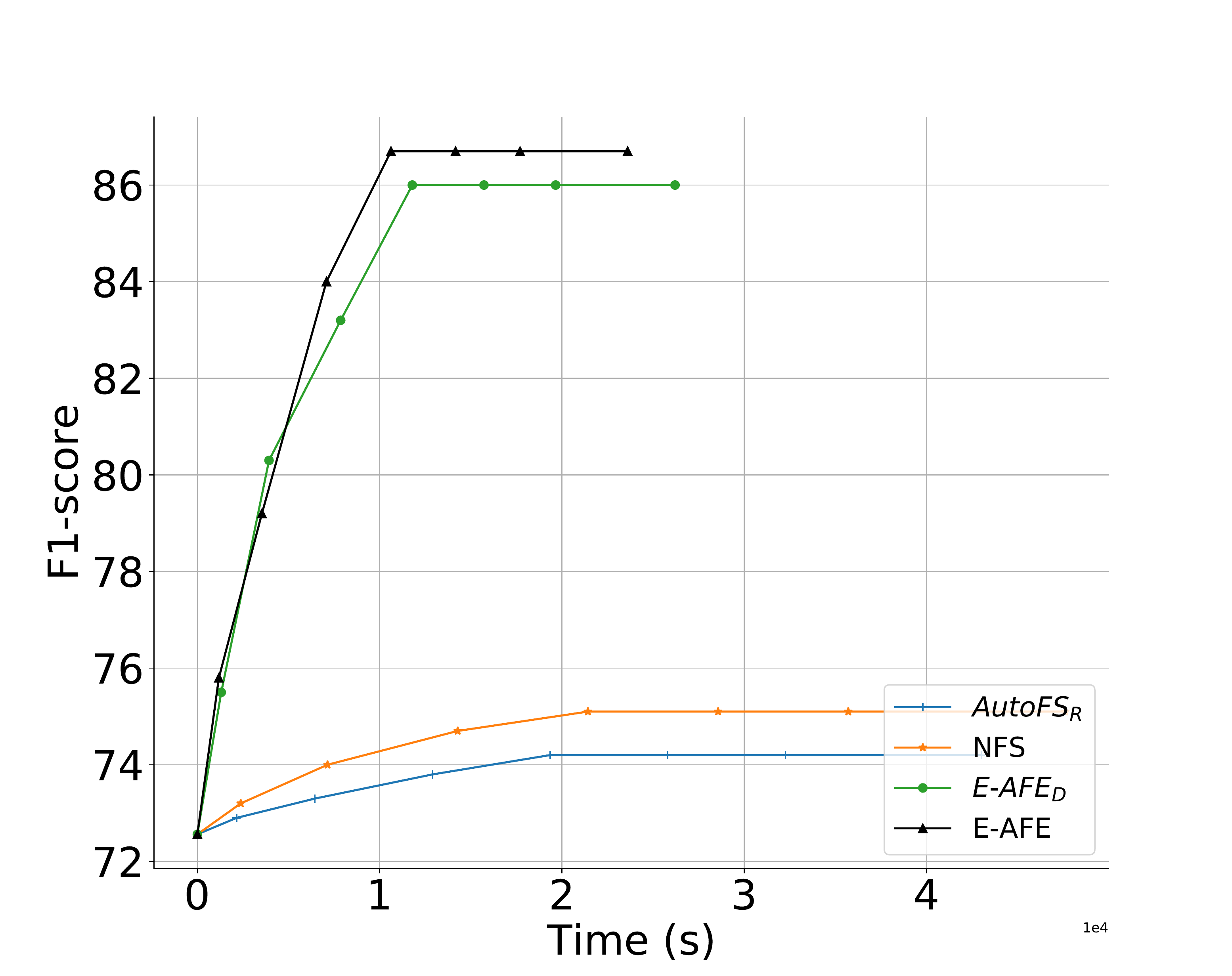}} \
    \subfloat[megawatt1]{\includegraphics[width=0.16\textwidth, height=0.10\textheight]{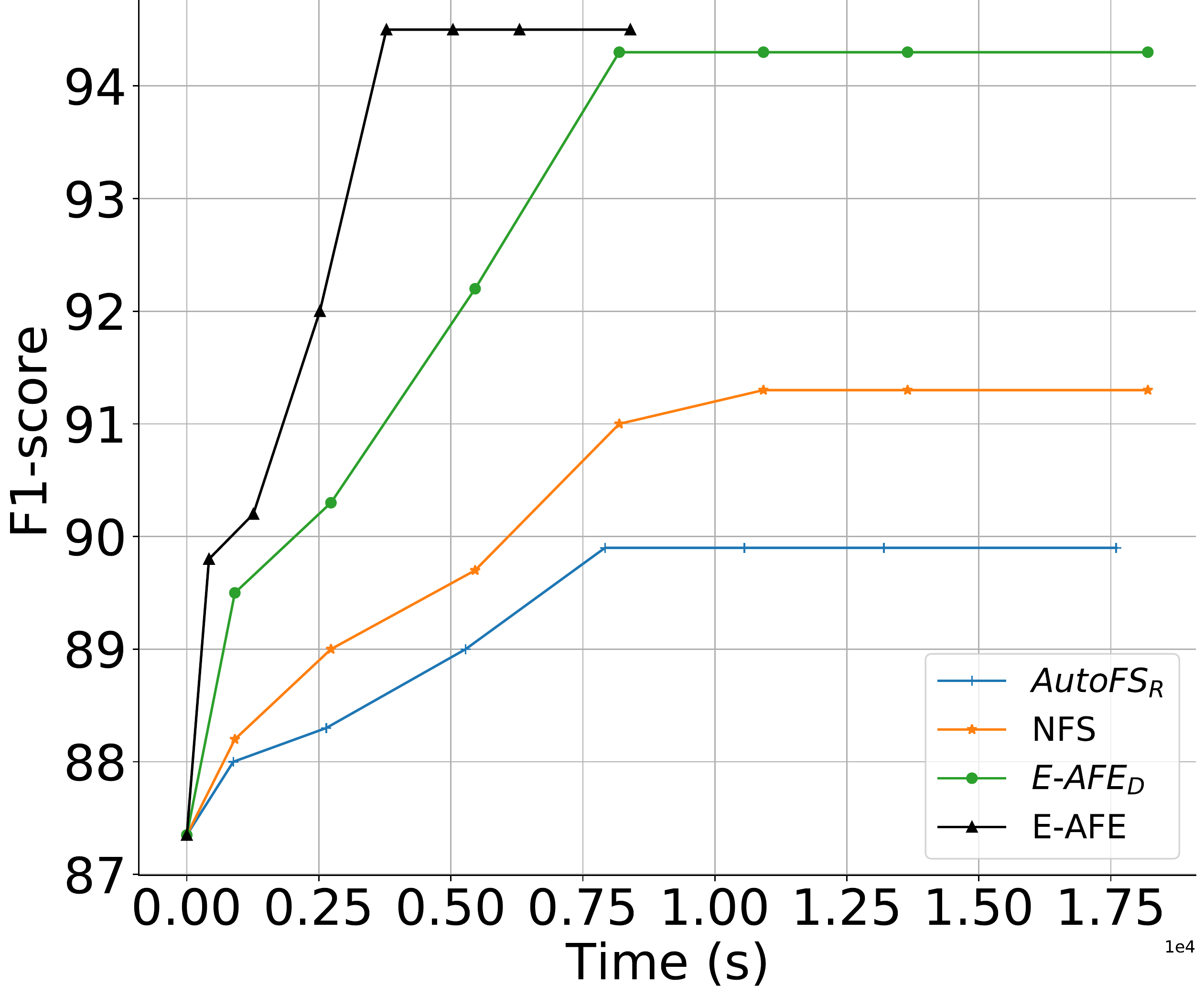}} \
    
    \subfloat[secom]{\includegraphics[width=0.16\textwidth, height=0.10\textheight]{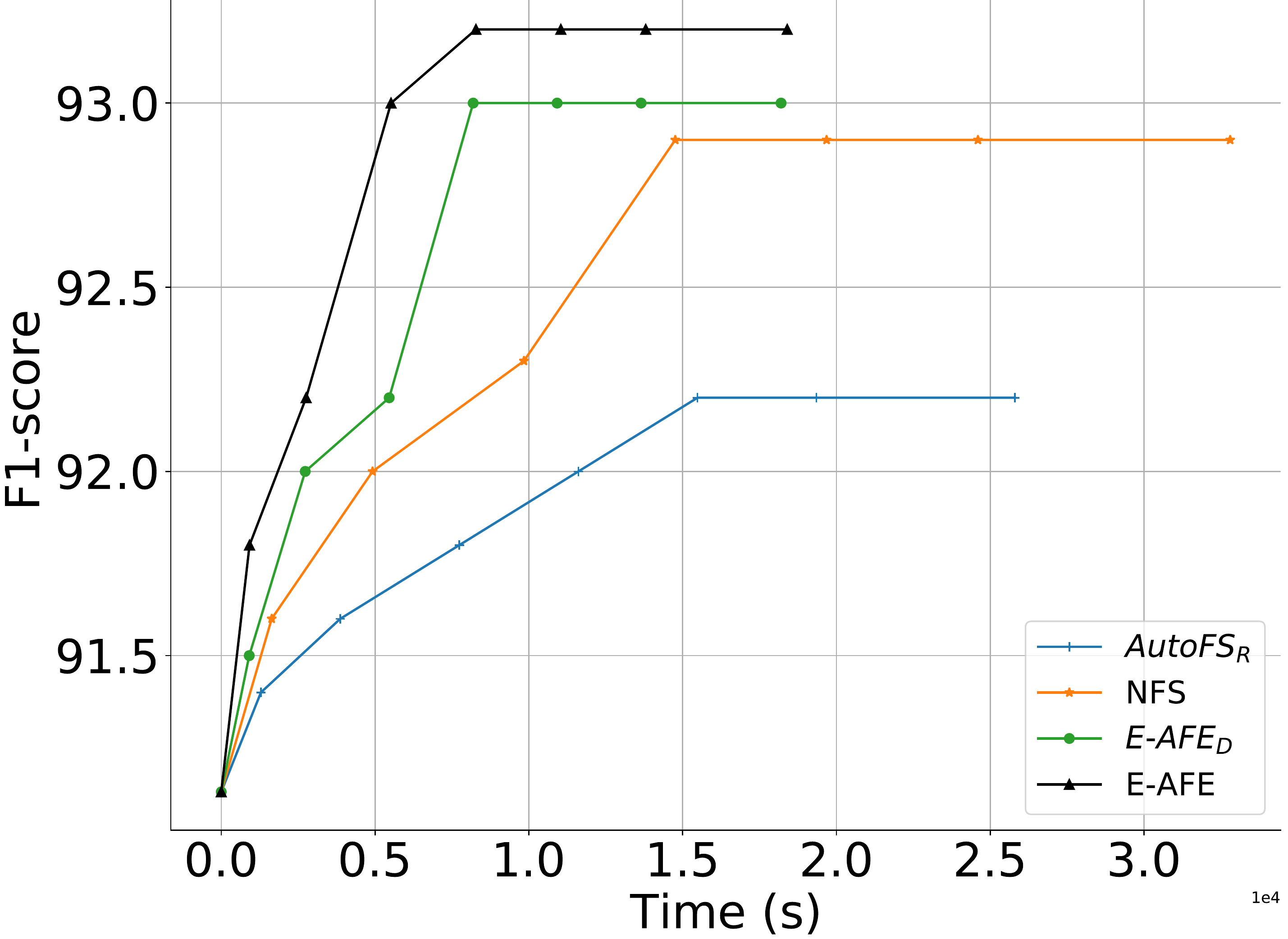}} \
    \subfloat[sonar]{\includegraphics[width=0.16\textwidth, height=0.10\textheight]{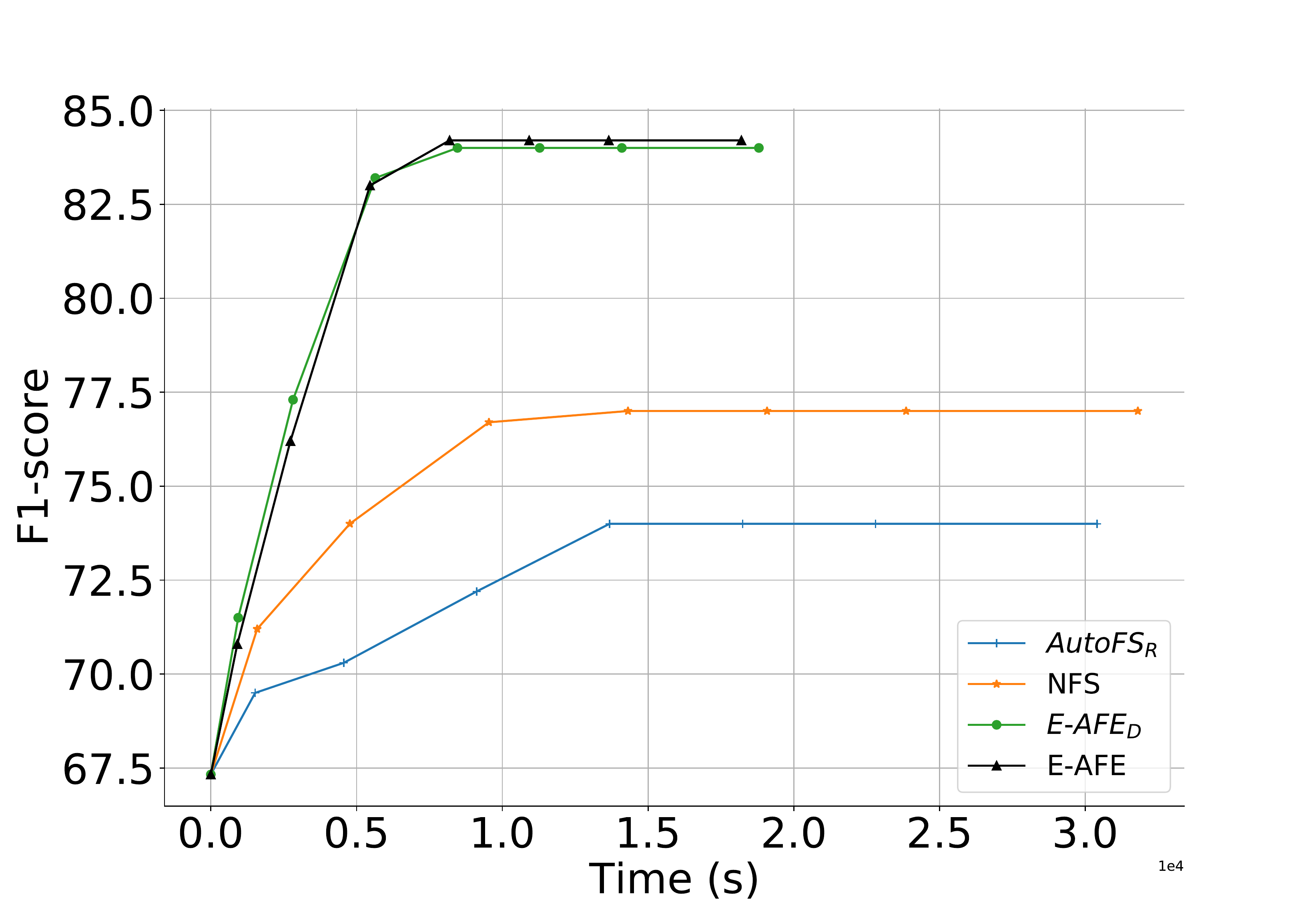}} \
    % gression
    \subfloat[Bikeshare DC]{\includegraphics[width=0.16\textwidth, height=0.10\textheight]{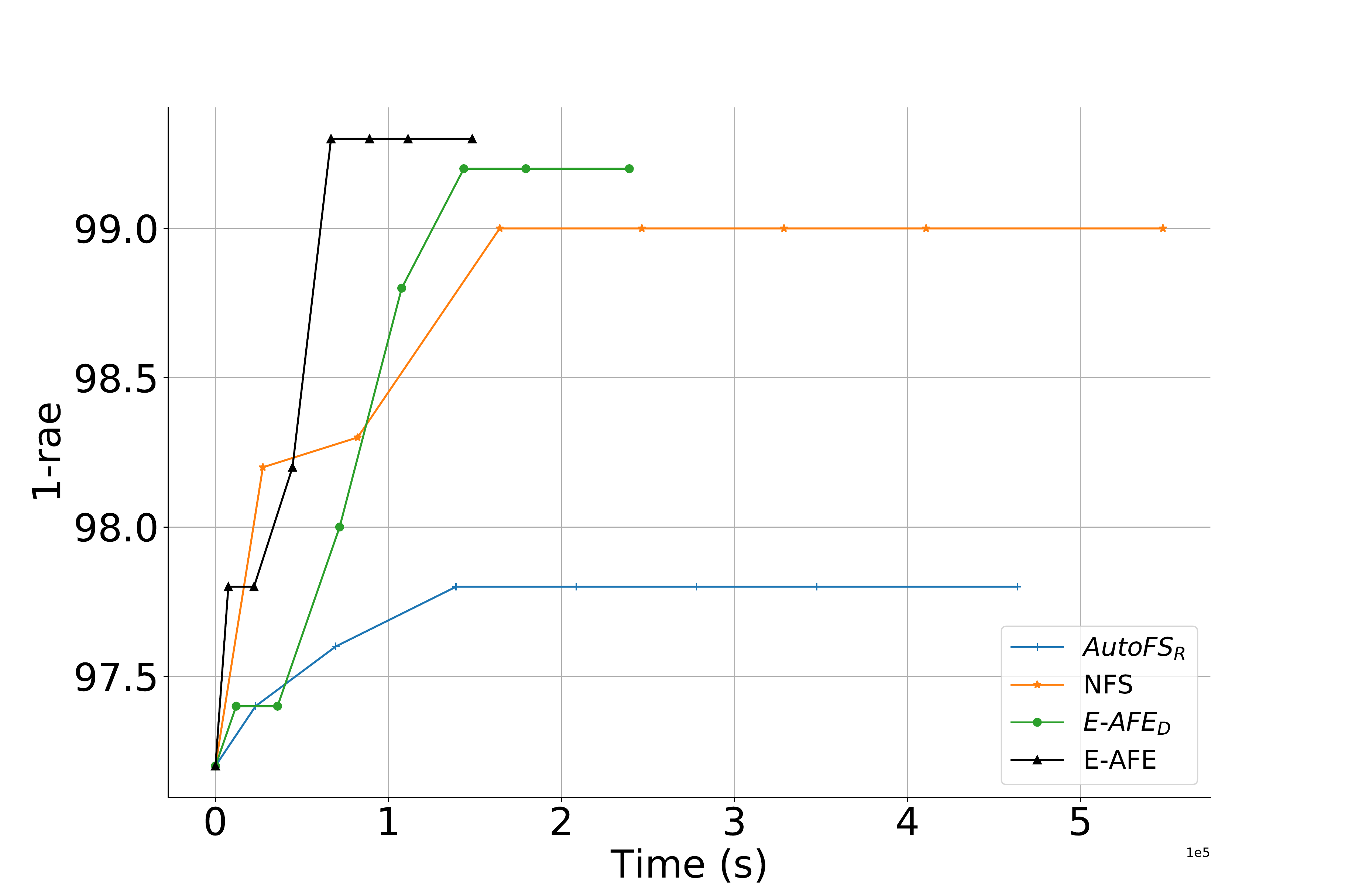}} \
    \subfloat[Housing Boston]{\includegraphics[width=0.16\textwidth, height=0.10\textheight]{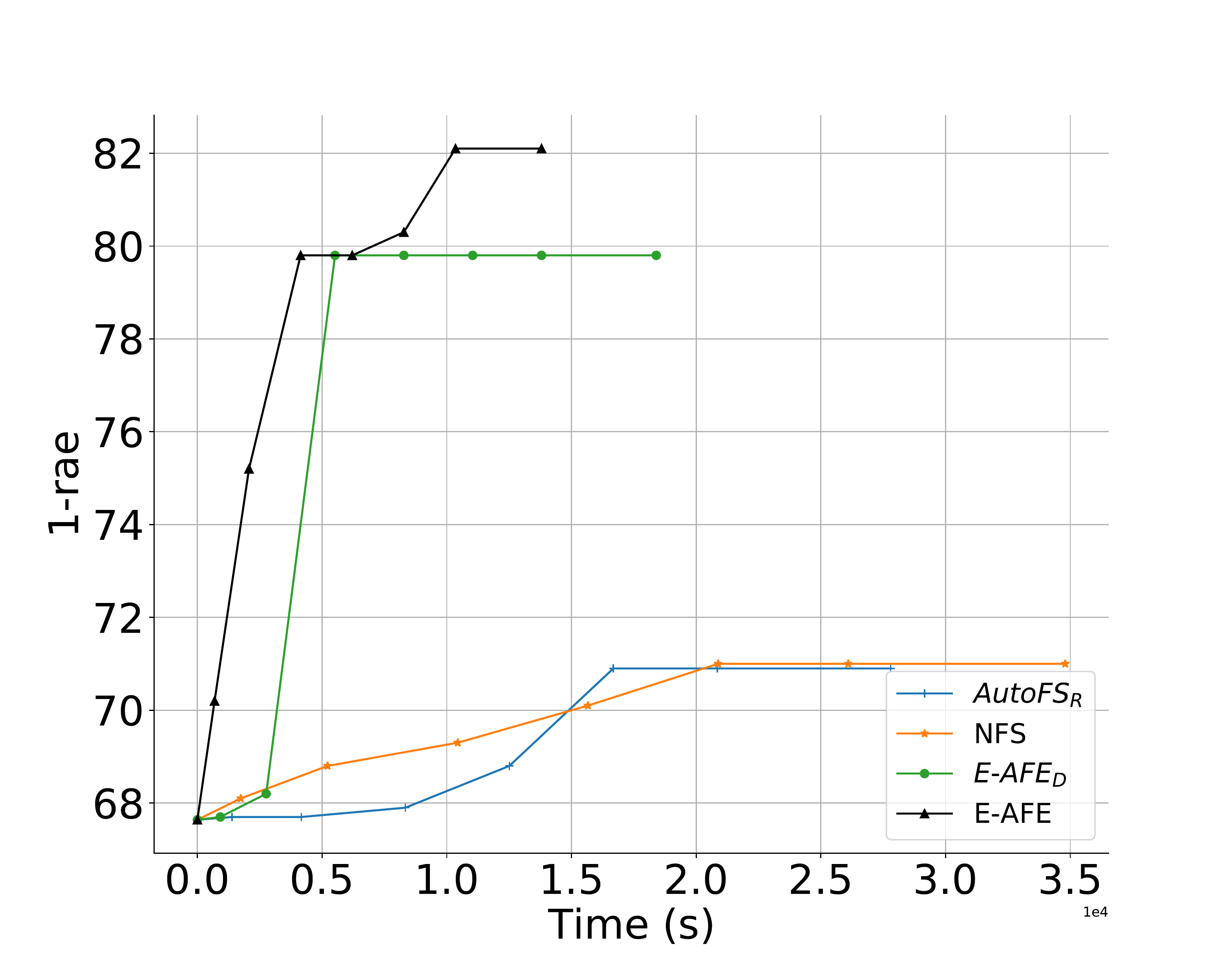}} \
    \subfloat[Airfoil]{\includegraphics[width=0.16\textwidth, height=0.10\textheight]{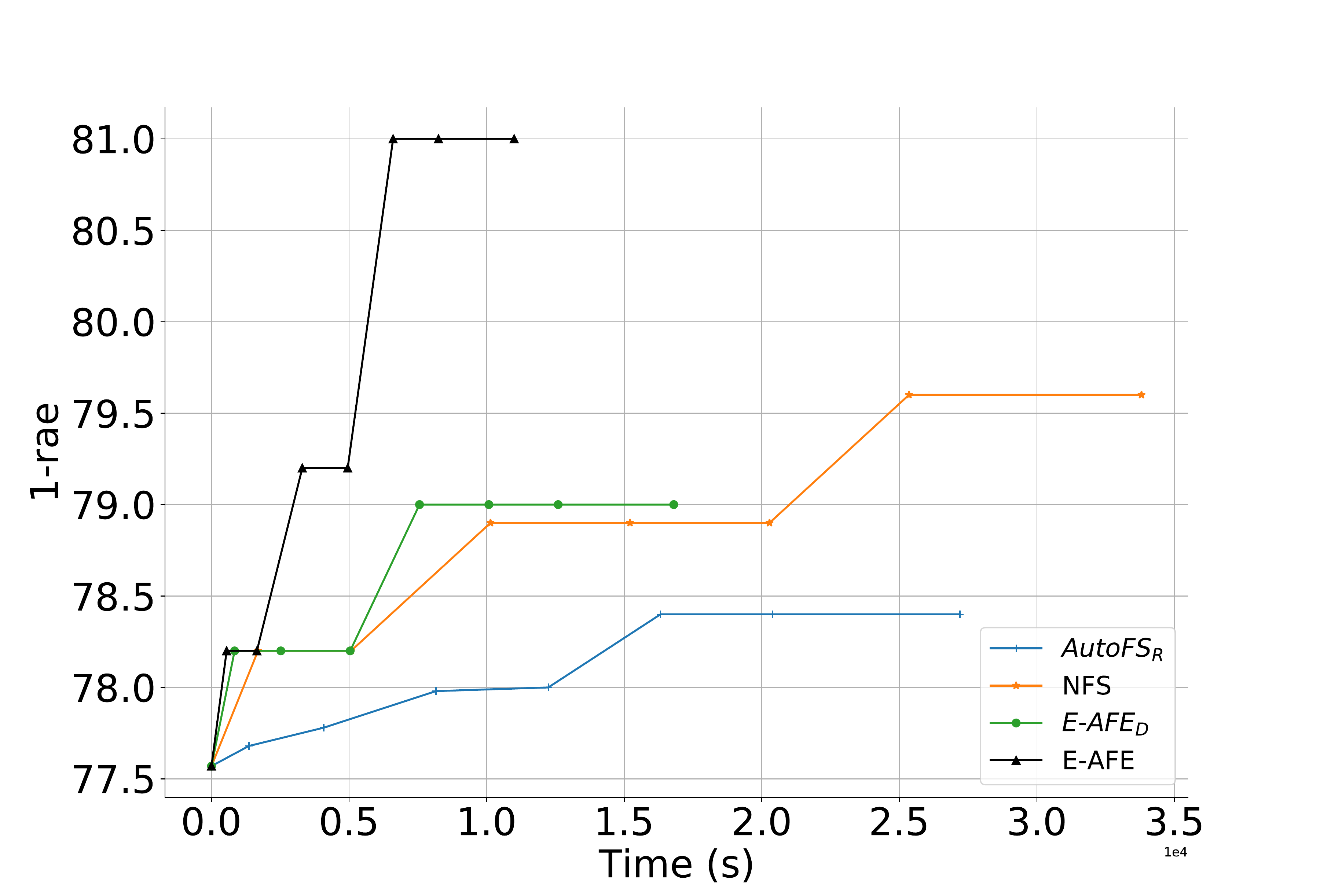}} \
    \subfloat[Openml 618]{\includegraphics[width=0.16\textwidth, height=0.10\textheight]{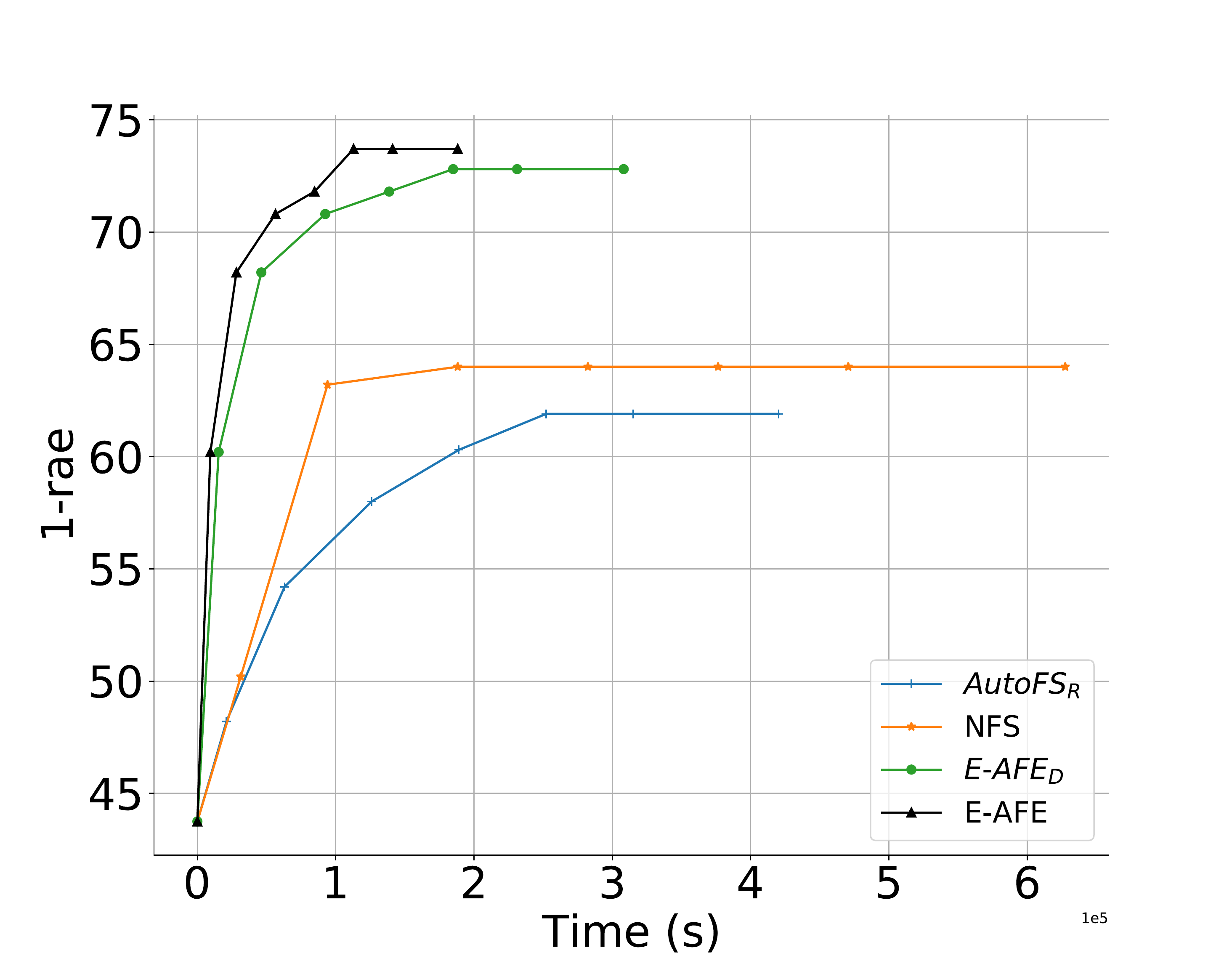}} \

    \subfloat[Openml 589]{\includegraphics[width=0.16\textwidth, height=0.10\textheight]{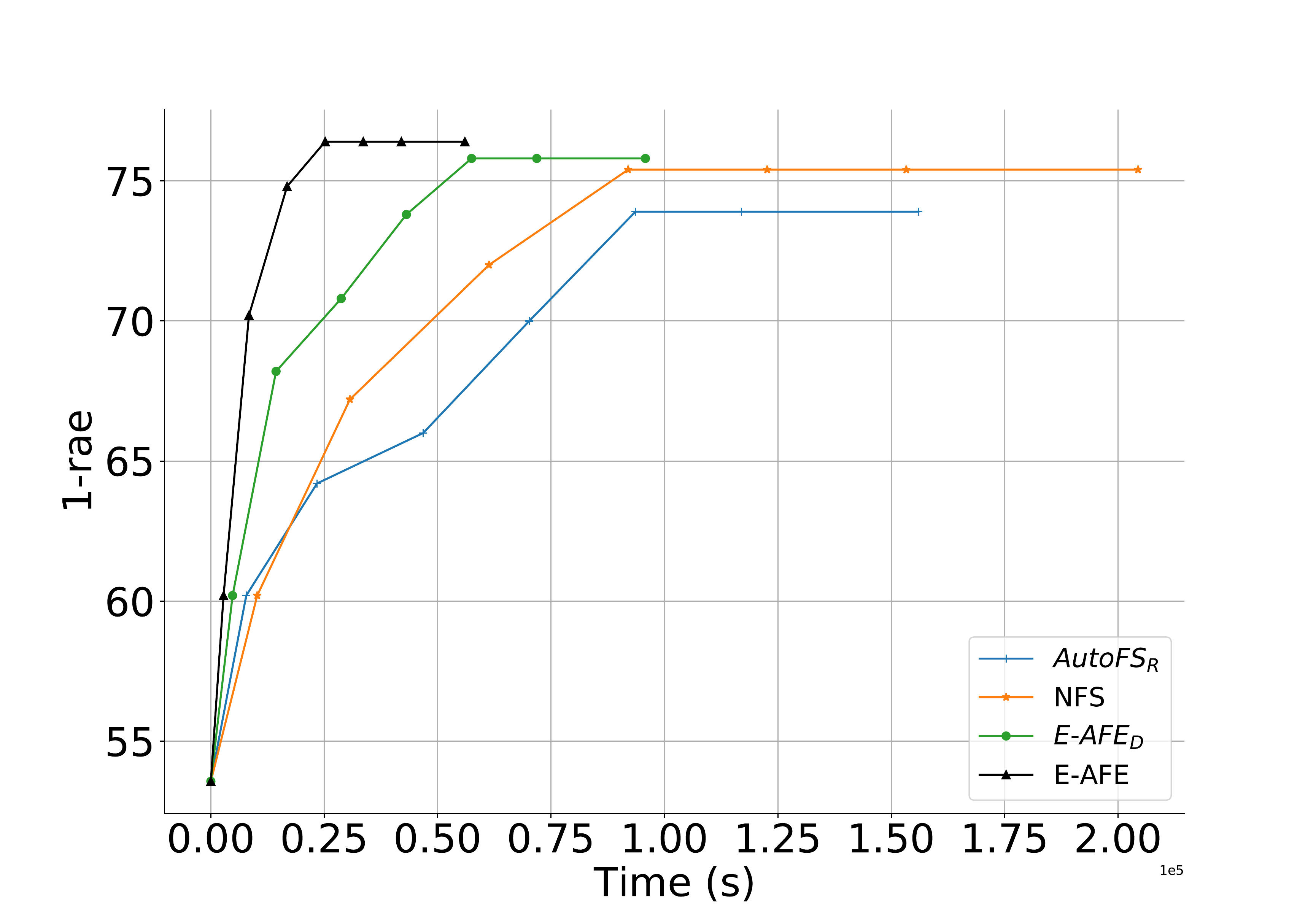}} \
    \subfloat[Openml 616]{\includegraphics[width=0.16\textwidth, height=0.10\textheight]{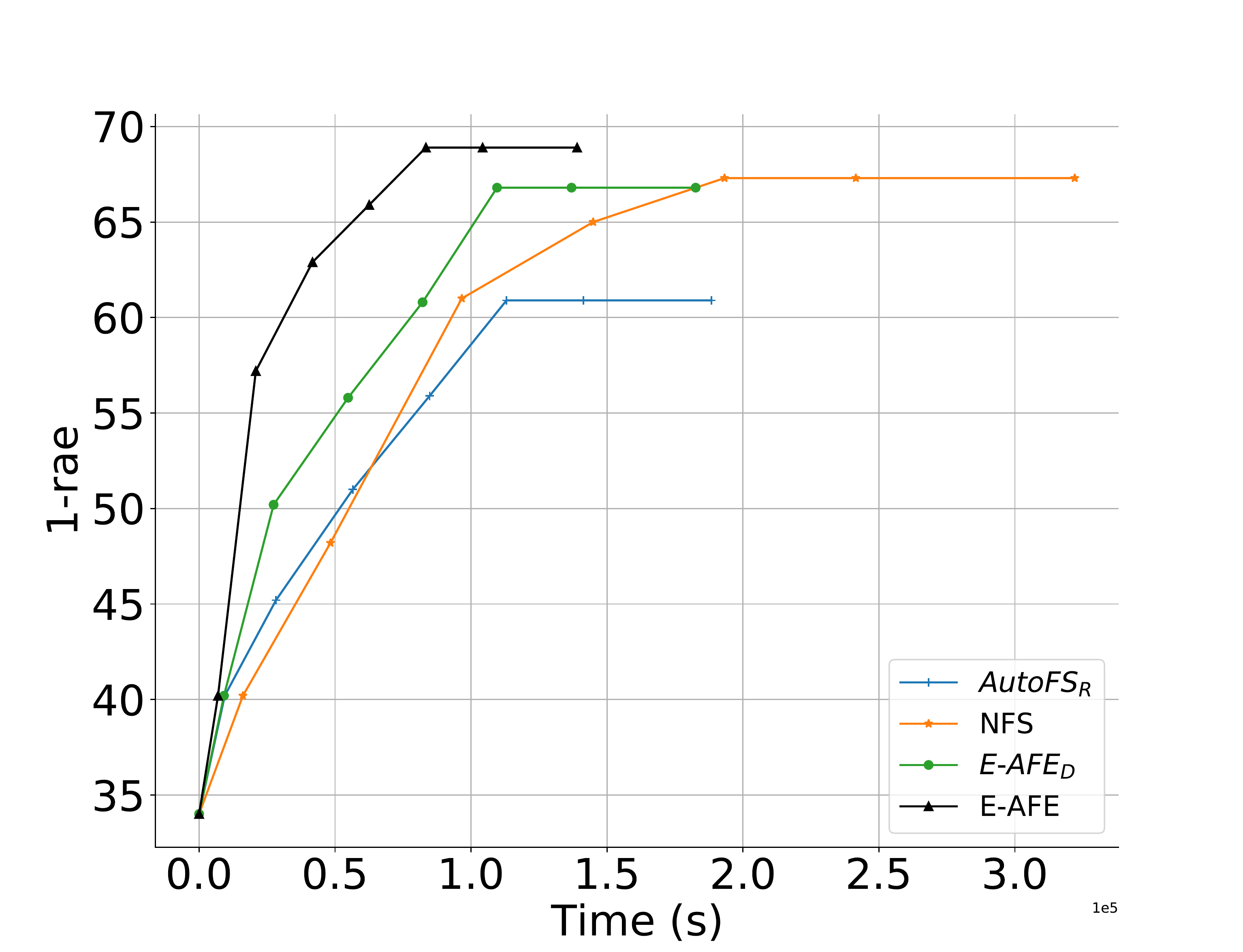}} \
    \subfloat[Openml 607]{\includegraphics[width=0.16\textwidth, height=0.10\textheight]{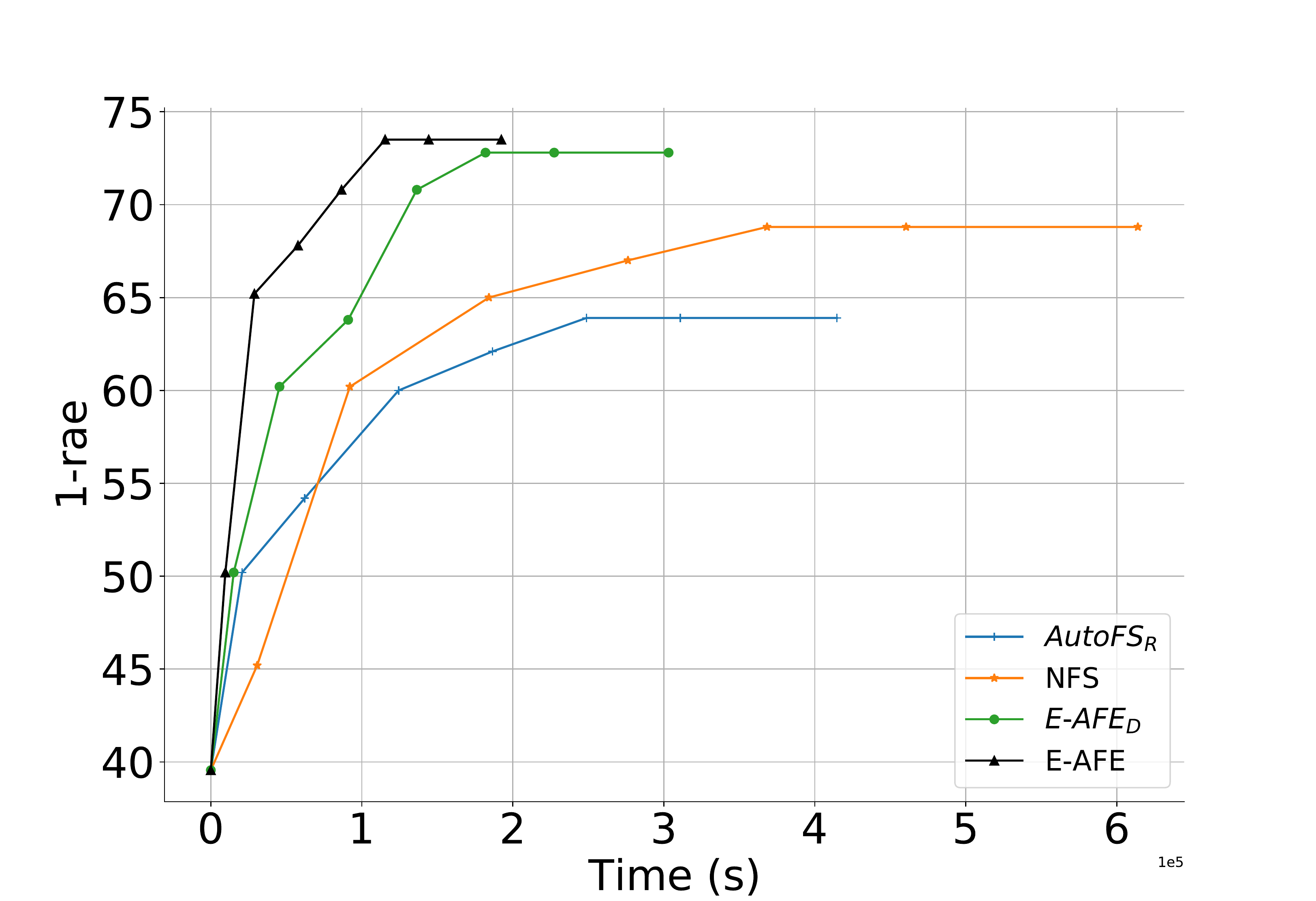}} \
    \subfloat[Openml 620]{\includegraphics[width=0.16\textwidth, height=0.10\textheight]{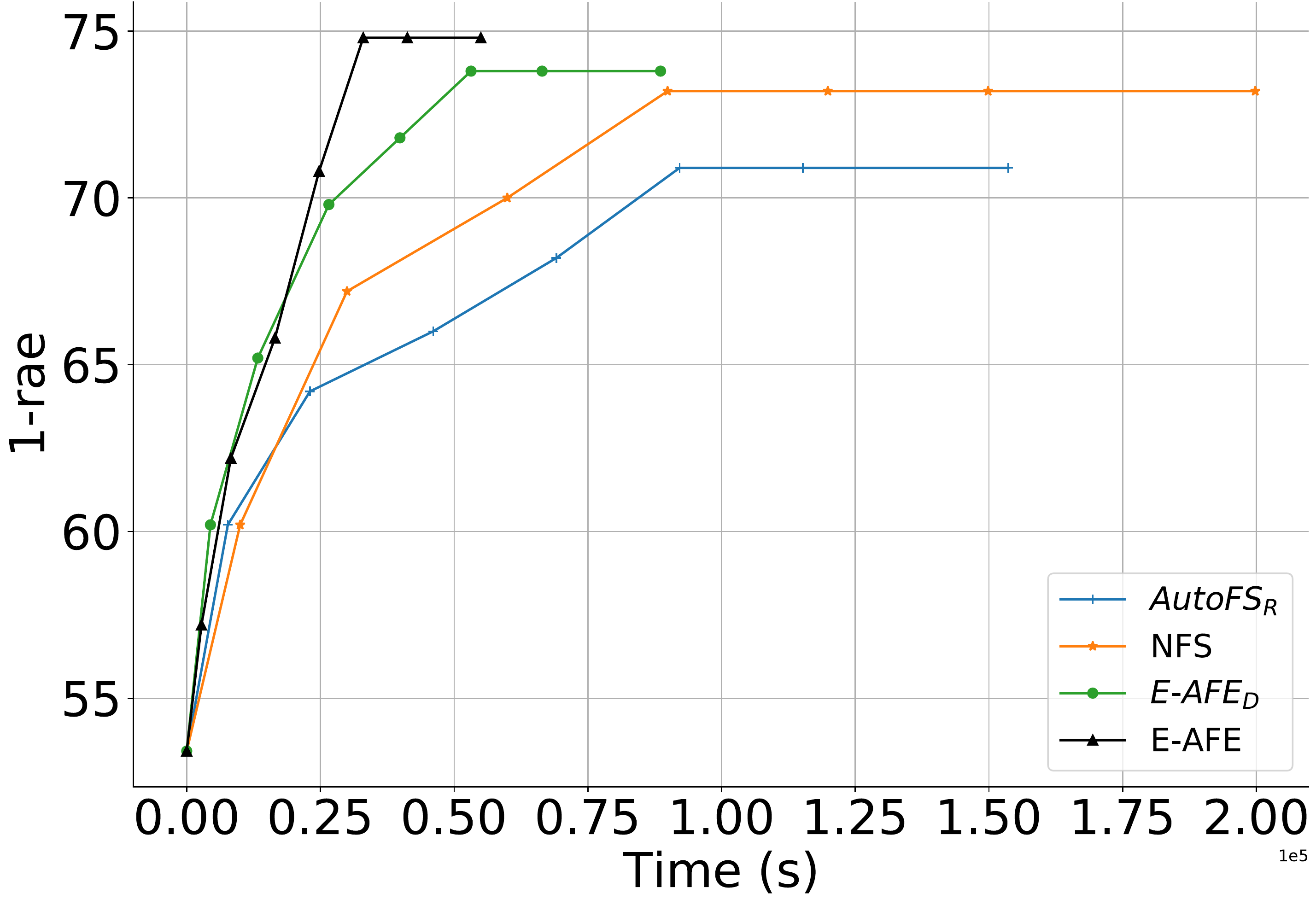}} \
    \subfloat[Openml 637]{\includegraphics[width=0.16\textwidth, height=0.10\textheight]{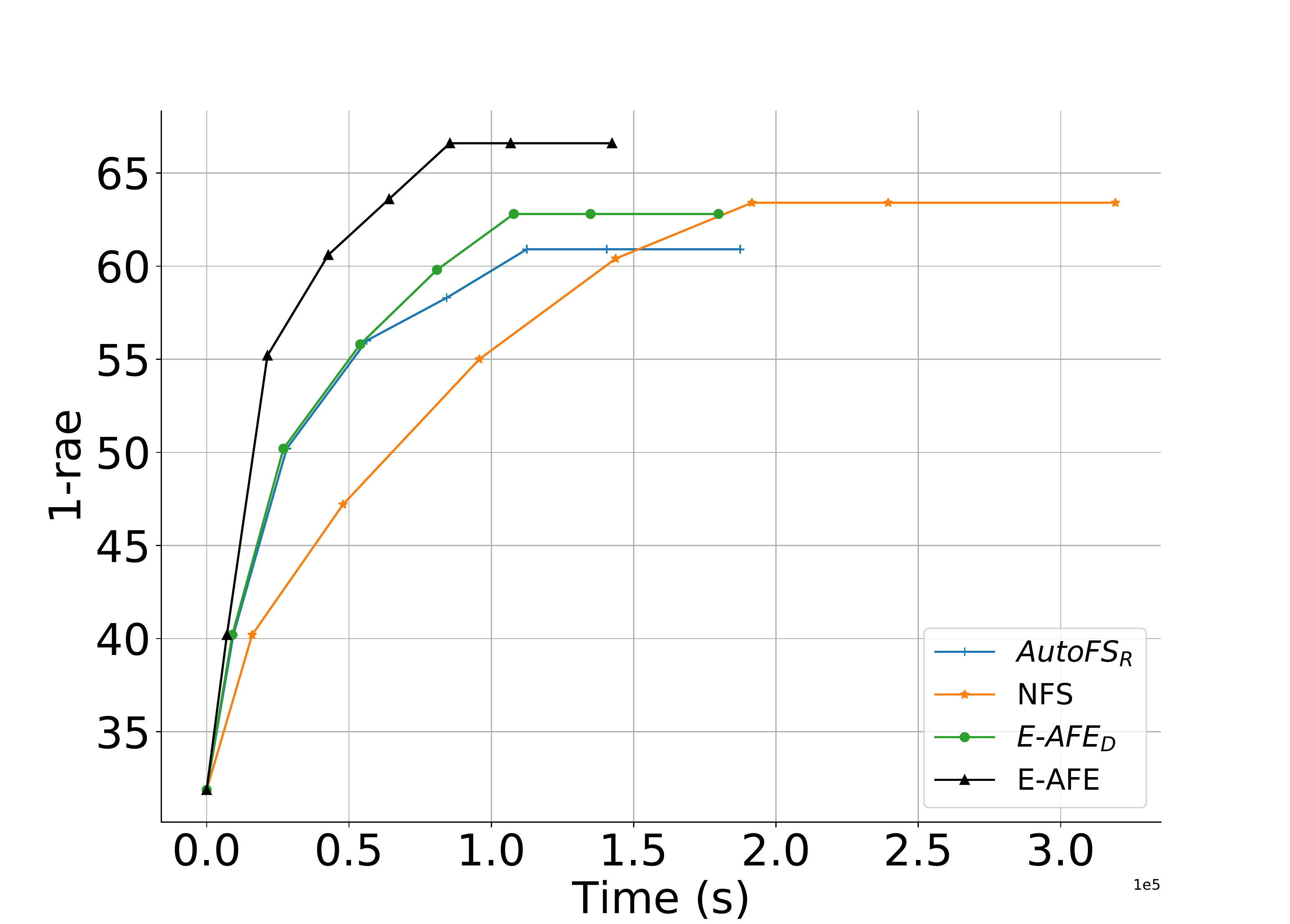}} \
    \subfloat[Openml 586]{\includegraphics[width=0.16\textwidth, height=0.10\textheight]{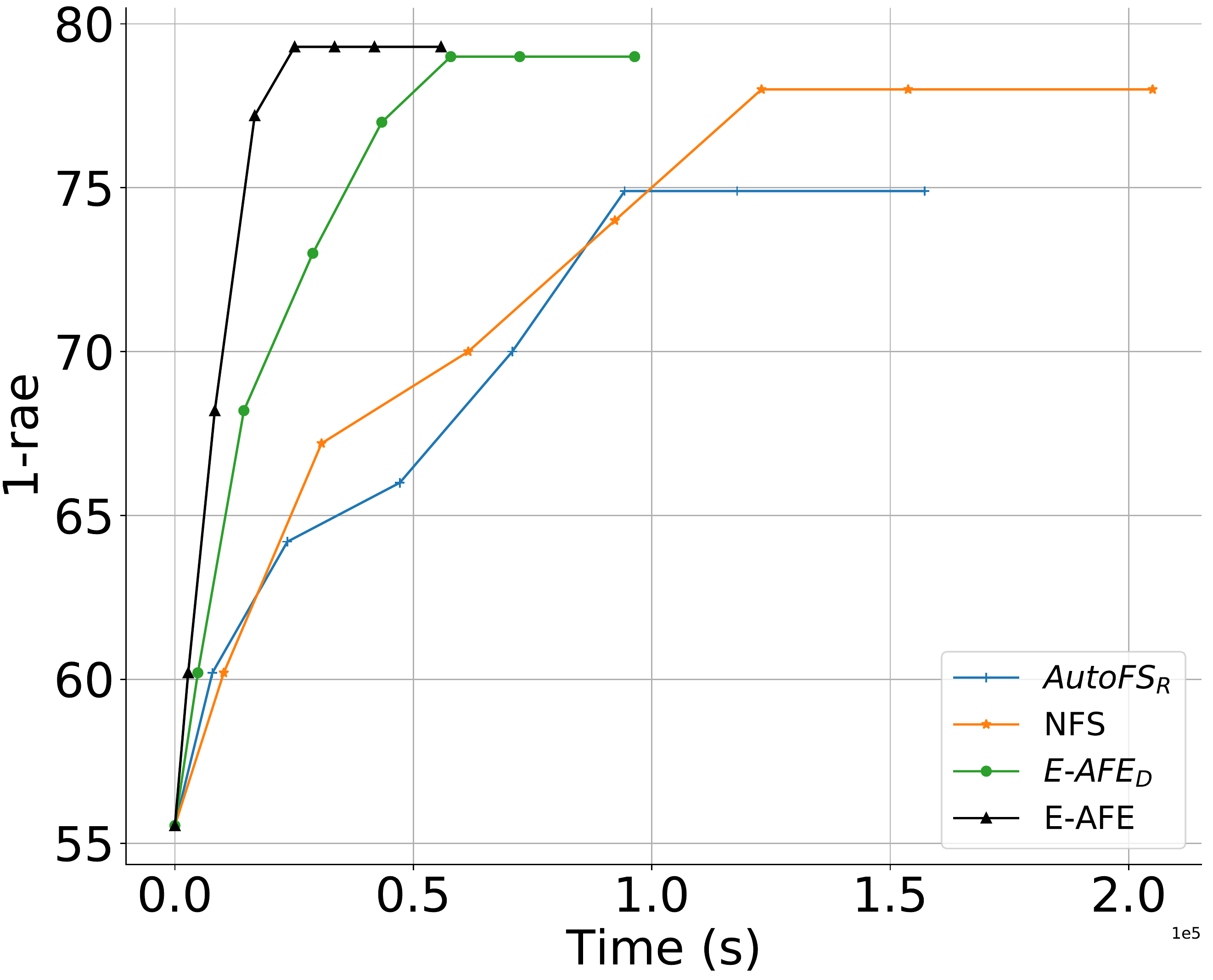}} \
    
    \caption{Performance comparison. We show the converging curve of four methods on target datasets.  $\TheName$ is our method. $\mathrm{AutoFS_{R}}$ \cite{fan2020autofs,fan2021interactive} is feature selection from randomly generated features. $\mathrm{NFS}$ \cite{chen2019neural} is feature generation and evaluation. $\TheName _{D}$ is the ablation study of ~\TheName~, replacing the approximate hashing model with a random dropout method.}
    \vspace{-0.4cm}
    \label{fig:learning_curve}
\end{figure*}

In addition to comparing ~\TheName~ with state-of-the-art techniques, we aim to understand the proposed framework better and evaluate the critical components of the FPE model. 
Mainly, we aim to answer the following question: 
How is the performance of ~\TheName~ variants with different combinations of critical components in the RL framework? 
Hence, in our evaluation, we consider the random drop feature method $\TheName _{D}$ for the ablation study:

In Table \ref{tab:comparsion_score}, \ref{tab:feature_evaluated}, and Figure \ref{fig:learning_curve}, the score of comparison between ~\TheName~ and $\TheName _{D}$, our FPE model gets a higher score. 
According to evaluated generated features by downstream task, our method is mostly evaluated less than other methods.
The results of $\TheName _{D}$ prove that the new features have redundancy for AFE. The above ablation study results also demonstrate that our method does learn more effective knowledge of discarding redundant features than dropping new features randomly.

\begin{table}
\centering
\caption{Comparison of feature evaluation numbers of one epoch in the target dataset. $\mathrm{AutoFS_{R}}$ ($\mathrm{FS_{R}}$) \cite{fan2020autofs,fan2021interactive} is feature selection from randomly generated features. $\mathrm{NFS}$ \cite{chen2019neural} is feature generation and evaluation. $\TheName _{D}$ is the ablation study of ~\TheName~, replacing the approximate hashing model with a random dropout method.}
\vspace{-0.2cm}
% \begin{adjustbox}{scale=0.65}
\begin{tabular}{c|l|l|l|l} 
\hline
Dataset           & $\mathrm{FS_{R}}$ & NFS  & $\TheName _D$ &    \TheName            \\ 
\hline
Higgs Boson       & 4640              & 4585 & 2273          & \textbf{2182}  \\ 
\hline
A. Employee       & 1440              & 1322 & 667           & \textbf{603}   \\ 
\hline
PimaIndian        & 1280              & 1230 & 585           & \textbf{474}   \\ 
\hline
SpectF            & 7040              & 6964 & 3372          & \textbf{3116}  \\ 
\hline
SVMGuide3         & 3360              & 3269 & 1577          & \textbf{1573}  \\ 
\hline
German Credit     & 3840              & 3748 & 1634          & \textbf{1473}  \\ 
\hline
Bikeshare DC      & 1600              & 1537 & 743           & \textbf{327}   \\ 
\hline
Housing Boston    & 1920              & 1826 & 898           & \textbf{559}   \\ 
\hline
Airfoil           & 800               & 698  & 333           & \textbf{161}   \\ 
\hline
AP. ovary         & 8000              & 7905 & \textbf{3937} & 4008           \\ 
\hline
Lymphography      & 2880              & 2809 & 1339          & \textbf{1160}  \\ 
\hline
Ionosphere        & 5440              & 5364 & 2600          & \textbf{2139}  \\ 
\hline
Openml 618        & 8000              & 7905 & 3903          & \textbf{1572}  \\ 
\hline
Openml 589        & 4000              & 3889 & 1934          & \textbf{975}   \\ 
\hline
Openml 616        & 8000              & 7917 & 3967          & \textbf{2009}  \\ 
\hline
Openml 607        & 8000              & 7900 & 3960          & \textbf{1671}  \\ 
\hline
Openml 620        & 4000              & 3883 & 1924          & \textbf{929}   \\ 
\hline
Openml 637        & 8000              & 7894 & 4040          & \textbf{2038}  \\ 
\hline
Openml 586        & 4000              & 3891 & 1996          & \textbf{932}   \\ 
\hline
Credit Default    & 3680              & 3593 & 1823          & \textbf{1608}  \\ 
\hline
Messidor features & 3040              & 2958 & 1470          & \textbf{1448}  \\ 
\hline
Wine Q. Red       & 1760              & 1665 & 867           & \textbf{806}   \\ 
\hline
Wine Q. White     & 1760              & 1646 & 849           & \textbf{596}   \\ 
\hline
SpamBase          & 9120              & 9016 & 4347          & \textbf{4015}  \\ 
\hline
AP. lung          & 8000              & 7905 & \textbf{3954} & 3966           \\ 
\hline
credit-a          & 960               & 898  & \textbf{415}  & 483            \\ 
\hline
diabetes          & 1280              & 1208 & 589           & \textbf{489}   \\ 
\hline
fertility         & 1440              & 1361 & 670           & \textbf{552}   \\ 
\hline
gisette           & 8000              & 7875 & 3954          & \textbf{3667}  \\ 
\hline
hepatitis         & 960               & 836  & 407           & \textbf{363}   \\ 
\hline
labor             & 1280              & 1202 & \textbf{596}  & 695            \\ 
\hline
lymph             & 8000              & 7891 & 4042          & \textbf{4033}  \\ 
\hline
madelon           & 8000              & 7923 & 3549          & \textbf{3549}  \\ 
\hline
megawatt1         & 5760              & 5695 & 2845          & \textbf{2319}  \\ 
\hline
secom             & 3200              & 3087 & \textbf{1559} & 1647           \\ 
\hline
sonar             & 9600              & 9501 & \textbf{4747} & 4762           \\
\hline
\end{tabular}
\vspace{-0.5cm}
% \end{adjustbox}
\label{tab:feature_evaluated}
\end{table}

\subsection{Effect of RL in ~\TheName~ Framework (\textbf{Q3})}
To show the effect of the RL framework in our developed ~\TheName~, we replaced our designed RL framework with the policy gradient method $\TheName _{R}$ for the ablation study:

Table \ref{tab:comparsion_score} shows that our RL framework method ~\TheName~ is better than $\TheName _{R}$. 
Figure \ref{fig:overview} shows that our RL framework explores knowledge from offline public datasets and exploits new knowledge online from a target dataset. 
Especially our method not only use the final result of the downstream task but also cache the intermediate result of the downstream task in the process of training agents. This RL framework is also the source of improving the score of our method compared with the NFS \cite{chen2019neural} method, which uses the policy gradient to train the controller and get a result at the final test. However, NFS omitted the cross-validation results in the training process, resulting in time-consuming and poor results.

\subsection{Comparison with DNN method (\textbf{Q4})}
To compare the feature engineering method with the deep learning method, we invite the RTDL \cite{gorishniy2021revisiting} method for baseline.

In Table \ref{tab:comparsion_score}, we can see that feature engineering-related methods are more robust than ResNet from RTDL.
The results of ResNet have 0.0 or near 0.0.  
We consider that ResNet is less useful in mini datasets, such as Lymphography and hepatitis. In the mini dataset, NFS \cite{chen2019neural} and ~\TheName~ have significant advantages to ResNet. 
However, if the dataset is large, ResNet performs similarly to NFS \cite{chen2019neural} and ~\TheName~, such as Higgs Boson, SpamBase, AP. lung, and gisette. 
The robustness loss of DNN results comes from its pre-division of data sets into training, validation, and test sets. Especially for small data sets, this partition is a fatal disadvantage.
The robust result comes from the cross-validation downstream task of the feature engineering-related method, though this cross-validation is very time-consuming.
\com{We also mix deep learning and feature engineering methods in Table \ref{tab:comparsion_score}, and DL\textbar{}FE and FE\textbar{}DL also support the above analysis results.}

\subsection{Hyperparameter Sensitivity Studies (\textbf{Q5})}
~\TheName~ involves several parameters (e.g., Threshold $thre$, \com{MinHash signature output dimension} and Maximum Order).
 To investigate
the robustness of our method, we examine how the different choices of parameters affect the performance of ~\TheName~. Except for the parameter being tested, we set other parameters at the default values.

We use auto-sklearn for training and validation the \com{FPE} model.
$thre$ divides the features into positive and negative ones.
The \com{FPE} model is trained on the approximate hashing features from the original public training set and validated on the approximate hashing features from the validation set.

\begin{figure}[h]
    \centering
    \vspace{-0.2cm}
    \subfloat[Threshold $thre$.]{\includegraphics[width=0.16\textwidth, height=0.09\textheight]{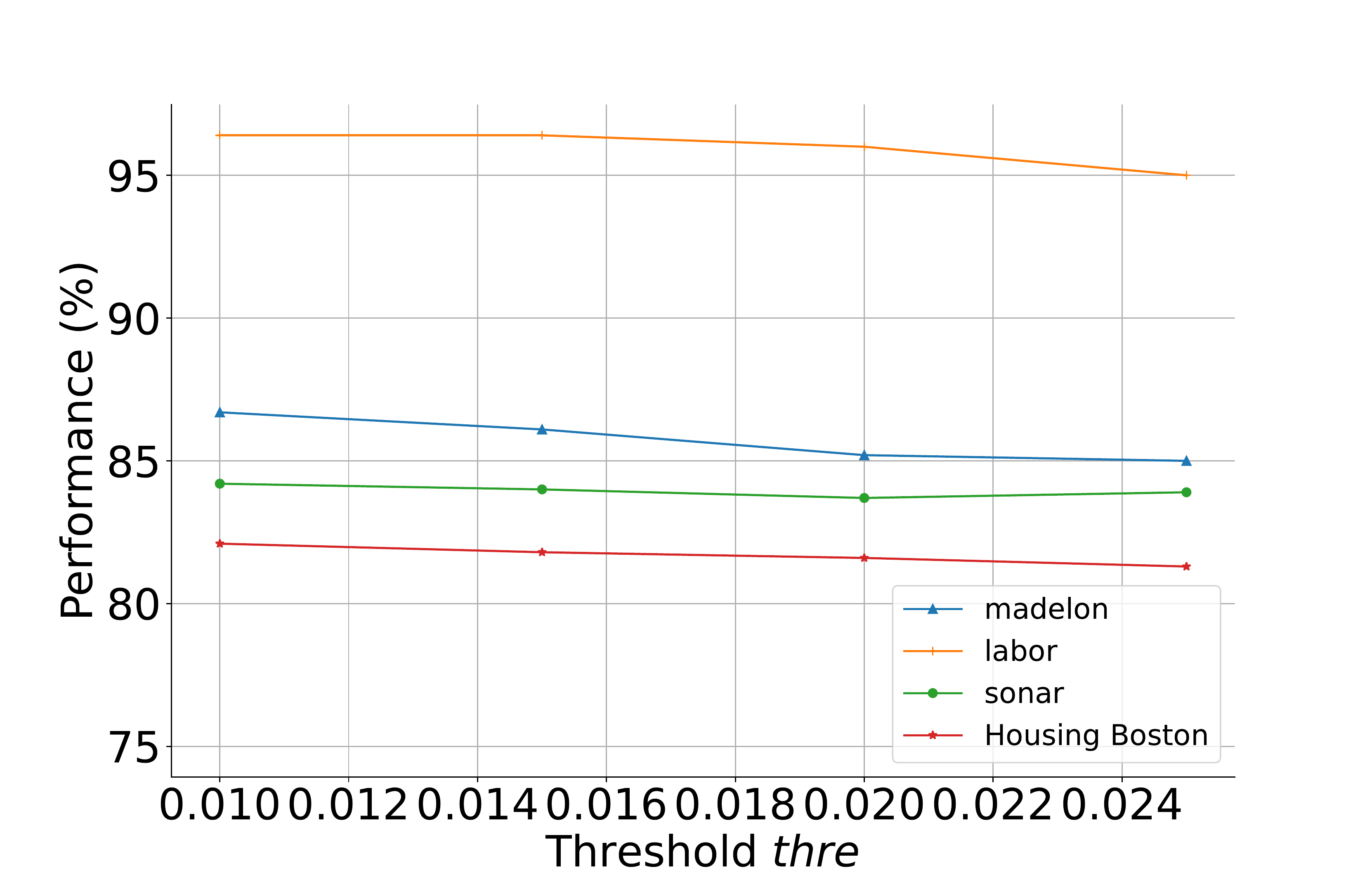}}
    \subfloat[
    \com{MinHash signature output dimension}
    .]{\includegraphics[width=0.16\textwidth, height=0.09\textheight]{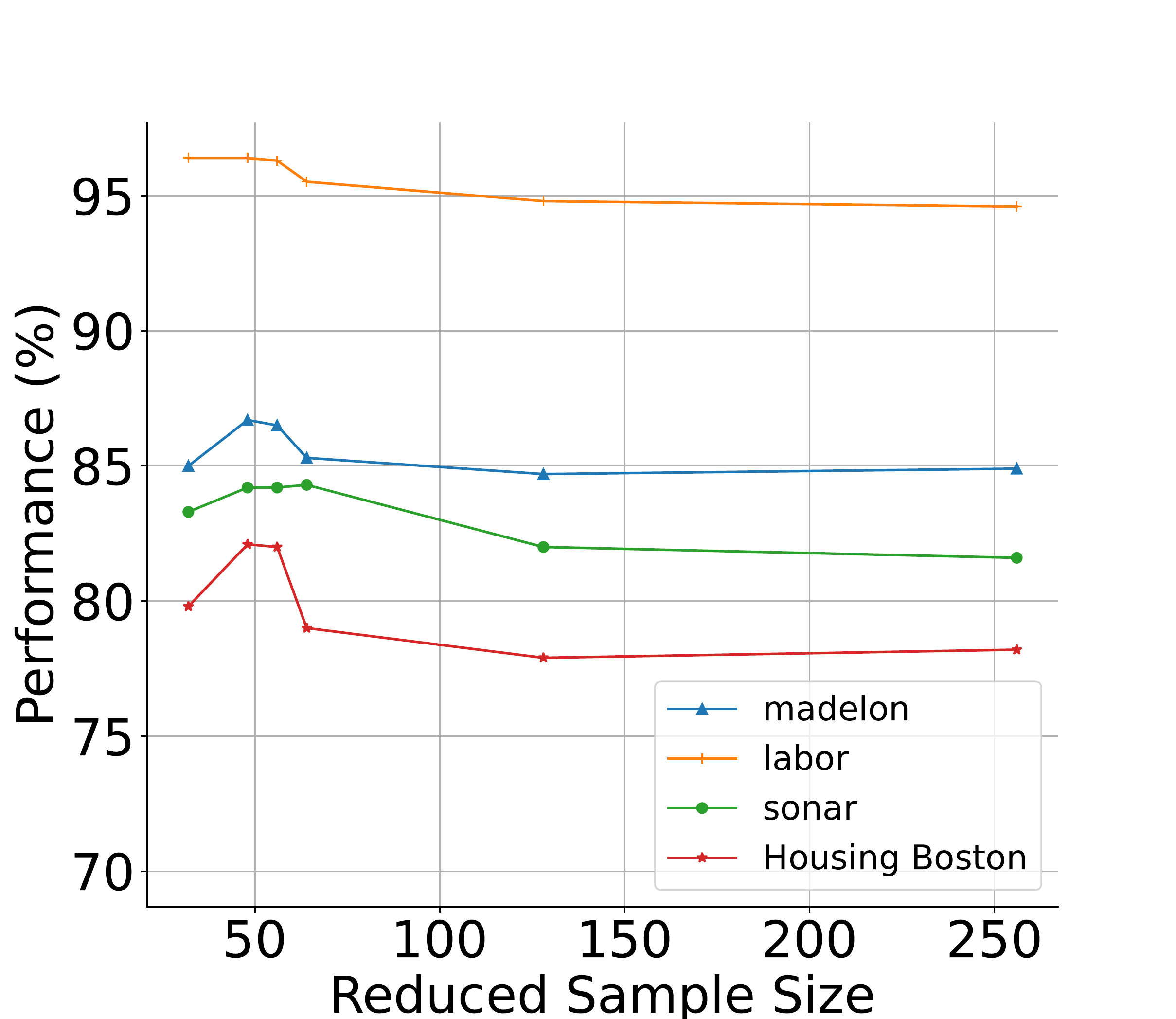}}\
    \subfloat[Maximum Order.]{\includegraphics[width=0.16\textwidth, height=0.09\textheight]{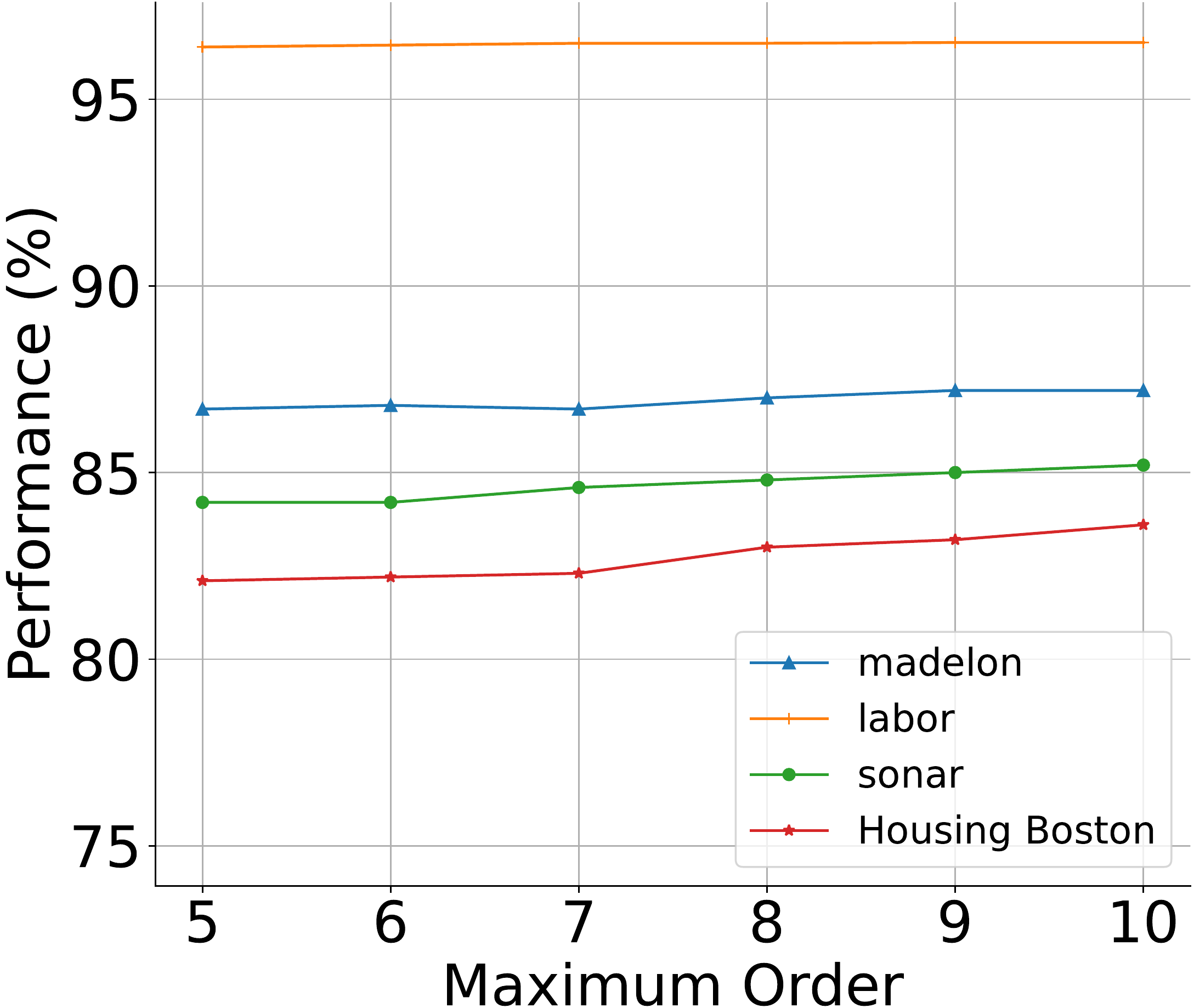}}\
    \caption{Hyperparameter sensitivity studies of ~\TheName~.}
    \vspace{-0.5cm}
    \label{fig:hyperparameter_sensitivity}
\end{figure}

Figure \ref{fig:hyperparameter_sensitivity} shows the evaluation results on data as a function of one selected parameter when fixing others. Overall, we observe that ~\TheName~ is not strictly sensitive to these parameters, demonstrating our proposed framework's robustness. In particular, we can observe that score achieves better performance with the decrease of the threshold. The reason is that a smaller threshold has a larger recall for the validation set. 
The MinHash output length have something with the score. 
The reason is that the approximate features generated by the hash function can affect training results. This paper uses a variety of hash functions and different compressed feature lengths to generate approximate features. 
If the \com{MinHash signature output dimension}
set is too small, it will extract too little information from the original feature and is not conducive to hash model training.
Some datasets may get a better score with the maximum order increase, but the evaluated features and training time will increase significantly. Our experiments set the maximum order as 5 due to the effectiveness and computational cost trade-offs.

\subsection{\com{MinHash Explanations and Signature Sensitivity (\textbf{Q6})}}
\com{On the one hand, MinHash can map arbirtary size samples into the fixed size (like other Hash functions do), which is necessary for across datasets data engineering; 
on the other hand, one unique property of MinhHash is to quickly estimate the similarity between two samples with hashing signature, which can capture and preserve the relationships between samples during hashing, leading to limited information loss caused by compression~\cite{wu2020review}.
Table \ref{tab:comparsion_score} shows little difference in the effect obtained by different MinHash functions.}

% \vspace{-0.2cm}
\subsection{\com{Replace downstream task (\textbf{Q7})}}
% \vspace{-0.3cm}
\begin{table}[h]
\centering
\caption{\com{Comparison results on 36 datasets. C is classification, R is regression. $\mathrm{AutoFS_{R}}$ \cite{fan2020autofs,fan2021interactive} is feature selection from randomly generated features. $\mathrm{NFS}$ \cite{chen2019neural} is feature generation and evaluation. 
SVM is support vector machines, NB is gaussian naive bayes for classification, GP is gaussian processes for regression, MLP is multi-layer perceptron.}}
\label{tab:downstream_task_all}
\vspace{-0.1cm}
\begin{adjustbox}{scale=0.65}
\com{
\begin{tabular}{|c|l|l|l|l|l|l|l|l|l|l|} 
\hline
\multirow{2}{*}{Dataset} & \multirow{2}{*}{C\textbackslash{}R} & \multicolumn{3}{c|}{$\mathrm{AutoFS_{R}}$} & \multicolumn{3}{c|}{NFS}       & \multicolumn{3}{c|}{E-AFE}                        \\ 
\cline{3-11}
                         &                                     & SVM   & NB\textbar{}GP & MLP               & SVM   & NB\textbar{}GP & MLP   & SVM            & NB\textbar{}GP & MLP             \\ 
\hline
fertility                & C                                   & 0.880 & 0.670          & 0.760             & 0.870 & 0.140          & 0.300 & \textbf{0.880} & \textbf{0.880} & \textbf{0.900}  \\ 
\hline
hepatitis                & C                                   & 0.794 & 0.819          & 0.561             & 0.774 & 0.316          & 0.329 & \textbf{0.813} & \textbf{0.826} & \textbf{0.813}  \\ 
\hline
labor                    & C                                   & 0.648 & 0.839          & 0.482             & 0.582 & 0.667          & 0.333 & \textbf{0.876} & \textbf{0.911} & \textbf{0.808}  \\ 
\hline
PimaIndian               & C                                   & 0.762 & 0.749          & 0.569             & 0.638 & 0.352          & 0.469 & \textbf{0.779} & \textbf{0.772} & \textbf{0.626}  \\ 
\hline
credit-a                 & C                                   & 0.655 & 0.657          & 0.536             & 0.545 & 0.555          & 0.445 & \textbf{0.733} & \textbf{0.748} & \textbf{0.713}  \\ 
\hline
diabetes                 & C                                   & 0.762 & 0.757          & 0.509             & 0.642 & 0.352          & 0.397 & \textbf{0.770} & \textbf{0.768} & \textbf{0.638}  \\ 
\hline
german credit            & C                                   & 0.713 & 0.727          & 0.663             & 0.688 & 0.302          & 0.390 & \textbf{0.734} & \textbf{0.754} & \textbf{0.731}  \\ 
\hline
ionosphere               & C                                   & 0.937 & 0.869          & 0.539             & 0.632 & 0.442          & 0.388 & \textbf{0.952} & \textbf{0.906} & \textbf{0.821}  \\ 
\hline
sonar                    & C                                   & 0.558 & 0.590          & 0.518             & 0.447 & 0.441          & 0.335 & \textbf{0.683} & \textbf{0.667} & \textbf{0.721}  \\ 
\hline
spambase                 & C                                   & 0.606 & 0.826          & 0.899             & 0.568 & 0.388          & 0.399 & \textbf{0.810} & \textbf{0.847} & \textbf{0.914}  \\ 
\hline
SPECTF267                & C                                   & 0.790 & 0.670          & 0.701             & 0.775 & 0.513          & 0.323 & \textbf{0.839} & \textbf{0.794} & \textbf{0.798}  \\ 
\hline
AP. lung                 & C                                   & 0.695 & 0.936          & 0.626             & 0.586 & 0.404          & 0.339 & \textbf{0.936} & \textbf{0.946} & \textbf{0.867}  \\ 
\hline
lymph                    & C                                   & 0.899 & 0.935          & 0.682             & 0.499 & 0.456          & 0.363 & \textbf{0.928} & \textbf{0.950} & \textbf{0.884}  \\ 
\hline
lymphography             & C                                   & 0.810 & 0.713          & 0.570             & 0.563 & 0.465          & 0.352 & \textbf{0.852} & \textbf{0.838} & \textbf{0.768}  \\ 
\hline
madelon                  & C                                   & 0.664 & 0.504          & 0.535             & 0.469 & 0.468          & 0.435 & \textbf{0.751} & \textbf{0.621} & \textbf{0.578}  \\ 
\hline
megawatt1                & C                                   & 0.893 & 0.806          & 0.565             & 0.873 & 0.166          & 0.222 & \textbf{0.897} & \textbf{0.893} & \textbf{0.893}  \\ 
\hline
messidor features        & C                                   & 0.693 & 0.611          & 0.517             & 0.513 & 0.467          & 0.476 & \textbf{0.709} & \textbf{0.643} & \textbf{0.673}  \\ 
\hline
AP. ovary                & C                                   & 0.804 & 0.789          & 0.636             & 0.695 & 0.284          & 0.280 & \textbf{0.822} & \textbf{0.829} & \textbf{0.775}  \\ 
\hline
secom                    & C                                   & 0.934 & 0.720          & 0.906             & 0.925 & 0.262          & 0.712 & \textbf{0.934} & \textbf{0.919} & \textbf{0.931}  \\ 
\hline
A. Employee              & C                                   & 0.942 & 0.912          & 0.890             & 0.942 & 0.746          & 0.683 & \textbf{0.942} & \textbf{0.942} & \textbf{0.942}  \\ 
\hline
svmguide3                & C                                   & 0.762 & 0.817          & 0.763             & 0.749 & 0.240          & 0.314 & \textbf{0.817} & \textbf{0.825} & \textbf{0.778}  \\ 
\hline
Wine Q. Red              & C                                   & 0.511 & 0.528          & 0.475             & 0.425 & 0.459          & 0.334 & \textbf{0.537} & \textbf{0.553} & \textbf{0.527}  \\ 
\hline
Wine Q. White            & C                                   & 0.458 & 0.436          & 0.374             & 0.441 & 0.159          & 0.337 & \textbf{0.477} & \textbf{0.479} & \textbf{0.476}  \\ 
\hline
credit default           & C                                   & 0.779 & 0.380          & 0.660             & 0.778 & 0.222          & 0.492 & \textbf{0.808} & \textbf{0.780} & \textbf{0.777}  \\ 
\hline
gisette                  & C                                   & 0.580 & 0.555          & 0.880             & 0.498 & 0.486          & 0.506 & \textbf{0.962} & \textbf{0.900} & \textbf{0.920}  \\ 
\hline
higgs boson              & C                                   & 0.704 & 0.685          & 0.695             & 0.669 & 0.331          & 0.432 & \textbf{0.735} & \textbf{0.701} & \textbf{0.720}  \\ 
\hline
Bikeshare DC             & R                                   & 0.977 & 0.773          & 0.998             & 0.976 & 0.773          & 0.998 & \textbf{0.977} & \textbf{0.773} & \textbf{0.999}  \\ 
\hline
boston housing           & R                                   & 0.614 & 0.468          & 0.509             & 0.612 & 0.467          & 0.476 & \textbf{0.614} & \textbf{0.468} & \textbf{0.525}  \\ 
\hline
Airfoil                  & R                                   & 0.680 & 0.311          & 0.323             & 0.670 & 0.311          & 0.323 & \textbf{0.680} & \textbf{0.314} & \textbf{0.341}  \\ 
\hline
openml 618               & R                                   & 0.631 & 0.523          & 0.218             & 0.630 & 0.523          & 0.438 & \textbf{0.633} & \textbf{0.523} & \textbf{0.671}  \\ 
\hline
openml 589               & R                                   & 0.638 & 0.548          & 0.543             & 0.637 & 0.548          & 0.249 & \textbf{0.641} & \textbf{0.548} & \textbf{0.605}  \\ 
\hline
openml 616               & R                                   & 0.552 & 0.486          & 0.324             & 0.549 & 0.486          & 0.459 & \textbf{0.558} & \textbf{0.486} & \textbf{0.636}  \\ 
\hline
openml 607               & R                                   & 0.623 & 0.485          & 0.369             & 0.620 & 0.485          & 0.497 & \textbf{0.624} & \textbf{0.485} & \textbf{0.606}  \\ 
\hline
openml 620               & R                                   & 0.619 & 0.521          & 0.340             & 0.618 & 0.518          & 0.488 & \textbf{0.622} & \textbf{0.521} & \textbf{0.265}  \\ 
\hline
openml 637               & R                                   & 0.519 & 0.468          & 0.404             & 0.517 & 0.468          & 0.409 & \textbf{0.523} & \textbf{0.473} & \textbf{0.575}  \\ 
\hline
openml 586               & R                                   & 0.647 & 0.523          & 0.756             & 0.646 & 0.523          & 0.726 & \textbf{0.649} & \textbf{0.523} & \textbf{0.759}  \\
\hline
\end{tabular}
}
\vspace{-0.3cm}
\end{adjustbox}
\end{table}
% \vspace{-0.2cm}

\com{We add one more experiment to verify that the features generated by our AFE are also robust to other downstream tasks. 
Specifically, we cache the features generated by different AFEs and replace Random Forest with other models ({\it i.e.}, taking widely-used SVM as an exmaple) to evaluate the quality of these features. 
We also list the results in Table \ref{tab:downstream_task_all} for your reference. 
The results indicate that with our proposed framework, the features obtained from random forest can consistently outperform baselines using SVM, which indicates the features selected by our method are robust to other models.}

\subsection{\com{Improvement analysis (\textbf{Q8})}}

\begin{table}[h]
\centering
% \vspace{-0.3cm}
\caption{\com{P-value of ~\TheName~ with other baseline methods}}
\label{tab:p-value}
\begin{adjustbox}{scale=0.85}
\com{
\begin{tabular}{|l|l|l|l|} 
\hline
     P-value  & $AutoFS_R$\textbar{} E-AFE         & $RTDL_N$  \textbar{} E-AFE           & NFS  \textbar{} E-AFE                \\ 
\hline
Performance  & $5.08 \times 10^{-2}$ & $9.87 \times 10^{-7}$  & $1.83 \times 10^{-1}$  \\ 
\hline
Time   & $1.97 \times 10^{-6}$ & $2.75 \times 10^{-11}$ & $3.75 \times 10^{-6}$  \\
\hline
\end{tabular}}
\end{adjustbox}
\vspace{-0.2cm}
\end{table}

\com{Specifically, we calculate the $p-value$ of the improvement over each baseline method in terms of both effectiveness (accuracy) and efficiency (running time) to check the significance.
The results in Table \ref{tab:p-value} indicate that for efficiency, $p-value$ of improvement over baselines NFS,  $AutoFS_{R}$, and $RTDL_{N}$ are $3.75 \times 10^{-6}$, $1.97 \times 10^{-11}$, and $2.75 \times 10^{-11}$, respectively, which shows that the improvement on efficiency is statistically significant; 
for effectiveness, the improvement over $RTDL_{N}$ is statistically significant with $p-value$ as $9.87\times 10^{-7}$, and the improvement over $AutoFS_{R}$ is statistically near-significant with $p-value$ as $5.08\times 10^{-2}$. 
But the effectiveness improvement over NFS is not statistically significant with the $p-value$ as $1.83\times 10^{-1}$. 
The reason is that the difference of our method over NFS is mainly on developing the two-stage training strategy with reduced sample/feature size to improve the efficiency, while both of the two methods exploits reinforcement learning-based cross-validation for feature evaluation. 
Thus, the improvement on efficiency is statistically significant, while the effectiveness improvement is incremental.}

\subsection{\com{Scalability analysis (\textbf{Q9})}}

\begin{figure}[h]
\centering
\vspace{-0.6cm}
\com{
\subfloat[Performance vs. Feature Number]{\includegraphics[width=0.11\textwidth, height=0.07\textheight]{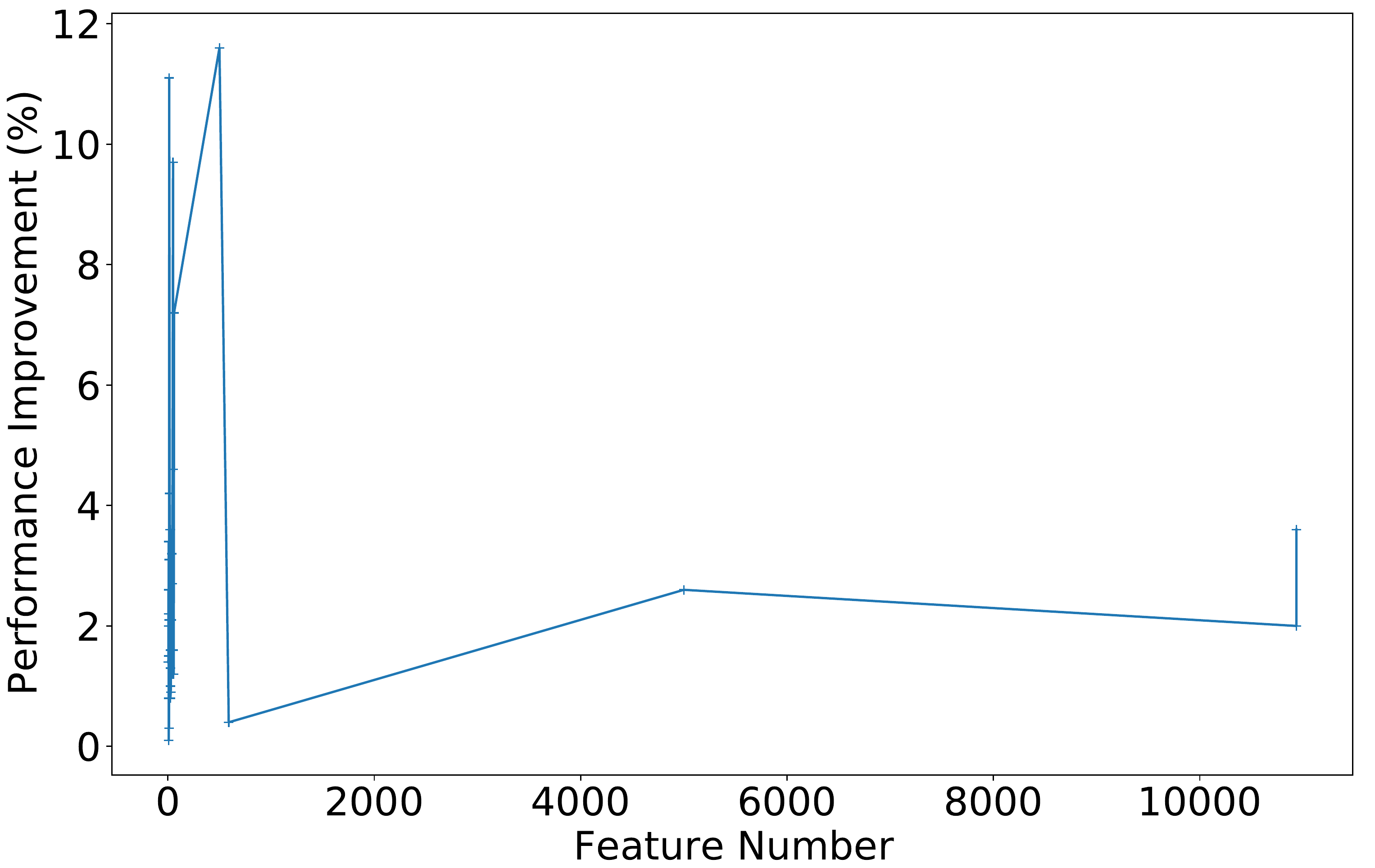}}\
\subfloat[Performance vs. Sample Number]{\includegraphics[width=0.11\textwidth, height=0.07\textheight]{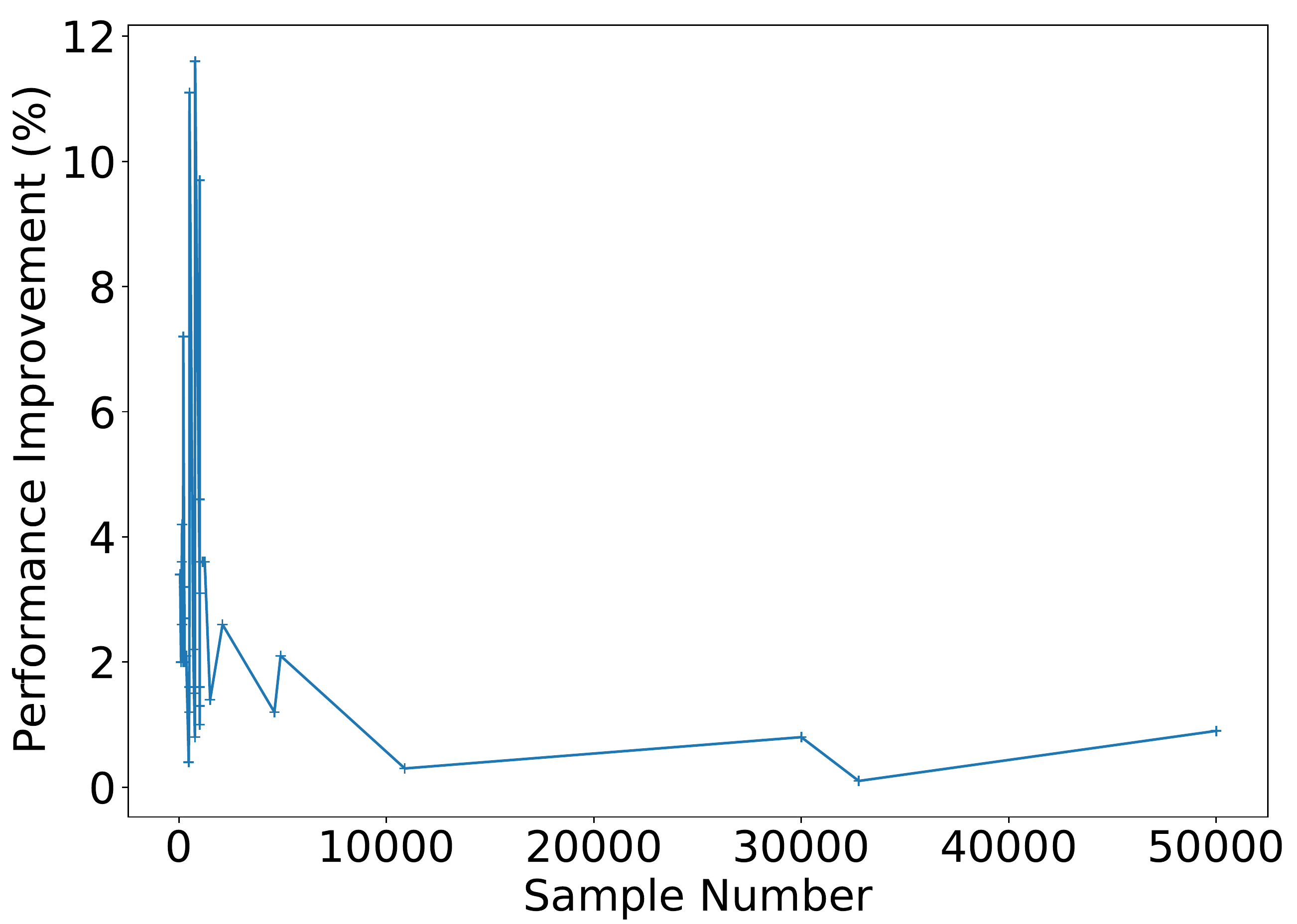}}\
\subfloat[Performance vs. Feature Number]{\includegraphics[width=0.11\textwidth, height=0.07\textheight]{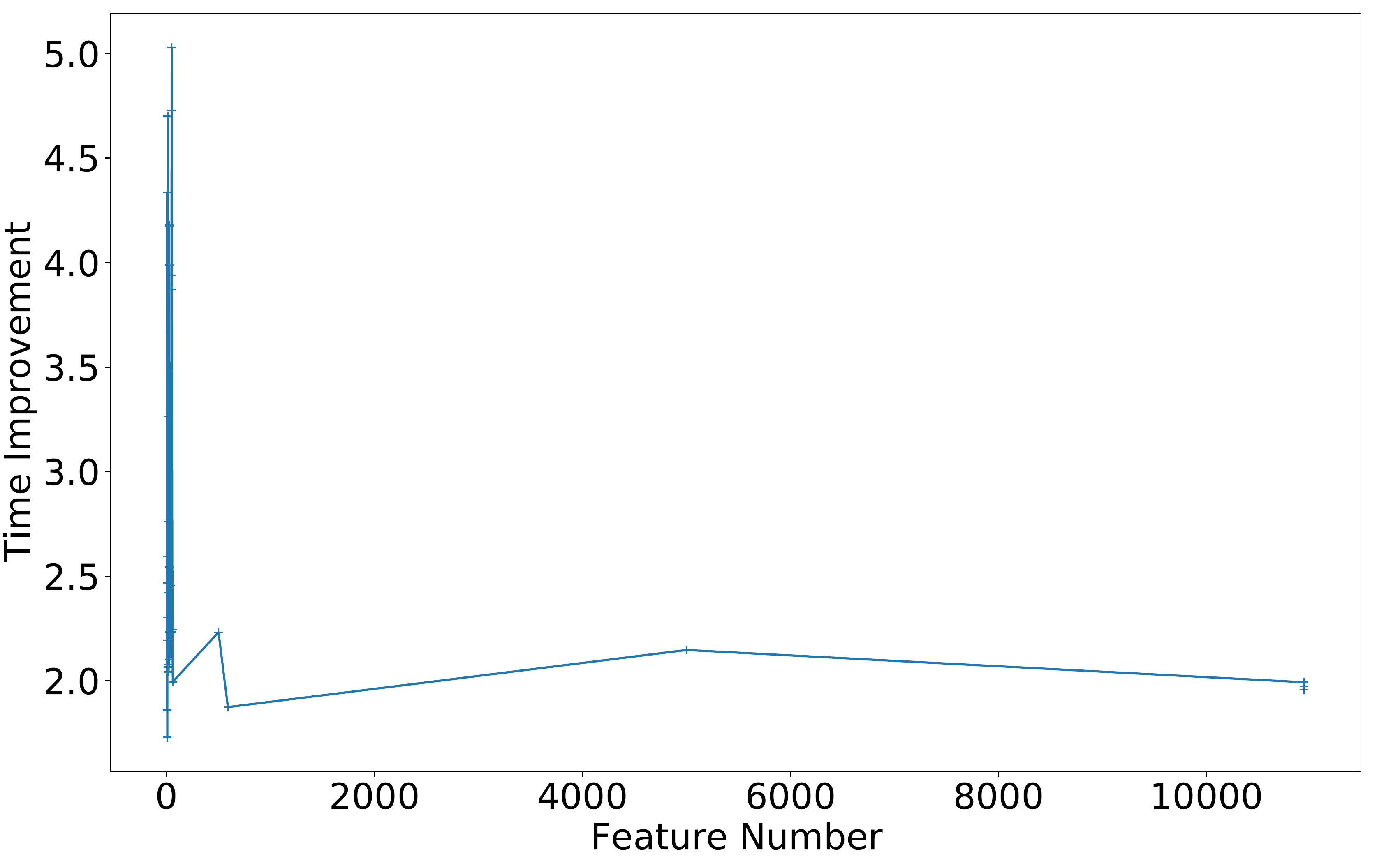}}\
\subfloat[Performance vs. Sample Number]{\includegraphics[width=0.11\textwidth, height=0.07\textheight]{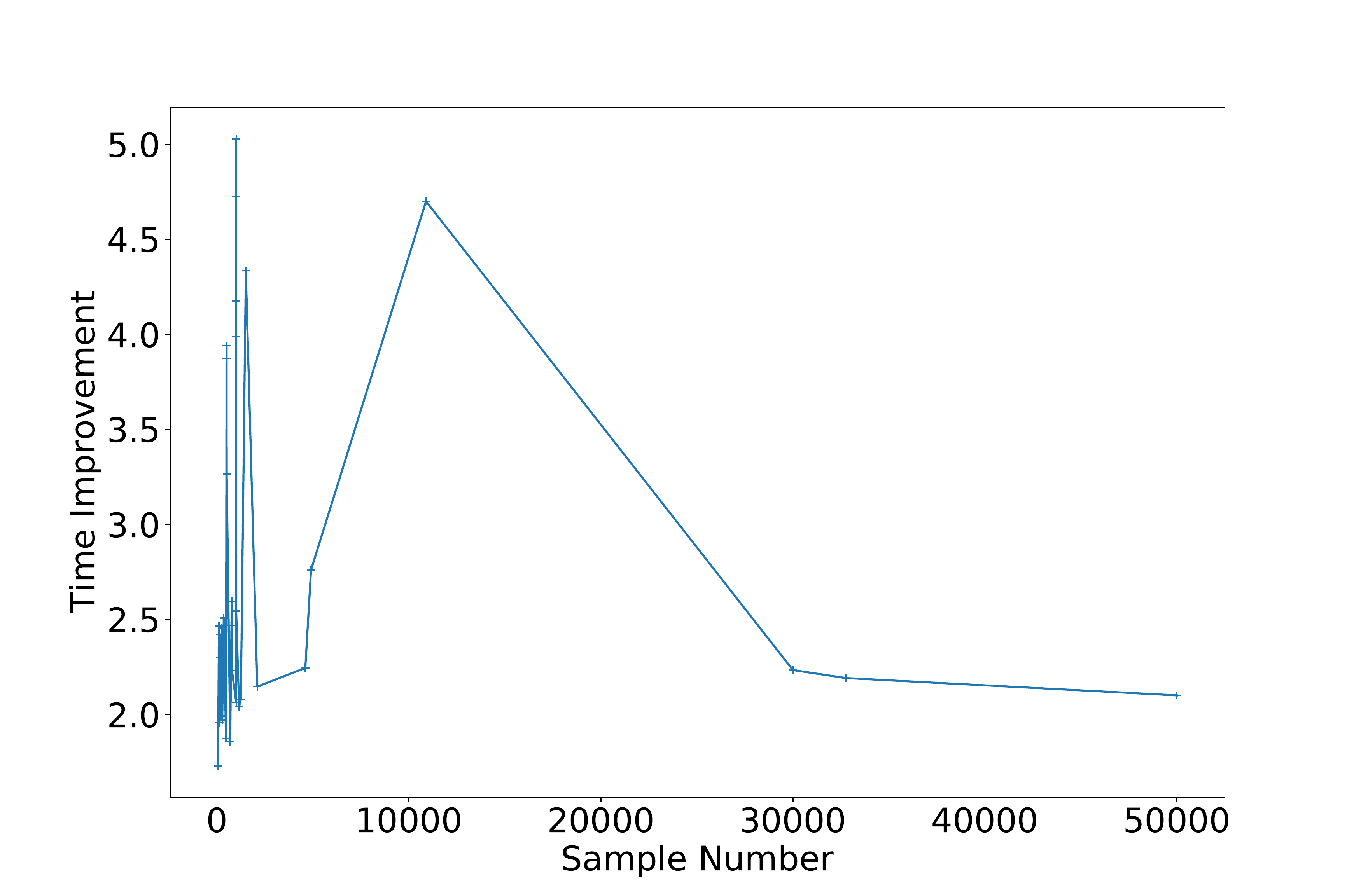}}\
}
\caption{\com{Time or performance improvement vs. feature or sample size.}}
\label{fig:scalability}
\vspace{-0.4cm}
\end{figure}

\com{Figure \ref{fig:scalability} results show the relationship between running time and performance improvement with data size. The performance improvement of E-AFE on larger datasets are better on smaller dataset, which demonstrate scalability ability of our proposed method.}
\section{Related Work}

\subsection{Automated Feature Engineering}
\com{
AFE method can be divided into three types: 
(1)The downstream task has no feedback to the feature generator.
ExploreKit \cite{katz2016explorekit} performs all transformation functions on the dataset. 
AutoLearn \cite{kaul2017autolearn} preprocesses raw features and discards features with low information gain. 
Learning Feature Engineering (LFE) \cite{nargesian2017learning} uses feature-class representation and a set of Multi-Layer Perception (MLP) classifiers to predict whether the transformation result is better than the original feature. 
(2)The downstream task gives feedback to the feature generator with RL. Transformation Graph \cite{khurana2018feature} builds DAG and uses Q-learning to exploit high-order features. 
Neural Feature Search (NFS) \cite{chen2019neural} predicts the most appropriate transformation for each feature by policy gradient for better feature generation.
Group-wise Reinforcement Feature Generation (GRFG) \cite{wang2022group} proposes a principled framework to address the automation, explicitness, optimal issues in representation space reconstruction.
(3)Feature selection with RL ignores feature generation.
Multi-Agent Reinforcement Learning Feature Selection (MARLFS), Interactive Reinforced Feature Selection (IRFS), Single-Agent Reinforcement Learning Framework (SADRLFS), Monte Carlo based reinforced feature selection (MCRFS), Group-based Interactive Reinforced Feature Selection (GIRFS) and Combinatorial Multi-Armed Bandit (CMAB) on feature selection \cite{liu2019automating,fan2020autofs,liu2021automated,fan2021interactive,zhao2020simplifying,liu2021efficient,fan2021autogfs,liu2021multi} use RL in feature selection. However, those RL frameworks can't consider feature generation. 
}

\subsection{Approximate Feature}
\com{
To speed up feature engineering, many methods to obtain an approximate dataset from the original dataset are proposed. We summarize four classes of approximate feature methods. 
(1)Meta-Feature. Previous approaches used hand-crafted meta-features, including information-theoretic and statistical meta-features, to represent datasets \cite{michie1995machine,kalousis2001model,feurer2015efficient,katz2016explorekit}. 
Meta-Feature Extractor (MFE) \cite{rivolli2018towards,alcobacca2020mfe} extracts meta-features from the dataset to improve the reproducibility of machine learning. 
Automatic generation of meta-features \cite{pinto2016towards} presents a framework to generate meta-features in the context of meta-learning systematically. 
ExploreKit \cite{katz2016explorekit} generates two types of meta-features: dataset-based and candidate features-based. 
(2)Low-rank matrix approximation \cite{frieze2004fast,nguyen2009a} finds a smaller rank matrix similar to the original matrix. 
(3)Quantile Data Sketch was used in LFE \cite{nargesian2017learning} to represent feature values.
(4)Hashing method. 
Feature Hashing \cite{weinberger2009feature} introduces specialized hash functions with unbiased inner products that are directly applicable to a large variety of kernel methods. 
Hash kennel \cite{shi2009hasha,shi2009hashb} computes the kernel matrix for data streams and sparse feature spaces.
}

\section{Conclusion Remarks}
In this paper, we studied how to improve the efficiency of AFE. 
Based on the empirical studies, we identified that the time-consuming feature evaluation procedure is the core reason of interfering with AFE's running efficiency. 
We further validated that smaller sample size and feature size would accelerate feature evaluation. 
Therefore, we proposed to improve the efficiency of AFE by reducing the sample and feature size. 
Specifically, we develop FPE model, including a sample compressor with $\text{MinHash}$ to reduce sample size , and a feature pre-selector with a pre-trained binary classifier to distinguish the effectiveness of the generated features. 
Moreover, we devised a two-stage training strategy to first initialize the policy using the binary feature-effectiveness classification as the evaluation task to borrow external knowledge from the pre-trained FPE. 
The enhanced initialization provideed the AFE policy with a good starting point for exploring the optimal actions, which further significantly save the running time. 
Finally, we conduct extensive experiments on 36 datasets on both the classification and regression tasks.
When compared to state-of-the-art AFE methods, the results show 2.9 percent higher average performance and 2x higher computational efficiency. 
The improvements over both the effectiveness and efficiency proved that FPE model can remove the redundancy in candidate features.

\section{Acknowledgements}
This paper is jointly funded by National Key R\&D Program of China (No. 2019YFB2102100), The Science and Technology Development Fund of Macau SAR (File no. 0015/2019/AKP), Key-Area Research and Development Program of Guangdong Province (NO.2020B010164003), GuangDong Basic and Applied Basic Research Foundation (No. 2020B1515130004),
The Science and Technology Development Fund, Macau SAR (File no. SKL-IOTSC-2021-2023 to Chengzhong Xu and Pengyang Wang), and The Start-up Research Grant of University of Macau (File no. SRG2021-00017-IOTSC to Pengyang Wang).

% \newpage
\bibliographystyle{IEEEtran}
\bibliography{reference}

\vspace{12pt}
% \color{red}
% IEEE conference templates contain guidance text for composing and formatting conference papers. Please ensure that all template text is removed from your conference paper prior to submission to the conference. Failure to remove the template text from your paper may result in your paper not being published.

\end{document}